\documentclass{article}

\usepackage[accepted]{icml2025}
\usepackage{times}
\usepackage{etoolbox}

\let\underbar\undefined

\usepackage{microtype}
\usepackage{hhline}
\usepackage{wrapfig}
\usepackage{amsthm}
\usepackage{mathtools}
\usepackage{amsmath}
\usepackage{bbm}
\usepackage{amsfonts}
\usepackage{amssymb}
\usepackage{dsfont}

\makeatletter
\newcommand{\neutralize}[1]{\expandafter\let\csname c@#1\endcsname\count@}
\makeatother

\newenvironment{thmmod}[2]
  {\renewcommand{\thetheorem}{\ref*{#1}#2}%
   \neutralize{theorem}\phantomsection
   \begin{theorem}}
  {\end{theorem}}

  \newenvironment{cormod}[2]
  {\renewcommand{\thecorollary}{\ref*{#1}#2}%
   \neutralize{corollary}\phantomsection
   \begin{corollary}}
  {\end{corollary}}

\usepackage{thmtools}
\usepackage{thm-restate}

\makeatletter
\let\save@mathaccent\mathaccent
\newcommand*\if@single[3]{%
  \setbox0\hbox{${\mathaccent"0362{#1}}^H$}%
  \setbox2\hbox{${\mathaccent"0362{\kern0pt#1}}^H$}%
  \ifdim\ht0=\ht2 #3\else #2\fi
  }
\newcommand*\rel@kern[1]{\kern#1\dimexpr\macc@kerna}
\newcommand*\widebar[1]{\@ifnextchar^{{\wide@bar{#1}{0}}}{\wide@bar{#1}{1}}}
\newcommand*\underbar[1]{\@ifnextchar_{{\under@bar{#1}{0}}}{\under@bar{#1}{1}}}
\newcommand*\wide@bar[2]{\if@single{#1}{\wide@bar@{#1}{#2}{1}}{\wide@bar@{#1}{#2}{2}}}
\newcommand*\under@bar[2]{\if@single{#1}{\under@bar@{#1}{#2}{1}}{\under@bar@{#1}{#2}{2}}}
\newcommand*\wide@bar@[3]{%
  \begingroup
  \def\mathaccent##1##2{%
    \let\mathaccent\save@mathaccent
    \if#32 \let\macc@nucleus\first@char \fi
    \setbox\z@\hbox{$\macc@style{\macc@nucleus}_{}$}%
    \setbox\tw@\hbox{$\macc@style{\macc@nucleus}{}_{}$}%
    \dimen@\wd\tw@
    \advance\dimen@-\wd\z@
    \divide\dimen@ 3
    \@tempdima\wd\tw@
    \advance\@tempdima-\scriptspace
    \divide\@tempdima 10
    \advance\dimen@-\@tempdima
    \ifdim\dimen@>\z@ \dimen@0pt\fi
    \rel@kern{0.6}\kern-\dimen@
    \if#31
      \overline{\rel@kern{-0.6}\kern\dimen@\macc@nucleus\rel@kern{0.4}\kern\dimen@}%
      \advance\dimen@0.4\dimexpr\macc@kerna
      \let\final@kern#2%
      \ifdim\dimen@<\z@ \let\final@kern1\fi
      \if\final@kern1 \kern-\dimen@\fi
    \else
      \overline{\rel@kern{-0.6}\kern\dimen@#1}%
    \fi
  }%
  \macc@depth\@ne
  \let\math@bgroup\@empty \let\math@egroup\macc@set@skewchar
  \mathsurround\z@ \frozen@everymath{\mathgroup\macc@group\relax}%
  \macc@set@skewchar\relax
  \let\mathaccentV\macc@nested@a
  \if#31
    \macc@nested@a\relax111{#1}%
  \else
    \def\gobble@till@marker##1\endmarker{}%
    \futurelet\first@char\gobble@till@marker#1\endmarker
    \ifcat\noexpand\first@char A\else
      \def\first@char{}%
    \fi
    \macc@nested@a\relax111{\first@char}%
  \fi
  \endgroup
}
\newcommand*\under@bar@[3]{%
  \begingroup
  \def\mathaccent##1##2{%
    \let\mathaccent\save@mathaccent
    \if#32 \let\macc@nucleus\first@char \fi
    \setbox\z@\hbox{$\macc@style{\macc@nucleus}_{}$}%
    \setbox\tw@\hbox{$\macc@style{\macc@nucleus}{}_{}$}%
    \dimen@\wd\tw@
    \advance\dimen@-\wd\z@
    \divide\dimen@ 3
    \@tempdima\wd\tw@
    \advance\@tempdima-\scriptspace
    \divide\@tempdima 10
    \advance\dimen@-\@tempdima
    \ifdim\dimen@>\z@ \dimen@0pt\fi
    \rel@kern{0.6}\kern-\dimen@
    \if#31
      \underline{\rel@kern{-0.6}\kern\dimen@\macc@nucleus\rel@kern{0.4}\kern\dimen@}%
      \advance\dimen@0.4\dimexpr\macc@kerna
      \let\final@kern#2%
      \ifdim\dimen@<\z@ \let\final@kern1\fi
      \if\final@kern1 \kern-\dimen@\fi
    \else
      \underline{\rel@kern{-0.6}\kern\dimen@#1}%
    \fi
  }%
  \macc@depth\@ne
  \let\math@bgroup\@empty \let\math@egroup\macc@set@skewchar
  \mathsurround\z@ \frozen@everymath{\mathgroup\macc@group\relax}%
  \macc@set@skewchar\relax
  \let\mathaccentV\macc@nested@a
  \if#31
    \macc@nested@a\relax111{#1}%
  \else
    \def\gobble@till@marker##1\endmarker{}%
    \futurelet\first@char\gobble@till@marker#1\endmarker
    \ifcat\noexpand\first@char A\else
      \def\first@char{}%
    \fi
    \macc@nested@a\relax111{\first@char}%
  \fi
  \endgroup
}
\makeatother

\usepackage[utf8]{inputenc} % allow utf-8 input
\usepackage[T1]{fontenc}    % use 8-bit T1 fonts
\usepackage{hyperref}       % hyperlinks
\usepackage{url}            % simple URL typesetting
\usepackage{booktabs}       % professional-quality tables
\usepackage{amsfonts}       % blackboard math symbols
\usepackage{nicefrac}       % compact symbols for 1/2, etc.
\usepackage{microtype}      % microtypography
\usepackage{xcolor}         % Allow colored comments
\usepackage{amssymb}        % For mathbb formats
\usepackage{amsmath}        % More complex maths cmds
\usepackage{amsthm}         % Include Theorems
\usepackage{natbib}         % Bibliography
\usepackage[english]{babel}
\usepackage{cleveref}       % for commands like Cref
\usepackage{accents}        % for commands like underbrace
\usepackage{algorithm}      % for printing algorithms
\usepackage{graphicx}       % for importing images
\usepackage{subcaption}

\usepackage{multirow}       % Multirow for tabular results
\usepackage{tabularx,ragged2e} % Dataset tables

\usepackage{tcolorbox}

\usepackage{math_commands_nsdpo}  % Bring across our math commands
\usepackage{caption}
\usepackage[toc,page,header]{appendix}
\usepackage{minitoc}

\definecolor{mygreen}{RGB}{0, 126, 0}

\DeclareMathOperator*{\argmax}{arg\,max}
\DeclareMathOperator*{\argmin}{arg\,min}
\newcolumntype{C}{>{\Centering\arraybackslash}X}

\icmltitlerunning{Right Now, Wrong Then: Non-Stationary Direct Preference Optimization under Preference Drift}

\begin{document}

\twocolumn[
\icmltitle{\quad \quad \quad \quad \quad Right Now, Wrong Then:\newline Non-Stationary Direct Preference Optimization under Preference Drift}

\icmlsetsymbol{equal}{*}

\begin{icmlauthorlist}
\icmlauthor{Seongho Son}{equal,uclcs}
\icmlauthor{William Bankes}{equal,uclcs}
\icmlauthor{Sayak Ray Chowdhury}{iit}
\icmlauthor{Brooks Paige}{uclcs}
\icmlauthor{Ilija Bogunovic}{uclee}
\end{icmlauthorlist}

\icmlaffiliation{uclcs}{Department of Computer Science, University College London, London, United Kingdom}
\icmlaffiliation{uclee}{Department of Electronic and Electrical Engineering, University College London, London, United Kingdom}
\icmlaffiliation{iit}{Department of Computer Science and Engineering, IIT Kanpur, India}

\icmlcorrespondingauthor{Seongho Son}{seong.son.22@ucl.ac.uk}

\icmlkeywords{Machine Learning, ICML}

\vskip 0.3in
]
\printAffiliationsAndNotice{\icmlEqualContribution} % otherwise use the standard text.

\begin{abstract}
Current Large Language Model (LLM) preference optimization algorithms do not account for temporal preference drift, which can lead to severe misalignment. To address this limitation, we propose \textit{Non-Stationary Direct Preference Optimisation} (NS-DPO) that models time-dependent reward functions with a Dynamic Bradley-Terry model. NS-DPO proposes a computationally efficient solution by introducing only a single discount parameter in the loss function, which is used for exponential weighting that proportionally focuses learning on more time-relevant datapoints. We theoretically analyze the convergence of NS-DPO in a general setting where the exact nature of the preference drift is not known, providing upper bounds on the estimation error and regret caused by non-stationary preferences. Finally, we demonstrate the effectiveness of NS-DPO
% \footnote{For code, see \href{https://github.com/geronest/ns-dpo}{https://github.com/geronest/ns-dpo}.} 
for fine-tuning LLMs under drifting preferences. Using scenarios where various levels of preference drift is introduced, with popular LLM reward models and datasets, we show that NS-DPO fine-tuned LLMs remain robust under non-stationarity, significantly outperforming baseline algorithms that ignore temporal preference changes, without sacrificing performance in stationary cases.\looseness=-1
\end{abstract}

\section{Introduction}
% \vspace{-5pt}
The application of Reinforcement Learning from Human Feedback (RLHF) to fine-tune Large Language Models (LLMs) \citep{christiano2017deep, stiennon2020learning, ziegler2019fine, ouyang2022training, bai2022constitutional} has led to more precise control over the behavior they exhibit. This control is crucial when looking to safely deploy models in the real world \citep{amodei2016concrete, hendrycks2022x}. Human preference datasets enable the training of proxy \emph{reward models} (see, e.g., RewardBench \citep{lambert2024rewardbench}) that can evaluate complex human behaviour. These reward models are used in conjunction with RL to fine-tune the LLM. Recent works \citep{rafailov2024direct, azar2024general, hong2024reference} seek to improve the efficiency and stability of these approaches \citep{chaudhari2024rlhf} by training the LLM straight from human preference data, avoiding the need to learn a proxy reward model.\looseness=-1

A key assumption made in these preference optimization algorithms is that human preferences are \emph{stationary}, i.e., they do not change over time. However, a shift in preferences can occur due to new information becoming available \citep{zafari2019modelling, johnson2020temporal}, social influences, and cultural trends. As more preference datasets are gathered over long periods of time, the chance of the data containing varying preferences increases. In such cases, algorithms that do not account for these changes, view them as noise and treat outdated data as equally important as fresh data, often leading to deteriorated performance. An increasing body of evidence \citep{zhou2024lima, chen2024alpagasus} points to data quality as being a key factor in fine-tuning performance, thus preference drift can greatly affect the alignment of models \citep{carroll2024ai}.
The development of preference optimization algorithms and theory to handle preference drifts are therefore crucial.\looseness=-1

\begin{figure*}[ht]
    \centering
\includegraphics[width=1.0\textwidth]{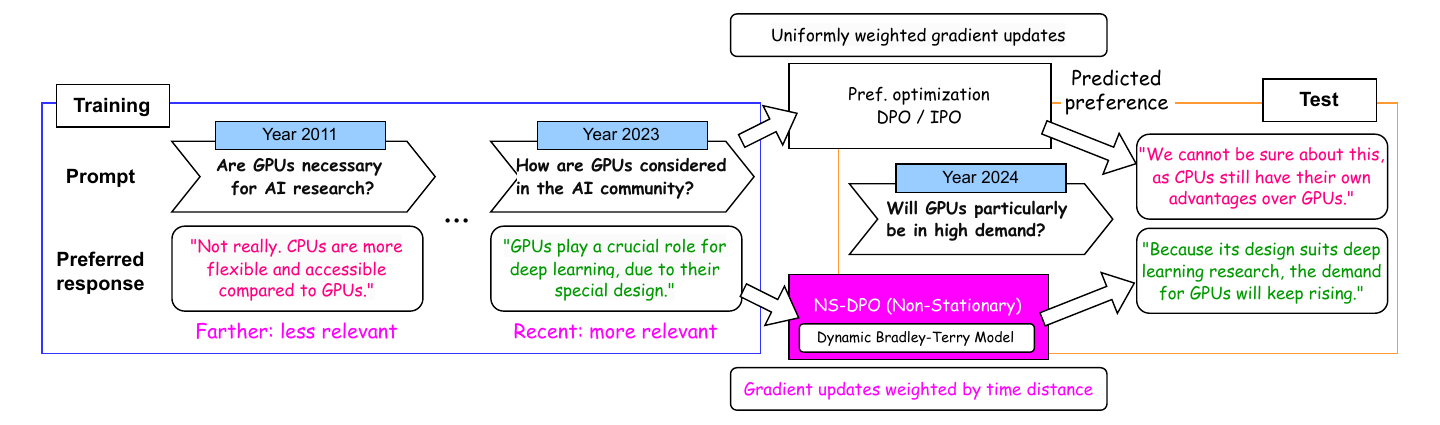}
    \caption{Human preferences are dynamic and influenced by a variety of factors (e.g. environment change and societal influence). However, standard preference optimization approaches (e.g., DPO and IPO \citep{rafailov2024direct, azar2024general}) do not account for this non-stationarity. In contrast, NS-DPO robustly learns on non-stationary data by using a \textbf{Dynamic Bradley-Terry model}, and adjusts the loss to discount older datapoints and concentrate learning on the latest data.\looseness=-1} 
    \label{fig: summary_NSDPO}
\end{figure*}

In this work, we propose \emph{Non-Stationary Direct Preference Optimization} (NS-DPO), a novel approach that uses a probabilistic \emph{Dynamic} Bradley-Terry model \citep{cattelan2013dynamic, bong2020nonparametric, tian2023spectral} to account for non-stationary drift in human preferences with just a single additional parameter. As we only assume knowledge of the total preference drift and not which specific preferences change over time, NS-DPO re-weights each training datapoint, down-weighting older data with potentially stale preferences and up-weighting more recent ones. We empirically show the effectiveness and robustness of NS-DPO compared to stationary approaches.
Our approach is summarized in \Cref{fig: summary_NSDPO}. \looseness=-1

\textbf{Related work.} One of the primary applications of the RLHF framework is fine-tuning large language models (LLMs) \citep{christiano2017deep, stiennon2020learning, ziegler2019fine, ouyang2022training, bai2022constitutional}. A key component of this is the Bradley-Terry model \citep{bradley1952rank} which learns a reward signal from paired human preferences. \cite{rafailov2024direct} propose Direct Preference Optimization (DPO), which implicitly uses the Bradley-Terry model, to fine-tune an LLM directly from a preference dataset. A variety of alternatives to DPO have been proposed which adapt or do not use the Bradley-Terry model \citep{azar2024general, amini2024direct, meng2024simpo, cen2024value, xu2023some}. Other approaches such as maximizing a utility function \citep{ethayarajh2024kto}, improving from the policy in the previous iteration \citep{munos2023nash, rosset2024direct, tang2025game}, and improving the response at inference-time \citep{mudgal2023controlled, son2025robust} have been investigated. Our work is the first to consider a direct preference algorithm using a Dynamic Bradley-Terry model.\looseness=-1

A variety of work has analyzed the RLHF problem from a theoretical standpoint. \cite{xiong2023iterativepl} provide suboptimiality bounds of policies in the offline, online and hybrid settings under linear rewards. They do not directly analyze the performance of DPO, but propose it as a practical implementation of the oracle. \cite{zhu2023principled, chowdhury2024provably} analyse the offline preference learning and DPO settings, respectively. \cite{chowdhury2024provably} address noisy preferences with a modified version of the DPO algorithm, presenting confidence bounds for neural policy classes and suboptimality bounds for the setting with log-linear policies. 

Parameter drift has been widely studied in the bandit literature. \cite{cheung2019learning} propose using a sliding window to estimate parameters with data points close to the current timestep, whilst \cite{bogunovic2016time, zhao2020simple} investigate a restarting strategy.
Similarly to the strategy of \cite{russac2019weighted}, we use an exponentially weighted discounting term to re-weight points close to the current timestep. \cite{faury2021regret, wang2023revisiting} apply this approach to the case of generalised linear bandits first proposed by \cite{filippi2010parametric}. \cite{pacchiano2021dueling, saha2021optimal, mehta2023sample} focus on the duelling bandit setting, where only preference feedback between two actions is provided by the environment. In this work, we provide the \emph{first} theoretical guarantees for the popular offline setting where the true reward parameter (used to label training data) is allowed to change over time. \looseness=-1

\textbf{Main contributions.} We propose NS-DPO, a direct preference optimization method that accounts for non-stationary preferences in the dataset via a Dynamic Bradley-Terry model. NS-DPO modifies the training loss with a single exponential weighting parameter $\gamma$, and thus represents a \emph{simple} and \emph{computationally efficient} extension of the popular DPO algorithm. We provide an upper bound on the regret of NS-DPO for log-linear policies given standard data coverage assumptions used in offline learning. To explore the performance of NS-DPO, we construct \emph{non-stationary preference datasets} from a variety of existing popular datasets; including GlobalOpinionsQA \citep{durmus2023towards}, Helpful \& Harmless \citep{dai2023safe}, and UltraFeedback \citep{cui2023ultrafeedback}. We demonstrate that NS-DPO significantly outperforms stationary DPO and other relevant baselines on these non-stationary datasets with varying degrees of preference drift on Llama LLM models \cite{touvron2023llama, dubey2024llama}.

\section{Preliminaries}
\label{sec:prelims}

\textbf{Stationary RLHF.} In the stationary RLHF setting \citep{ziegler2019fine, ouyang2022training}, the goal is to find a suitable LLM policy $\pi$, whose response $a$, to a prompt $x$, maximise a reward function $r(x,a)$, i.e., \looseness=-1
\begin{align} 
    \J(\pi) &= \Eb_{x \sim \X, a \sim \pi}\Big[r(x, a) - \tau \DKL[\pi(\cdot | x)\| \piref(\cdot | x)]\Big]. \label{eq: stationary rlhf objective}
\end{align}
Here, the KL-divergence prevents the learnt policy from deviating too far from some reference policy $\piref$, that has characteristics we wish to preserve in the final model. This is controlled by the parameter $\tau > 0$. In practical settings, human feedback is too complex to capture in a hand designed reward model, and we resort to learning a model from human preference data.

\textbf{Bradley-Terry Model.} A human preference dataset consists of prompts and two possible responses $\mathcal{D}= \{(x_i, a_i, a_i')\}_{i \in [n]}$, where $a_i$ is the response preferred to $a_i'$, and $n$ is the number of datapoints. To learn a reward model from this dataset, we assume the preferences are generated by a Bradley-Terry (BT) model \citep{bradley1952rank} where the probability that $a_i$ is preferred to $a'_i$ is  
\begin{equation}\label{eq: stationary Bradley-Terry}
    p(a_i \succ a'_i|x_i) = \sigma(r(x_i,a_i) - r(x_i,a'_i)).
\end{equation}
In \Cref{eq: stationary Bradley-Terry}, $\sigma(\cdot)$ is the logistic sigmoid function and $r(x,a)$ is the reward model of human preferences we do not have access to and wish to learn. We parameterise the reward, typically as a single layer MLP on the last layer of the reference policy model $\piref$ \citep{ziegler2019fine}, and then learn the parameters using a maximum likelihood estimator. An LLM can then be fine-tuned on the objective in \Cref{eq: stationary rlhf objective} using Reinforcement Learning (RL). It is important to note that the BT model captures many of the inherent assumptions we make about our data, which include the stationary nature of the underlying data generating process.\looseness=-1

\textbf{Direct Preference Optimization.} Recent work by \citep{rafailov2024direct} avoids the training of an explicit reward model in the stationary RLHF process by optimizing the LLM policy directly from human preference data. To do this, the analytical solution to the stationary RLHF objective is rearranged into \Cref{eq: stationary rlhf objective} to derive an implicit reward 
\begin{equation}\label{eq: stationary implicit reward}
    r(x, a) = \tau \log \frac{\pi(a | x)}{\piref(a | x)} + \tau \log Z(x),
\end{equation}
where $Z(x)$ is a normalisation constant. This is substituted into the negative log likelihood of the Bradley-Terry model (see \Cref{eq: stationary Bradley-Terry}) resulting in the direct preference optimization (DPO) objective\looseness=-1 
\begin{equation}\label{eq: DPO Loss}
\mathcal{L}(\pi) = \sum_{(x, a, a') \in \mathcal{D}} -\log \sigma \left( \tau h_\pi(x, a, a') \right),
\end{equation}
Where $h_\pi(x, a, a') =  \log \tfrac{\pi(a|x)}{\piref(a|x)} - \log \tfrac{\pi(a'|x)}{\piref(a'|x)}$. All the methods introduced in this section, including DPO, are all stationary as they assume the reward model does not change with time. However, this assumption does not hold when training on real-world data. The changes in preferences over time, captured in the dataset, appear as label noise to the stationary methods.\looseness=-1

\section{Learning Under Preference Drift}
\label{sec: NS-DPO}

To address the problem of preference drift, in datasets collected over a period of time, we propose \emph{Non-Stationary Direct Preference Optimization} (NS-DPO). NS-DPO incorporates the \emph{Dynamic Bradley-Terry} model, which includes a non-stationary reward model $r(x,a,t)$. Here $t \in \{1, \dots, T-1\}$ denotes a time step in the past, and $T \in \mathbb{N}_{+}$ denotes the \emph{current time step}, where we are evaluating the trained policy. Under the Dynamic Bradley-Terry model, the probability of response $a_i$ being preferred to $a'_i$ is
\begin{equation}\label{eq: Dynamic Bradley-Terry}
    p(a_i \succ a'_i|x_i, t_i) = \sigma(h_r(x_i, a_i, a_i', t_i)),
\end{equation}
where $h_r(x, a, a', t) = r(x,a,t) - r(x,a',t)$. We also assume the dataset has temporal information about when the human preference between the two responses is expressed, $\mathcal{D} = \{(x_i, a_i, a'_i, t_i\}_{i \in [n]}$. For the ease of indexing datapoints, we assume $t_i \leq t_j$ if $i < j$.

Rather than making an explicit assumption on how the reward function varies over time, we consider a setting in which the degree the reward can change is upper bounded. We denote this upper bound as $B_T$. This is a mild assumption on the temporal variation, and allows the reward to vary drastically at any point in time over all $T-1$ steps in the training data. We formalise this in \Cref{assumption: variation budget} (\Cref{sec: theoretical analysis}), and use it to show that the convergence of NS-DPO depends upon $B_T$. To learn given this drift, we employ an \emph{exponentially weighted maximum likelihood estimator} \citep{faury2021regret, russac2019weighted, wang2023revisiting}, where the datapoints are re-weighted such that losses incurred at the most recent datapoints are prioritised.\looseness=-1

To learn a suitable reward model in this setting, we define the reward at time step $T$ as $r(x,a,T) \in \mathcal{R}$, where $\mathcal{R}$ is the space of reward values. We estimate the reward function at timestep $T$, by maximising the exponentially weighted negative log-likelihood of the Dynamic Bradley-Terry model:\looseness=-1
\begin{align}
    &\mathcal{L}_{DBT}(r) \nonumber \\
    &= \sum_{(x_i, a_i, a'_i, t_i) \in \mathcal{D}} -\gamma^{T-t_i-1} \log \sigma \left(h_r(x_i, a_i, a_i', T)\right).\label{eq: dbt reward estimator}
\end{align}
In \Cref{eq: dbt reward estimator}, $\gamma \in (0,1)$ controls the rate at which older datapoints are discounted. The loss recovers the stationary Bradley-Terry model as $\gamma \rightarrow 1$. We show in \Cref{theorem: regret bound - offline - uniform} that the optimal value of $\gamma$ can be obtained when we know $B_T$, the details of which we provide in \Cref{appendix: complexity analysis - regret - uniform}. 

\begin{remark}
    We do not assume any knowledge of which preferences change and which remain fixed as part of the total preference drift. As a result, we down-weight \emph{all} older points in the dataset with the exponential scaling term, $\gamma$.\looseness=-1
\end{remark}

\textbf{Offline Non-Stationary Direct Preference Optimization.} The derivation of NS-DPO follows as previously shown in \Cref{sec:prelims} for the stationary case. We first define the RLHF objective at timestep $T$ as\looseness=-1
\begin{equation} \label{eq: ns rlhf objective}
    \J_T(\pi) = \Eb_{x \sim \X, a \sim \pi}\Big[r(x, a, T) - \tau \DKL[\pi(\cdot | x)\| \piref(\cdot | x)]\Big],
\end{equation}
where we are interested in maximising the reward function $r(x, a, T)$ that reflects human preferences in the present (i.e., the current time step). We note the prompt distribution $\X$ and the reference model $\piref$ do not vary with time. As we consider the reward model at $T$, we derive an implicit reward of the same form as \Cref{eq: stationary implicit reward}. This relates the optimal policy and reward function of \Cref{eq: ns rlhf objective} as
\begin{equation}\label{eq: ns implicit reward}
r(x,a,T) = \tau \log \frac{\pi^*_T(a|x)}{\piref(a|x)} + \tau \log Z^*_T(x),
\end{equation}
where $\pi^*_T$ is the optimal policy that optimises \Cref{eq: ns rlhf objective} and $Z^*_T$ denotes the normalisation constant of $\pi^*_T$. We then parameterise the policy $\pi$ in \Cref{eq: ns rlhf objective} using the parameter $\theta_T$, which enables expressing the implicit reward with respect to the parameter as  
\begin{equation}\label{eq: policy parameterisation}
r_{\theta_T}(x,a,T) = \tau \log \frac{\pi_{\theta_T}(a|x)}{\piref(a|x)} + \tau \log Z_{\theta_T}(x),
\end{equation}
where $Z_{\theta_T}$ denotes the normalisation constant of $\pi_{\theta_T}$. We apply \Cref{eq: policy parameterisation} into the exponentially weighted negative log likelihood in \Cref{eq: dbt reward estimator} to derive the NS-DPO objective
\begin{align} 
    &\Lns(\theta_T) \nonumber \\
    &= \sum_{(x_i, a_i, a'_i, t_i) \in \mathcal{D}} -\gamma^{T-t_i-1} \log \sigma \left( \tau h_{\pi_{\theta_T}}(x_i, a_i, a_i') \right).\label{eq: NS-DPO Loss}
\end{align}

\section{Theoretical Analysis of Offline Non-stationary DPO} 
\label{sec: theoretical analysis}

In this section, we analyse the performance of NS-DPO in the offline setting. We assume the use of log-linear policies, and present how the preference drift affects the estimation error and regret bound of the algorithm. We provide the sample complexity of the algorithm, which recovers $O(n^{-1/2})$ when the preferences are stationary. See \Cref{appendix: offline learning analysis} for further details.\looseness=-1 

\textbf{Policy Class.} We use the policies parameterised by $\theta \in \Theta \subset \Rb^d$ of the following form
\begin{equation}\label{eq: policy class Pi}
    \Pi = \left\{
        \pi_\theta(a | x) = \frac{\exp(f_\theta(x, a))}{\sum_{a' \in \A}\exp(f_\theta(x, a'))}
    \right\},
\end{equation}
where $f_\theta(x, a) \in \Rb$ is a differentiable function. For our analysis, we consider the case of log-linear policies where $f_\theta$ is linear: $f_\theta(x, a) = \phi(x, a)^\intercal \theta$, and the feature map $\phi(x, a)$ is a $d$-dimensional vector. This is motivated by the reward model introduced in \cite{ziegler2019fine} where the last hidden layer of the LLM is used as the feature embedding function $\phi(x,a)$.

\textbf{Loss Function with $\ell_2$ regulariser.} For the analysis of log-linear policies, we regularise the NS-DPO loss with squared $\ell_2$-norm of $\theta$, $\tau^2$ and a non-linearity coefficient $\cs$ (explained in \Cref{appendix: offline learning analysis}): \looseness=-1
\begin{align}
    \Lns_{\texttt{reg}}(\theta) &= \frac{1}{n}\Lns(\theta) + \frac{\lambda \cs \tau^2}{2} \|\theta\|^2~. \label{eq: NS-DPO loss - offline analysis}
\end{align}
\textbf{Performance measure and Optimal Policy.} Let $\thetatilde_T \in \Theta$ denote the parameter that minimises the (regularised) NS-DPO loss defined in \Cref{eq: NS-DPO loss - offline analysis}. We assess the performance of the policy $\pi_{\thetatilde_T}$, using the difference of non-stationary RLHF objectives between $\pi_{\thetatilde_T}$ and $\pi^*_T$ in \Cref{eq: ns rlhf objective}:\looseness=-1
\begin{align}
    \Roff &= \J_T(\pi^*_T) - \J_T(\pi_{\thetatilde_T}) \nonumber \\
    &= \Eb_{x \sim \X}\Big[
        \Eb_{a \sim \pi^*_T(\cdot | x)}[r(x, a, T)] \nonumber \\
    &\quad\quad\quad\quad\quad- \tau \DKL[\pi^*_T(\cdot | x)\| \piref(\cdot | x)] \nonumber \\
    &\quad\quad\quad\quad\quad- \Eb_{a' \sim \pi_{\thetatilde_T}(\cdot | x)}[r(x, a', T)] \nonumber \\
    &\quad\quad\quad\quad\quad+ \tau \DKL[\pi_{\thetatilde_T}(\cdot | x)\| \piref(\cdot | x)]
    \Big], \label{eq: expected regret}
\end{align}
where $r(\cdot, \cdot, T)$ denotes the true reward function at time $T$, and $\pi^*_T$ denotes the optimal policy against which we compare the performance of our algorithm. Given a reference policy $\piref$, the optimal policy $\pi^*_T$ is defined as the policy which optimises the RLHF objective at time step $T$ \looseness=-1
\begin{align} 
     \argmax_{\pi \in \Pi}
     \Eb_{x \sim \X, a \sim \pi}\Big[r(x, a, T) 
     -\tau \DKL[\pi(\cdot | x)\| \piref(\cdot | x)]\Big].\label{eq: def_optimal_policy}
\end{align}
Similarly, we can define the parameter $\theta^*_t$ of the optimal policy in each time step $t \in [T]$ as\looseness=-1
\begin{equation} \label{eq: def_optimal_parameter}
  \argmax_{\theta_t \in \Theta}\Eb_{x \sim \X, a \sim \pi_{\theta_t}}\Big[r(x, a, t) - \tau \DKL[\pi_{\theta_t}(\cdot | x)\| \piref(\cdot | x)]\Big].
\end{equation}

We now introduce assumptions on our setting. The preference drift is defined as the change in the true underlying parameter $\theta^*_t \in \Theta, \forall t \in [T]$ of the optimal policy $\pi^*$ at each time step. We do not constrain how the optimal parameter changes, but instead upper bound the total possible preference drift in the environment before time step $T$. This upper bound is known as the variation budget. \looseness=-1
\begin{assumption}\label{assumption: variation budget}
    (Variation Budget Bound) The parameter drift of $\theta_t^* \in \Theta$ across $T$ timesteps is upper bounded as $\sum_{t=1}^{T-1}\|\theta^*_{t+1} - \theta^*_{t} \|_{2} \leq B_T$ where $B_T > 0$ is a known constant.
\end{assumption}

We move on to general assumptions for the learning process. We bound the 2-norm of the feature and parameter spaces. \looseness=-1
\begin{assumption}\label{assumption: boundedness}
    (Boundedness) The parameters and features are bounded: 
    $\theta \in \Theta$ where $\Theta = \{\theta \in \mathbb{R}^d\ |\ \|\theta\|_2 \leq W \}$ and $\Phi = \{\phi(x, a) \in \Rb^d\ |\ \|\phi(x,a)\|_2 \leq L\}$. 
\end{assumption}
It is known that an equivalence class of reward models leads to the same preferences under the Bradley-Terry model \citep{rafailov2024direct}. This is similarly true in the case of the Dynamic Bradley-Terry model, because the implicit reward of NS-DPO, shown in \Cref{eq: ns implicit reward}, relates the reward to the policy parameters $\theta$. We thus construct the following constraint on the policy class to properly specify the problem  \citep{chowdhury2024provably}. \looseness=-1
\begin{assumption}\label{assumption: identifiability}
    (Identifiability) The optimal policy in each time step $t$ corresponds to a single parameter in $\Theta$, which satisfies \Cref{eq: def_optimal_parameter}:
        $\textbf{1}_d^\intercal \theta^*_t = 0 \ \forall t \in [T]$, where $\textbf{1}_d^\intercal \in \Rb^d$ is a vector of 1s.
\end{assumption} 

In the offline setting, our learning is constrained by the available dataset $\mathcal{D}$. A standard assumption in the offline learning literature is that of data coverage \citep{chowdhury2024provably, zhu2023principled}. The data coverage assumption ensures that the reference policy $\piref$ suitably explores the space of plausible responses of the optimal policy. We define the population covariance matrix as $\Sigma_{\pi} = \Eb[\phi(x, a)\phi(x, a)^\intercal] - \Eb[\phi(x, a)]\Eb[\phi(x, a)]^\intercal$, where the expectation is calculated over samples $x \sim \mathcal{X}, a \sim \pi(\cdot|x)$. The condition number $\kappa_\pi$ compares the coverage of the two policies $\pi$ and $\piref$
\begin{equation}
    \forall \pi \in \Pi:\ \kappa_\pi = \sup_{v \in \Rb^d}\frac{v^\intercal \Sigma_\pi v}{v^\intercal \Sigma_\piref v} = \frac{\lmax(\Sigma_\pi)}{\lmin(\Sigma_\piref)},
    \label{eq: relative condition number}
\end{equation}
while we use $\kappa = \max_{\pi} \kappa_\pi$ to denote the maximum value of $\kappa_\pi$. The definition of $\kappa_\pi$ requires that the reference policy sufficiently explores the feature space, which leads to the following assumption.
\begin{assumption}\label{assumption: feature coverage}
    (Feature Coverage) The reference policy $\piref$ satisfies $\lmin(\Sigma_\piref) > 0$.
\end{assumption}
In a time-varying setting, the quality of the dataset $\mathcal{D}$ also depends upon its temporal coverage. We use the following assumptionm which also guarantees a minimal amount of data in each time step. Having enough data in each time step is motivated by the fact that we are assuming no knowledge of the dynamics of the actual preference drift. Note that $\Theta(T)$ in the assumption is the notation for the complexity, which is different from the parameter set $\Theta$ in \Cref{assumption: boundedness}.
\vspace{-1em}
\begin{assumption}\label{assumption: temporal_distribution_new}
(Temporal Coverage) For each time step \( t \in [T-1] \), the number of datapoints in the training set is between \( \underline{m} \) and \( \bar{m} \), where \( \underline{m} > 0 \) and \( \bar{m} > \underline{m} \) are constants (i.e., \( n = \Theta(T) \)).
\end{assumption}

\begin{figure*}[h!]
    \centering
    \includegraphics[width=0.23\textwidth]{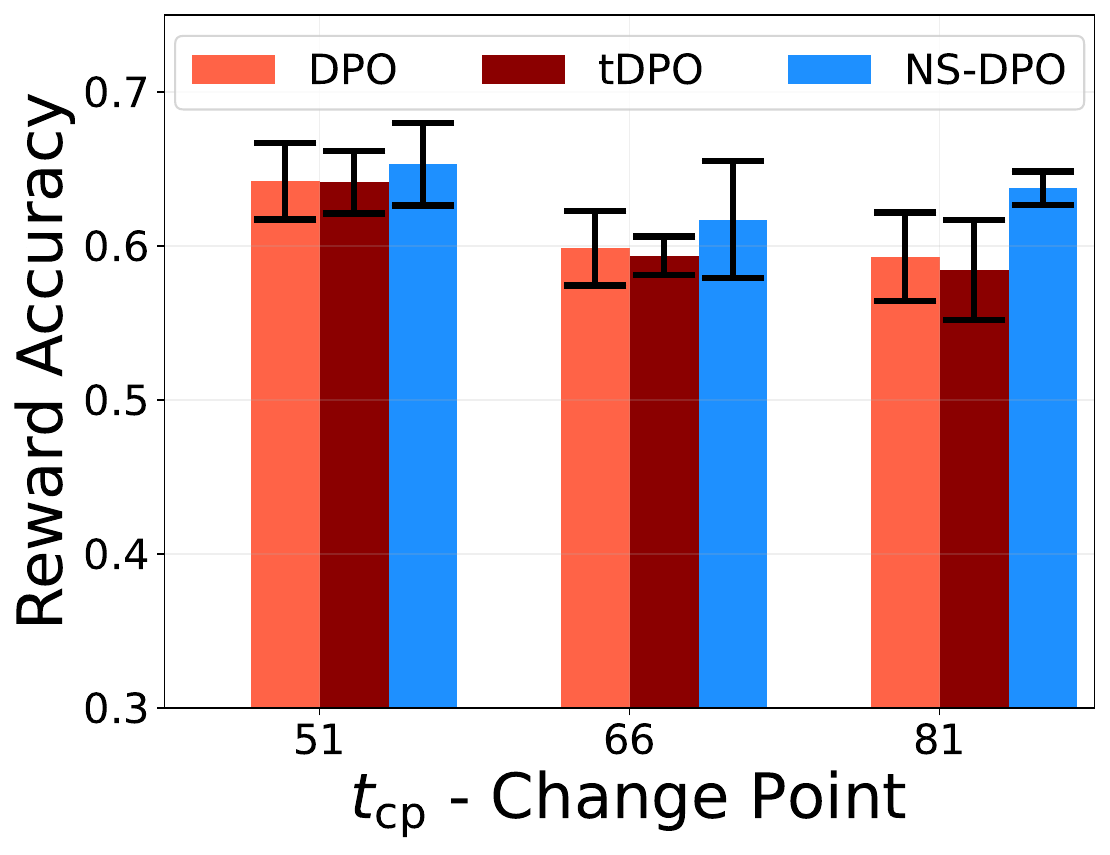}
    \includegraphics[width=0.23\textwidth]{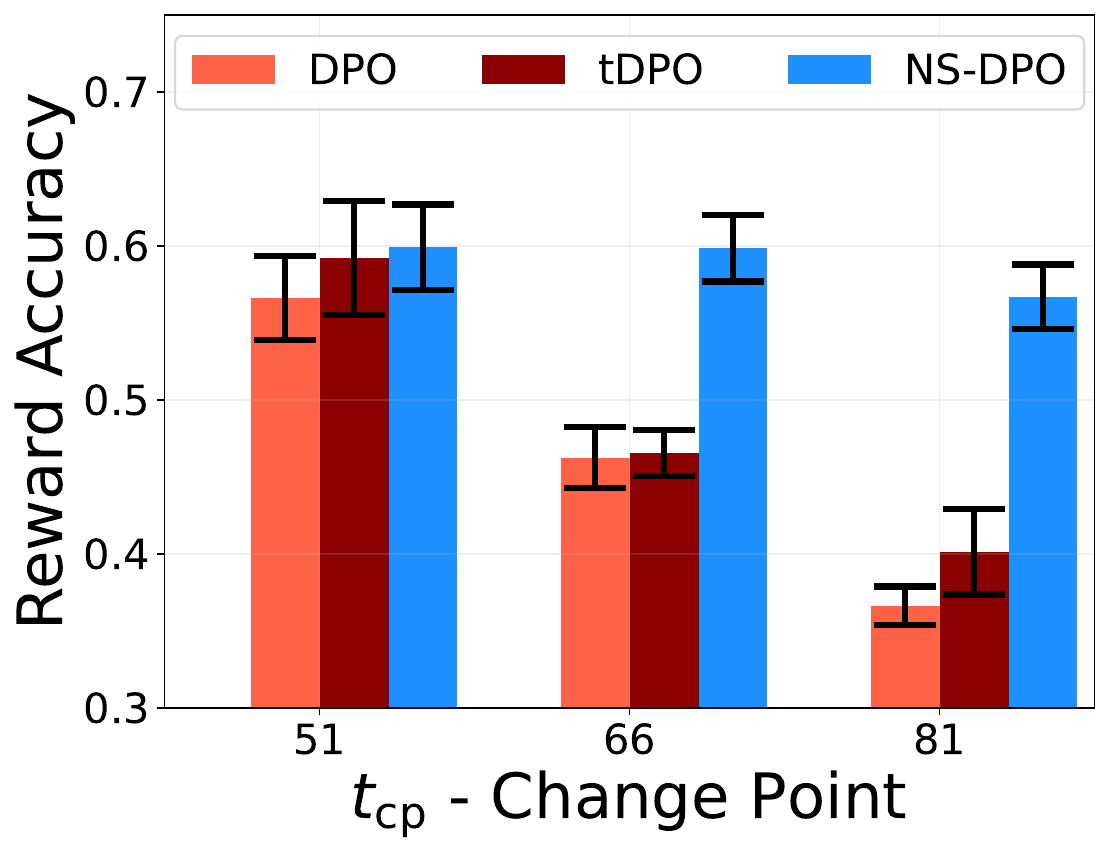}
    \includegraphics[width=0.23\textwidth]{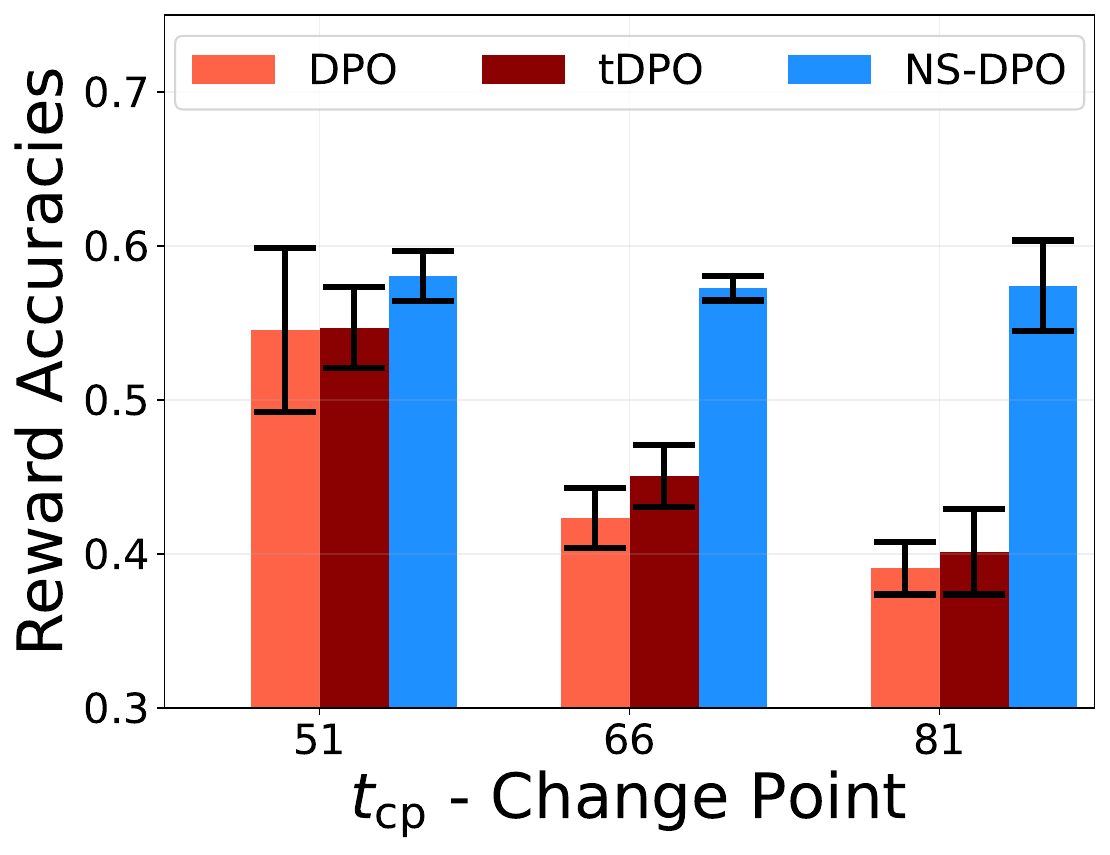}
    \includegraphics[width=0.23\textwidth]{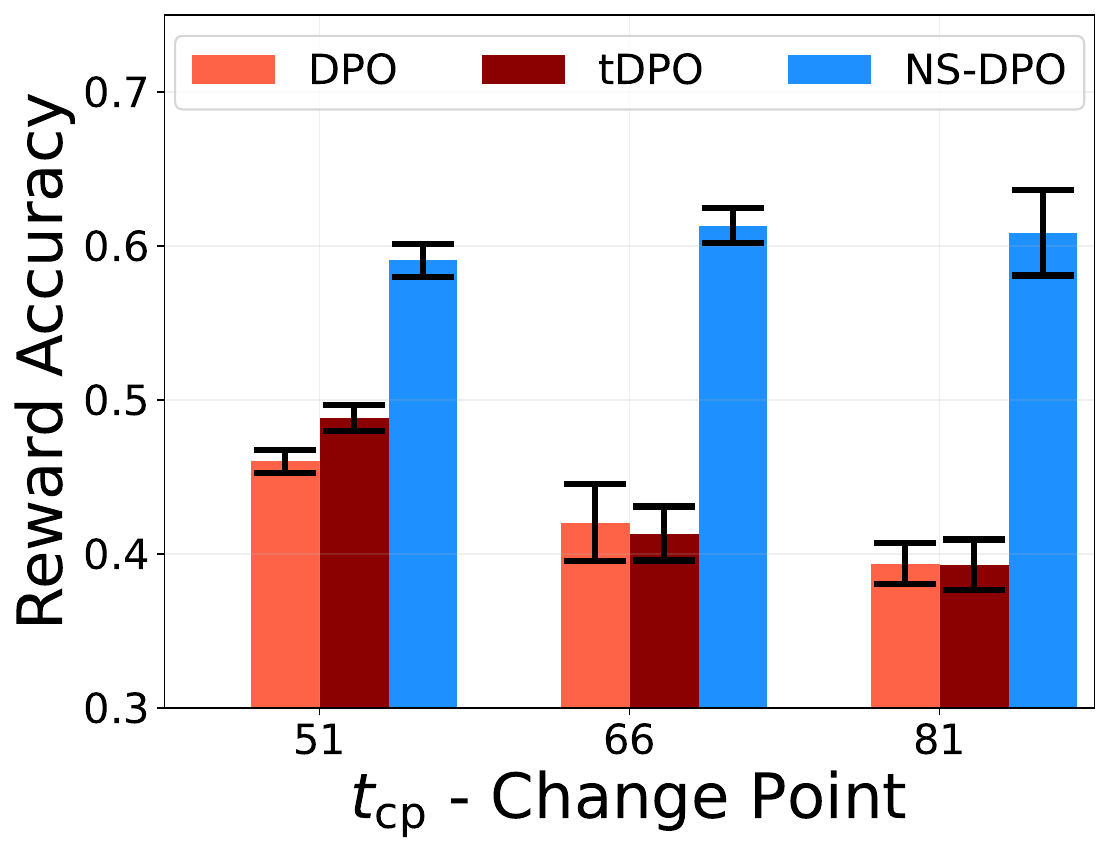}
    \captionsetup{font=footnotesize}
    % \vspace{-5pt}
    \caption{
    % [Left] Reward model shift at $\tcp=51$. [Middle] Reward model shift at $\tcp=66$. [Right] Reward model shift at $\tcp=81$. 
    Experiment results conducted on \textbf{UltraFeedback-RM} dataset with preference drift.[Left] $\rhodiff=0.7$. [Center Left] $\rhodiff=0.9$. [Center Right] $\rhodiff=0.95$. [Right] $\rhodiff=1.0$. As $\rhodiff$, the percentage of training datapoints with flipped preference increases, DPO fails to learn the preference distribution at $T=101$. Meanwhile, NS-DPO shows robust performance under various values of $\rhodiff$, maintaining reward accuracies above 50\%. As $\tcp$, the change point of the reward model happens later in time, the gap between stationary approaches and NS-DPO gets larger. The experiments are run under a reward model shift from \textsc{PairRM} to \textsc{ArmoRM}. \texttt{Llama-2-7b-chat-hf} is used, and the training dataset consists of 100 time steps.}
    \label{fig: ufb-2rm-l27b}
\end{figure*}

\subsection{Theoretical Results}
\label{sec: theoretical results}

\textbf{Estimation Error.} To bound the expected regret of the policy trained with NS-DPO, bounding the difference between the optimal and the learnt parameter is required. To analyse the parameter estimation error, we define the discounted covariance matrix of the offline dataset as
\begin{equation}
\label{eq: sigma_hat - weighted sample covariance matrix}
\sigmahat = \frac{1}{n}\sum_{i=1}^n \gamma^{T-t_i-1}\phihat_i \phihat_i^\intercal,
\end{equation}
where $\phihat_i = \phi(x_i, a_i) - \phi(x_i, a'_i)$ is also introduced for brevity. Under the assumptions from \Cref{sec: theoretical analysis}, we introduce bounds on the estimation error of the parameter $\thetatilde_T$, which minimises the NS-DPO loss in \Cref{eq: NS-DPO loss - offline analysis}, with respect to the true parameter $\theta^*_T$ and $\sigmahat$:
\begin{equation}\label{eq: def_estimation error}
    \|\theta^*_T - \thetatilde_T\|_{\sigmahat + \lambda I},
\end{equation}
where $\lambda > 0$ is introduced to guarantee the inversion of the matrix $\sigmahat + \lambda I$. The upper bound on the estimation error is shown in \Cref{theorem: estimation error - offline - uniform} and a detailed proof of the result is provided in \Cref{appendix: confidence bounds - offline}. Our analysis differs from the stationary case \citep{chowdhury2024provably}, as we consider the temporally discounted datapoints in the NS-DPO loss. This is reflected in the covariance matrix $\sigmahat$ by the inclusion of the $\gamma^{T-t_i-1}$ term, which decreases the influence of observations that happened further in the past. As part of our analysis, we separate the estimation error into a \emph{learning} term and \emph{tracking} term. This tracking term accounts for the error introduced by the non-stationary nature of the environment, depending upon $B_T$ and the choice of $\gamma$ in the algorithm to upper bound it. We outline a suitable choice for $\gamma$ below. 

\begin{theorem}\label{theorem: estimation error - offline - uniform}
    (Estimation error of $\thetatilde_T$.) Let $\delta \in (0, 1], \lambda > 0, \tau > 0$. Let $\thetahat_T$ denote the minimiser of the NS-DPO loss defined in \Cref{eq: NS-DPO loss - offline analysis}. Let $\thetatilde_T \in \Theta$ denote the parameter obtained by performing the parameter projection procedure on $\thetahat_T$. Then with probability at least $1 - \delta$:
    \begin{align}
        \|\thetatilde_T - \theta^*_T\|_{\sigmahat + \lambda I} \leq& \underbrace{2\sqrt{\lambda}W + \frac{2C_1}{\tau \cs}\sqrt{\frac{d + \log(1 / \delta)}{n}}}_{\text{learning}} \nonumber \\
        &+ \underbrace{\frac{16L \Rs \mbar}{T(1 - \gamma)^{\tfrac{3}{2}}} \sqrt{\frac{d\mbar}{n}} B_T}_{\text{tracking}}
        \label{eq: confidence bound - offline - combined main text}
    \end{align}
    where $C_1>0$ is a constant.
\end{theorem}

\begin{figure*}[t]
    \centering
    \includegraphics[width=0.32\textwidth]{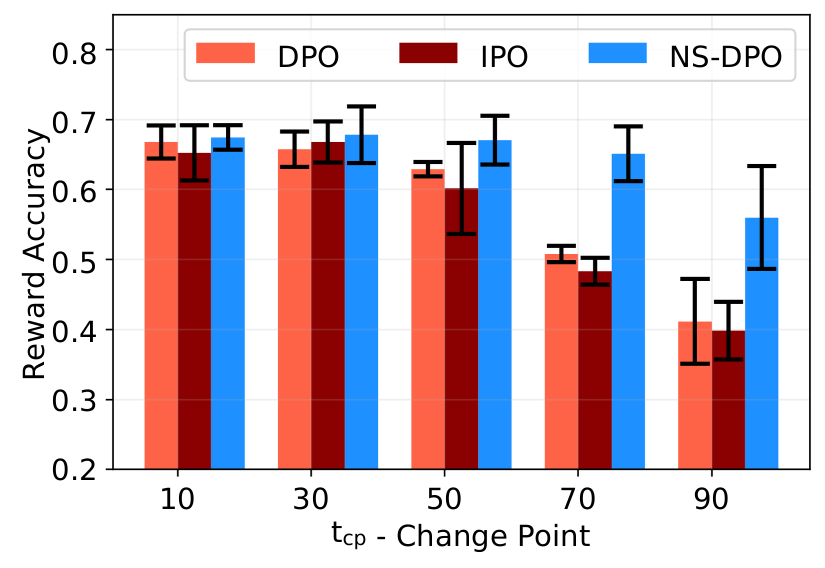}
    \includegraphics[width=0.32\textwidth]{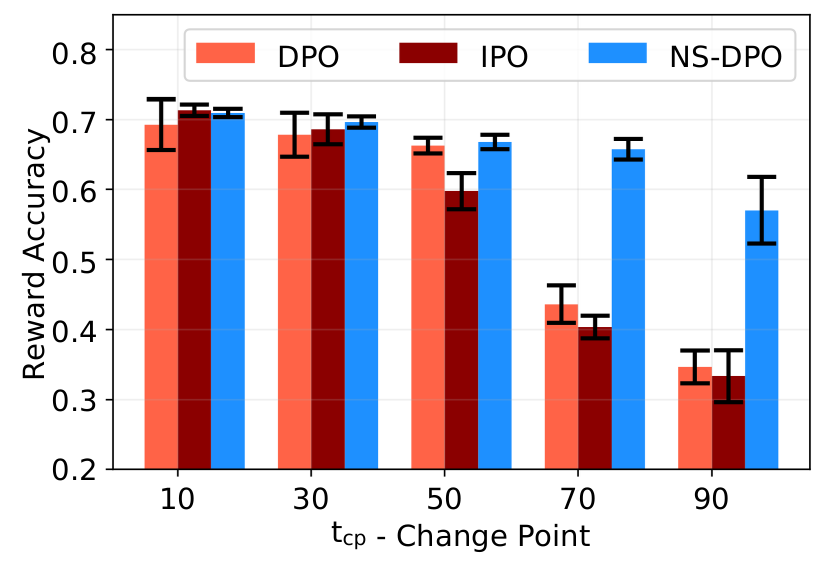}
    \includegraphics[width=0.32\textwidth]{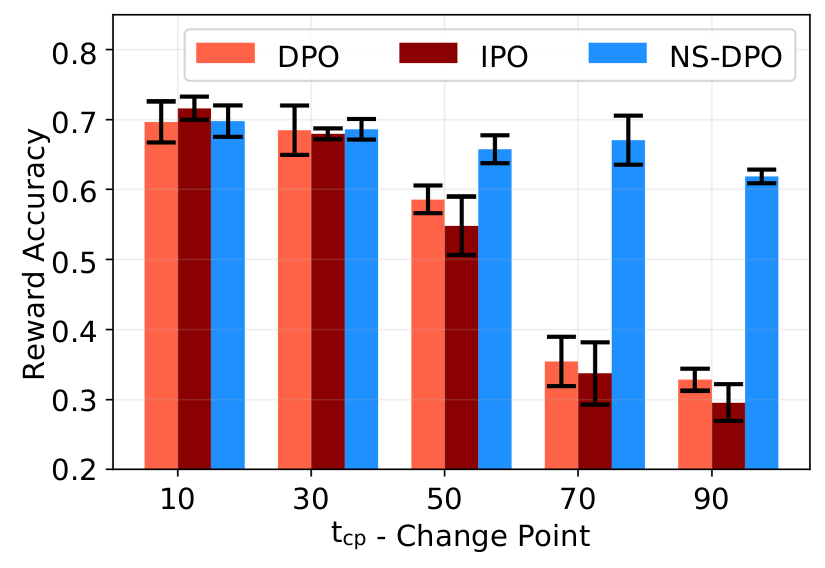}
    \captionsetup{font=footnotesize}
    \caption{NS-DPO consistently outperforms DPO and IPO as the change point, $t_{cp}$ nears the present $T=101$ for varying strengths of preference shift on the \textbf{TV-HH} dataset using the \texttt{Llama-2-7b-chat-hf} model.  [Left] $\rhodiff=0.7$. [Middle] $\rhodiff=0.8$. [Right] $\rhodiff=0.9$. We note that as the value of $\tcp$ increases, the performance difference between NS-DPO and the baselines increases. This is because as the change point moves closer to the present time step, the number of samples available from the updated preference distribution decreases. NS-DPO discounts samples with old preferences, focusing learning upon the small number of samples with up-to-date preference labels.} 
    \label{fig: NSDPO_TVHH2_changepoint_llama2}
\end{figure*}

\textbf{Expected Regret Bound.} Starting from the definition of the expected regret in \Cref{eq: expected regret}, the regret can be expressed with the estimation error in \Cref{eq: confidence bound - offline - combined main text}. We then use our results in \Cref{theorem: estimation error - offline - uniform} to complete the analysis. The details of the regret analysis are deferred to \Cref{appendix: regret bound}.
\begin{theorem}\label{theorem: regret bound - offline - uniform}
    (Regret bound of $\thetatilde_T$) Let $\delta \in (0, \tfrac{1}{2}], \tau > 0$. 
    Let $\thetatilde_T$ denote the parameter in $\Theta$ which minimises the NS-DPO loss (\Cref{eq: NS-DPO loss - offline analysis}) on an offline dataset. The following bound holds with probability at least $1 - 2\delta$ and when $\lambda \geq \conccoef$:\looseness=-1
    \begin{align}
        \Roff \leq \frac{\tau \kappa \mbar T (1 - \gamma)}{2 \mubar (1 - \gamma^{T-1})}&  \|\thetatilde_T - \theta^*_T\|_{\sigmahat + \lambda I}^2, \nonumber
    \end{align}
    where $C_1 > 0$ denotes a constant. When $\gamma = 1- \left(\tfrac{B_T}{T}\right)^{3/4}$, $\Roff$ satisfies:\looseness=-1
    \begin{align}
        \Roff = \tilde{O}\left(d\ B_T^{3/4}\ n^{-1/4}\right). \nonumber
    \end{align}
\end{theorem}

Standard offline bandits and RL algorithms assuming the stationarity of the underlying \emph{scalar-valued reward} achieve $O(n^{-1/2})$ regret \citep{wang2020statistical, zhan2023provable, qiao2024offline, cen2024value}. For stationary preference-based rewards, \cite{chowdhury2024provably} show an $O(n^{-1/4})$ regret/sub-optimality gap for DPO algorithm, whereas \cite{nika2024reward} obtain an $O(n^{-1/2})$ regret. Unlike these prior work assuming stationary preferences, NS-DPO uses the discount weight $\gamma = 1- \left(\tfrac{B_T}{T}\right)^{3/4}$ to address the non-stationarity in the dataset, which results in the regret bound above. However, our approach is general enough to capture the stationary setting, which corresponds to $B_T \rightarrow 0$. By setting $\gamma = 1 - \left(\tfrac{B_T}{T}\right)^{\alpha}$ with $0 < \alpha < \tfrac{2}{3}$, we show that the tracking term in the estimation error bound goes to zero. \Cref{corollary: regret bound - offline - stationary}, shows that the widely considered stationary setting is a special case of NS-DPO. We provide the detailed proof in \Cref{appendix: stationary case}.
\looseness=-1
\begin{corollary}\label{corollary: regret bound - offline - stationary}
    (Regret bound under stationary preferences) Let   
    $B_T \rightarrow 0$, $\delta \in (0, \tfrac{1}{2}], \tau > 0$. Let $\thetatilde_T \in \Theta$ denote the minimiser of the NS-DPO loss (\Cref{eq: NS-DPO loss - offline analysis}). Then, for $\lambda \geq \conccoef$, some constant $C_1 > 0$, $\gamma = 1 - \left(\tfrac{B_T}{T}\right)^{\alpha}$ and $0 < \alpha < 2/3$, we have with probability at least $1 - 2\delta$:\looseness=-1
    \begin{align}
        \lim_{B_T \rightarrow 0}\Roff < \frac{4\tau \kappa \mbar}{\mubar} \Bigg(
        \sqrt{\lambda}W + \frac{C_1}{\tau \cs}\sqrt{\frac{d + \log(1 / \delta)}{n}} \Bigg)^2~, \nonumber
    \end{align}
    and recover the complexity of $\Roff = O(n^{-\tfrac{1}{2}})$ under stationary preferences.
\end{corollary}

\section{Experiments}

In this section, we empirically evaluate NS-DPO's ability to learn under preference drift. We analyse how NS-DPO performs under different types of preference drift and different strengths of preference change using \texttt{Llama-2-7b-chat-hf} \citep{touvron2023llama} and \texttt{Llama-3.2-1b-it}\citep{dubey2024llama}. We provide experiments in \Cref{appendix: synthetic experiments} to further support our theoretical results in \Cref{sec: theoretical results}. We provide code\footnote{\href{https://github.com/geronest/ns-dpo}{https://github.com/geronest/ns-dpo}} for our experiments.\looseness=-1

\subsection{Experimental Setup}
\label{sec: Experimental Setup}

To test NS-DPO in an LLM setting, we create three preference datasets with known and controlled preference drift. \looseness=-1 

\textbf{Creating Non-Stationary Preference Datasets.} To create datasets with varying preference drift, we select two reward models $r_1, r_2$ that result in different preferences for the responses $a$ and $a'$. We assign each datapoint an arbitrary time across 100 timesteps $t \in [100]$ and adjust the response preference according to two main modes of preference change, sudden or gradual. For sudden preference change, we select a change point $\tcp \in [100]$ for datapoints with a time before $\tcp$ we assign preferences based on $r_1$ and for points after $\tcp$ we assign preferences based on $r_2$. For gradual preference change, we linearly interpolate the reward of each prompt response pair $(x,a)$ between $r_1$ and $r_2$ over some subset of the timesteps $T_{\text{grad}} \subset [100]$ (see \Cref{app: non-stationary preference dataset creation}).

\textbf{Preference Change Strength.} Finally, we also adjust how the strength of preference change affects the performance of NS-DPO. We introduce $\rhodiff$, which is the portion of datapoints included in the dataset whose preferences change when assigning preferences according to $r_2$ instead of $r_1$. We provide further details in \Cref{appendix: controlling the strength of preference drift}.\looseness=-1
\begin{figure*}[t]
  \begin{center}
    \begin{subfigure}[t]{0.6\linewidth}
        \includegraphics[width=0.5\textwidth]{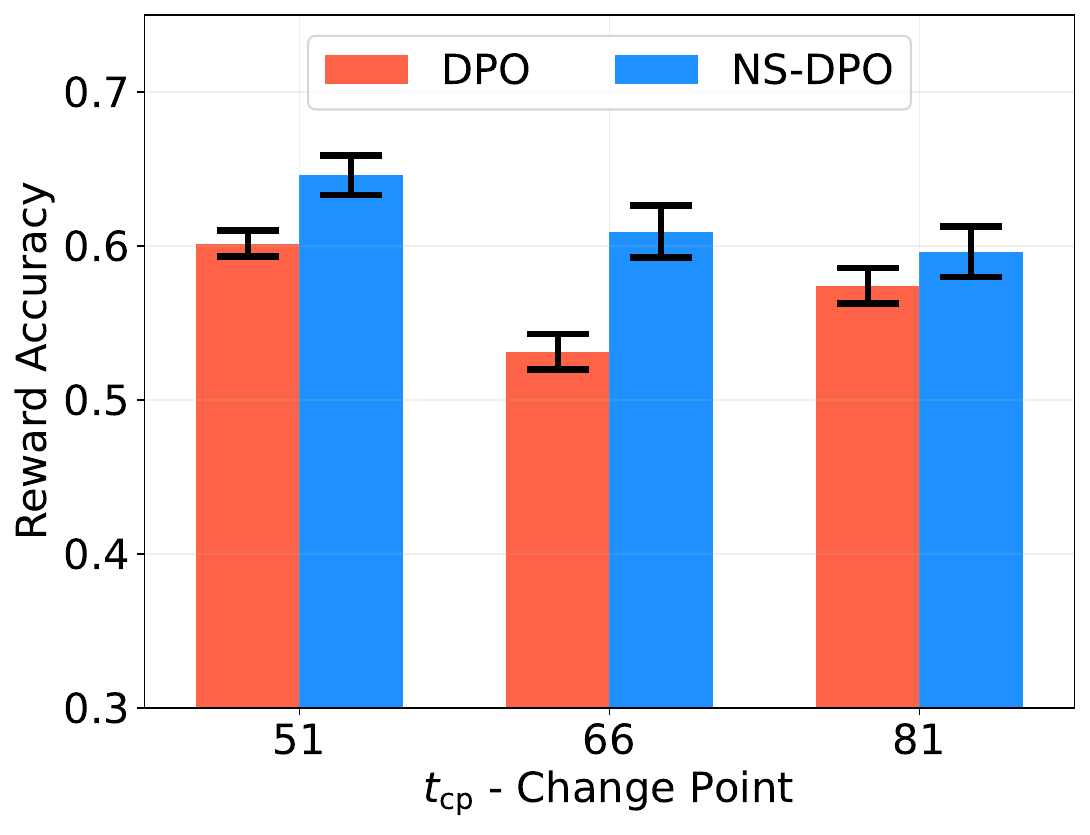}
        \includegraphics[width=0.5\textwidth]{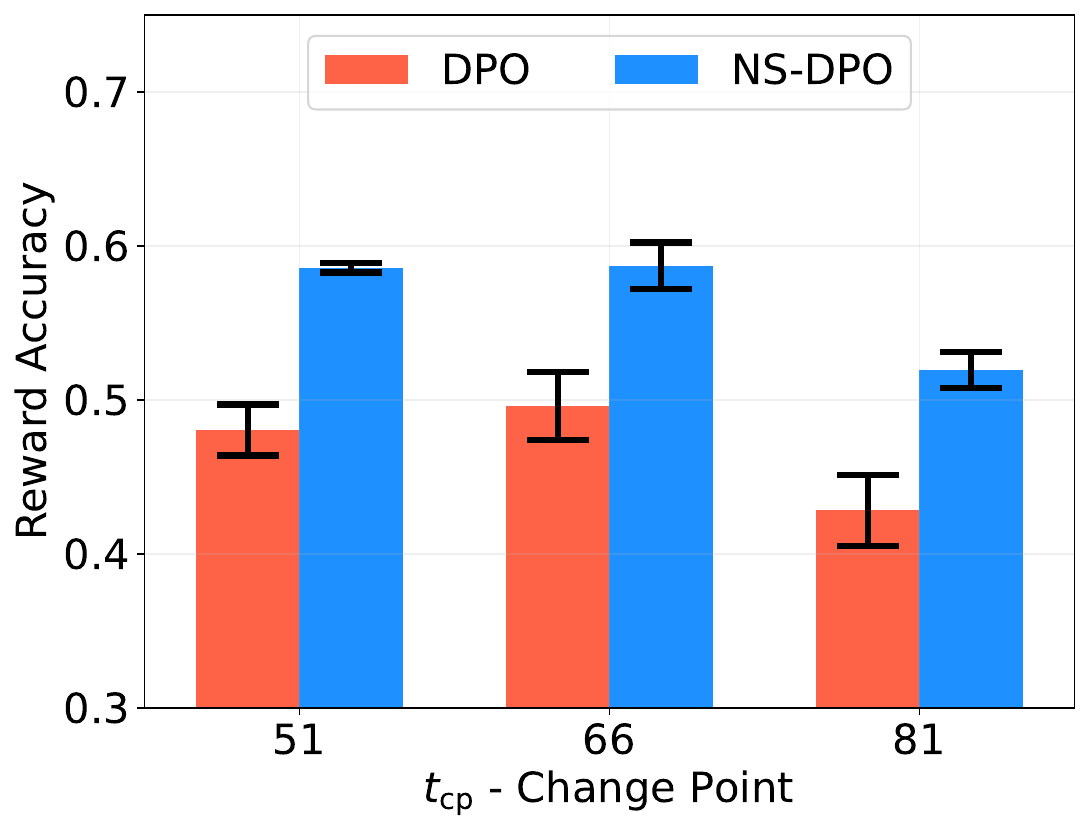}
    \end{subfigure}
    \hspace{1em}
    \begin{subfigure}[t]{0.3\linewidth}
        \includegraphics[height=0.78\linewidth]{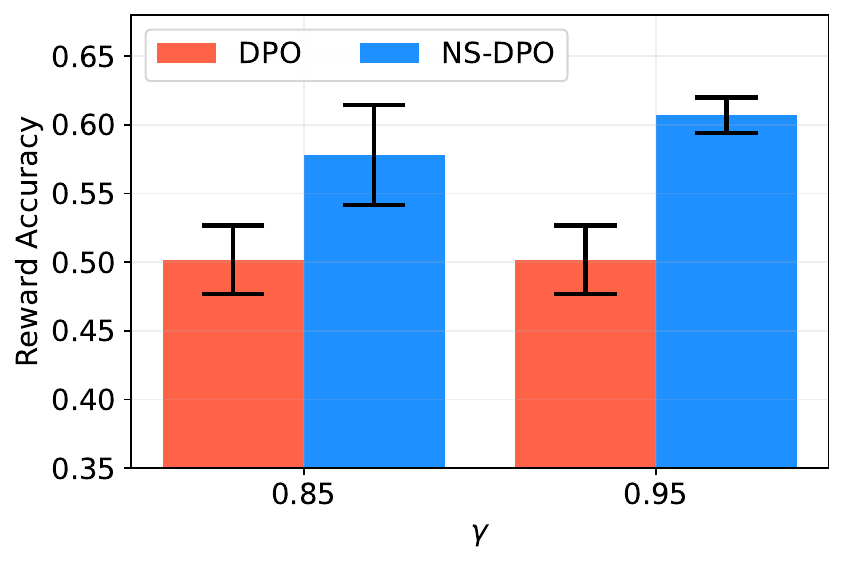}
    \end{subfigure}
    \captionsetup{font=footnotesize}
    \caption{[Left | Middle] NS-DPO outperforms DPO as the change point, $t_{cp}$ nears the present time, $T=101$, for $\rhodiff=0.7$ and $\rhodiff=1.0$ respectively on the TV-HH dataset finetuned on the \texttt{llama-3-1b-it} model. [Right] NS-DPO outperforms DPO in settings where preference drift is gradual across multiple timesteps on the TV-HH dataset.}
    \label{fig: full-finetuning+tvhh}
  \end{center}
\end{figure*}

\textbf{Datasets.} We create non-stationary preference datasets for the GlobalOpinionsQA dataset \citep{durmus2023towards} and Helpful-Harmless dataset \citep{bai2022training} using the \textit{helpsteer-helpfulness} and \textit{beavertails-is\_safe} outputs of the \textsc{ArmoRM} model. We use the Ultrafeedback dataset \citep{cui2023ultrafeedback} to create UltraFeedback-RM, which contains the preferences of the \textsc{PairRM} \citep{jiang2023llm} and \textsc{ArmoRM} \citep{wang2024interpretable} reward models. We also create UltraFeedback-LM, which uses the responses provided in the original dataset while preferences are determined based on the language model used to generate the response. We provide further details of how the non-stationary preference datasets are created in \Cref{app: non-stationary preference dataset creation}.\looseness=-1

\textbf{Language Models.} 
    We use \texttt{Llama-2-7b-chat-hf} \footnote{\href{https://huggingface.co/meta-llama/Llama-2-7b-chat-hf}{https://huggingface.co/meta-llama/Llama-2-7b-chat-hf}} and \texttt{Llama-3.2-1b-it} \footnote{\href{https://huggingface.co/meta-llama/Llama-3.2-1B-Instruct}{https://huggingface.co/meta-llama/Llama-3.2-1B-Instruct}} \citep{touvron2023llama, dubey2024llama} for both fine-tuning and the reference model. To reduce the compute demands of fine-tuning \texttt{Llama-2-7b-chat-hf}, we train LoRA weights \citep{hu2021lora} (see \Cref{sec: compute} for further details). We fine-tune all parameters of \texttt{Llama-3.2-1b-it}. 
\looseness=-1

\textbf{Evaluation Metrics.} To compare the performance of NS-DPO with the baseline algorithms, we evalute the Reward Accuracy and Win Rate. The reward accuracy is the proportion of examples for which the implicit reward of the preferred response is greater than the implicit reward of the least preferred response in the test split of a dataset. The win rate of NS-DPO versus a baseline algorithm in the TV-HH dataset is the portion of responses generated by the NS-DPO trained policy that score higher under the true reward model at time $T$, than those generated by the baseline policy. For each sample in the test set of the TV-HH dataset, we limit the maximum length of generated responses to 2048 tokens. For the UltraFeedback-LM experiment, we evaluate the performance of each algorithm by using AlpacaEval2 \citep{dubois2024length}. We report the Length-Controlled Win Rate (LCWR), which compares the responses generated by a given model to responses generated by GPT-4. 

\textbf{LLM Baselines.} We compare NS-DPO against stationary DPO and Identity Preference Optimization (IPO) \citep{azar2024general}. We also construct an In-Context Learning (ICL) algorithm referred to as tDPO, in which information about the time step is appended to the prompts of the data. All algorithms use the same supervised fine-tuned (SFT) model as the reference model. We use the SFT procedure from \cite{rafailov2024direct}, training the model on the preferred responses in the dataset. NS-DPO uses $\tau=0.1$ and $\gamma=0.95$ for fine-tuning \texttt{Llama-2-7b-chat-hf} with 2C NSGO dataset and UltraFeedback dataset. For the Time Varying Helpful-Harmless (TV-HH) dataset, we adjust the value of $\gamma$ as $\gamma = 1 - (\frac{1}{100 - \tcp})\log(100)$. For  \texttt{Llama-3.2-1b-it}, we use $\tau=1.0$ and $\gamma=0.85$.

\begin{figure}[h!]
  \begin{center}
    \includegraphics[width=0.45\textwidth]{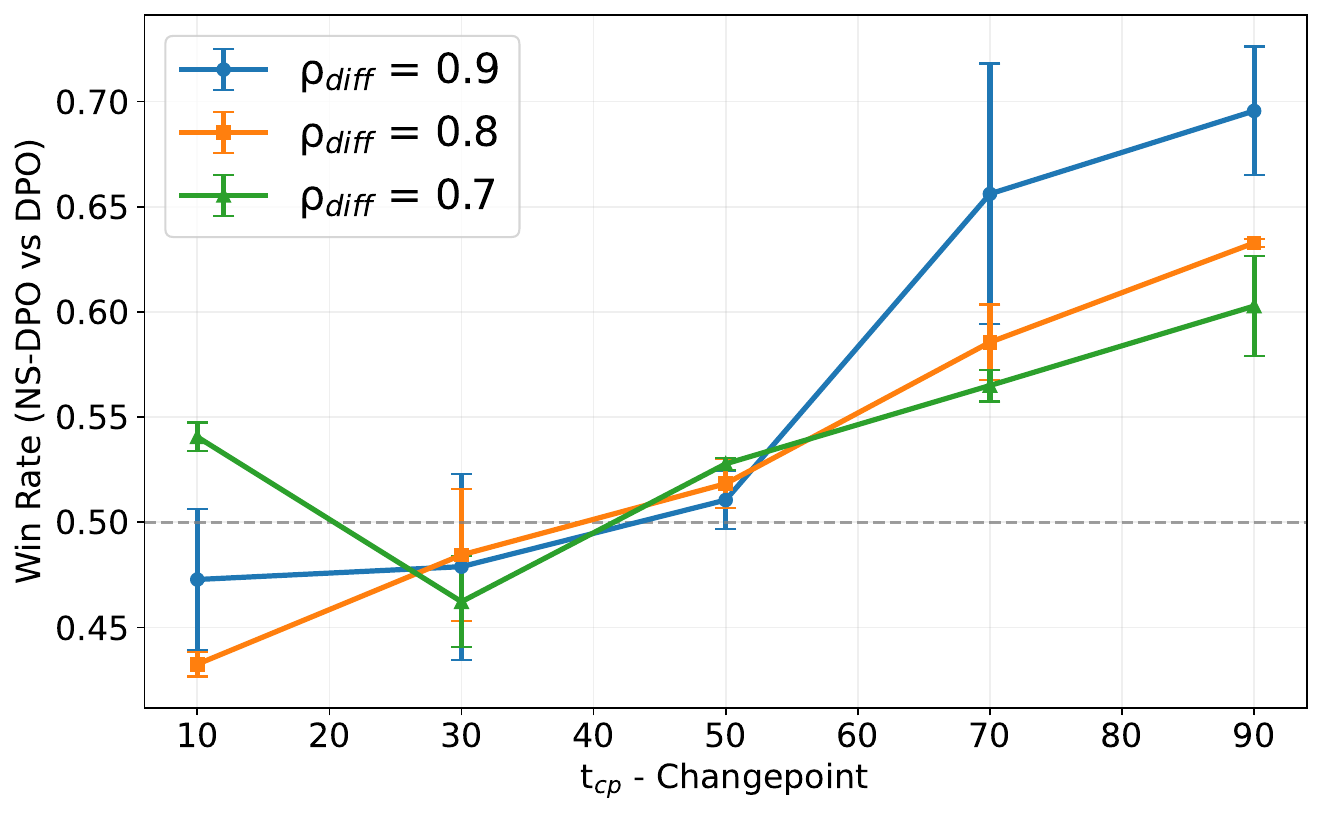}
  \end{center}
  \captionsetup{font=footnotesize}
  \caption{\textbf{NS-DPO returns more aligned responses than DPO, according to the reward model at $T=101$, when sudden preference shift occurs at later change points.} We finetune \texttt{llama-3-1b-it} on the TV-HH dataset across a range of change points and $\rhodiff$, and record the mean and std of the win rate across 600 samples from the test split over 3 runs.\looseness=-1} 
  \label{fig: llama3_1b_chpt_nsdpo_vs_dpo}
\end{figure}

\subsection{Experiment Results}

\begin{figure*}[t!]
    \centering
    \includegraphics[width=0.32\textwidth]{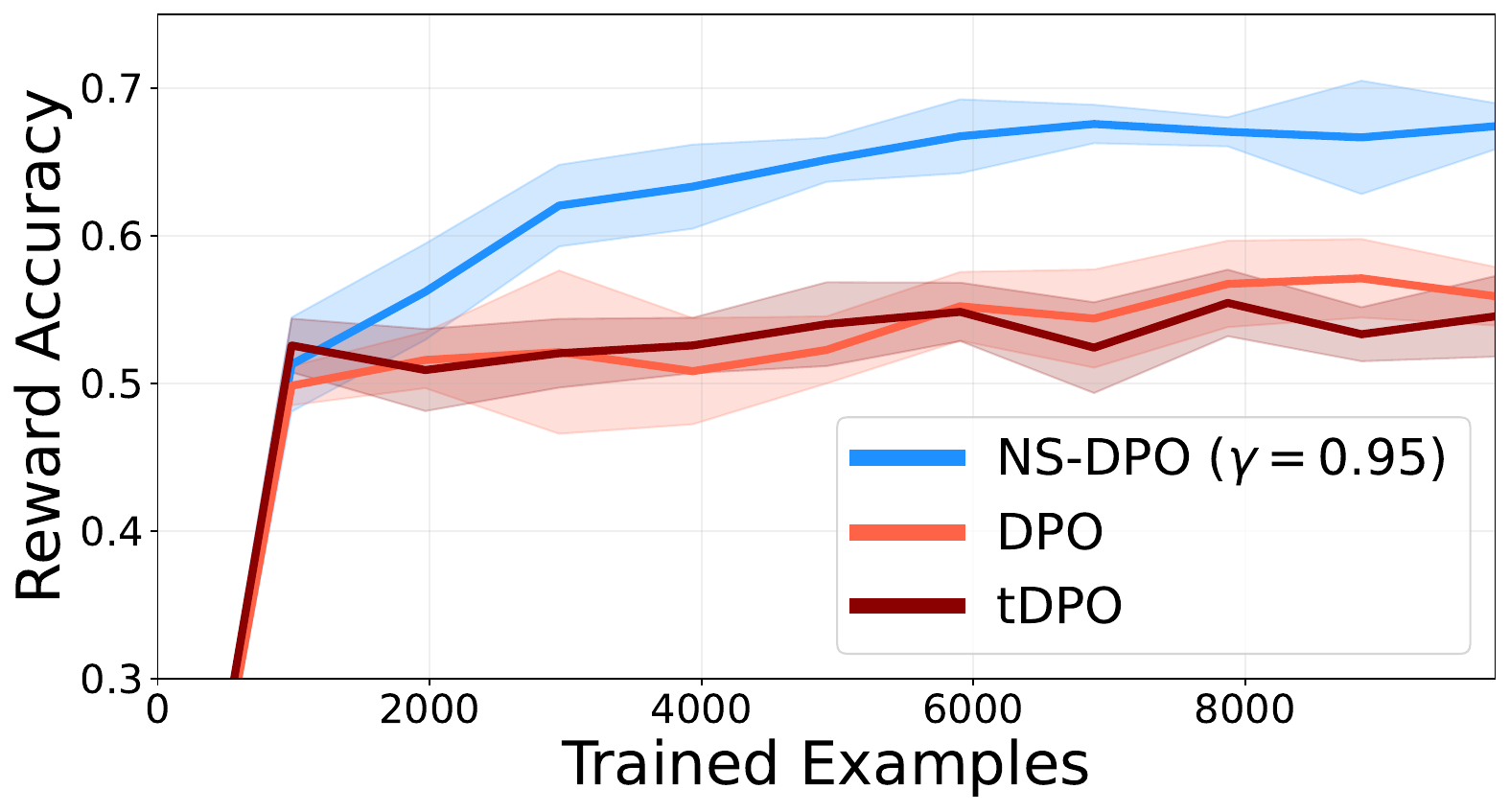}
    \includegraphics[width=0.32\textwidth]{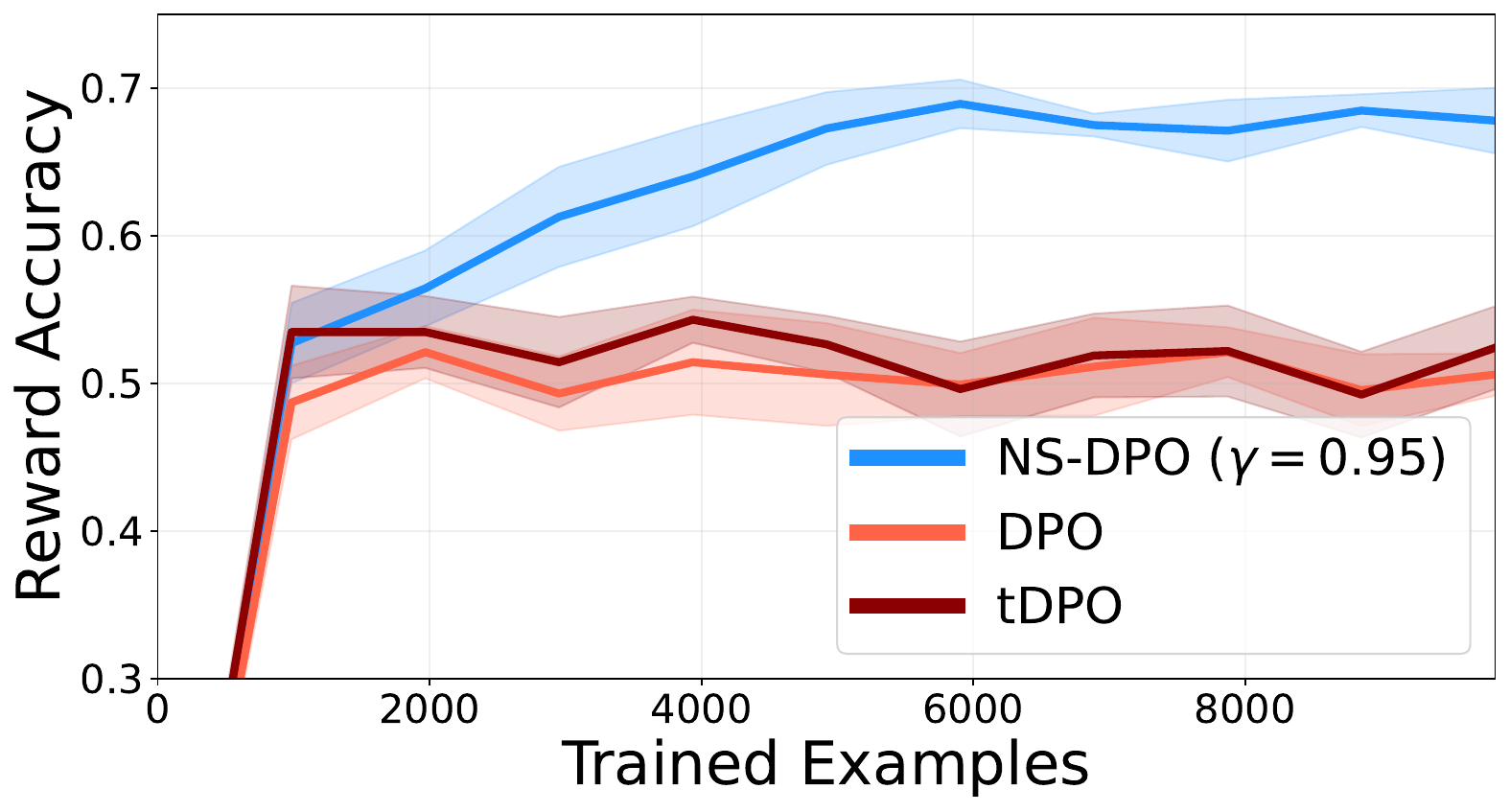}
    \includegraphics[width=0.32\textwidth]{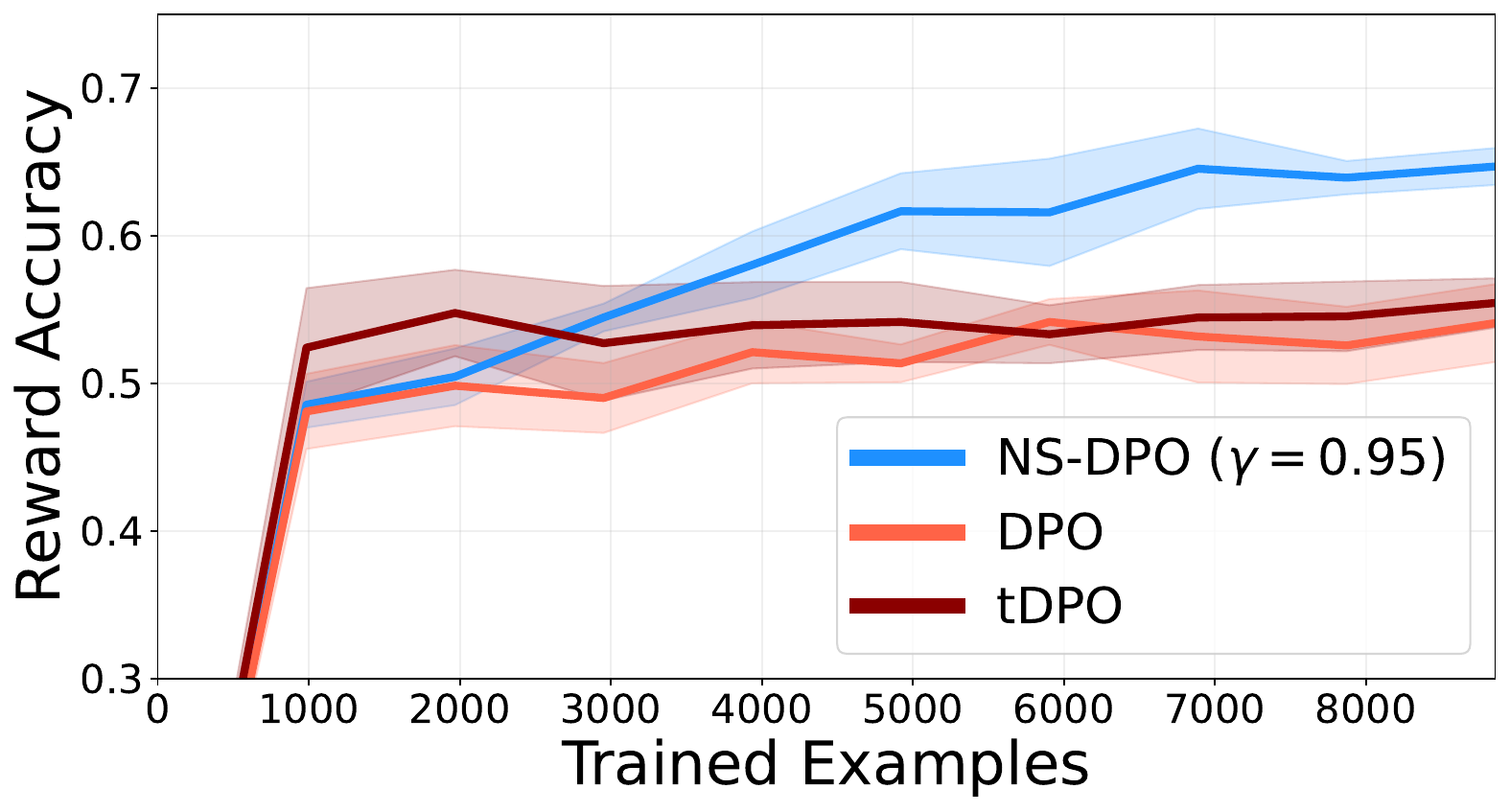}
    \captionsetup{font=footnotesize}
    % \vspace{-5pt}
    \caption{\texttt{Llama-2-7b-chat-hf} experiment results using \textbf{2C NSGO} dataset. [Left] Opinion drift from the US to Germany. [Middle] Opinion drift from the US to Japan. [Right] Opinion drift from the US to Brazil. NS-DPO stays robust to the non-stationarity present in the dataset and achieves reward accuracies above 60\%, while stationary methods show dropped reward accuracies of around 55\%. Including the time steps in the prompt (tDPO) does not help meaningfully improve the performance of stationary DPO. 
    } 
    \label{fig: experiment_nsgo2c}
\end{figure*}

\textbf{How robust and effective is NS-DPO under varying strengths of \emph{sudden} preference drift?} We investigate how different strengths of sudden preference drift affect the performance of NS-DPO. In \Cref{fig: ufb-2rm-l27b}, \Cref{fig: NSDPO_TVHH2_changepoint_llama2}, \Cref{fig: full-finetuning+tvhh} [Left | Middle], and \Cref{fig: llama3_1b_chpt_nsdpo_vs_dpo} we compare how different values of $\rhodiff$, the portion of datapoints with preferences that change, and the change point $t_{cp}$, the time at which the preference change occurs, affects the alignment of models trained with NS-DPO versus those trained with stationary preference algorithms. In \Cref{fig: llama3_1b_chpt_nsdpo_vs_dpo} we finetune \texttt{Llama-3.2-1b-it} on the TV-HH dataset and report the win rate against DPO. NS-DPO and DPO responses perform broadly the same when changes occur early $t_{cp} \in \{ 10, 30\}$, however, when the change point occurs later, NS-DPO consistently achieves a win rate of $>0.5$ beating DPO at all values of $\rhodiff$. The UltraFeedback-LM experiment in \Cref{tab:win rates} also shows that NS-DPO effectively addresses the preference drift, having consistently higher LCWR than the baselines including DPO. We observe a similar trend in \Cref{fig: ufb-2rm-l27b}, \Cref{fig: NSDPO_TVHH2_changepoint_llama2} and \Cref{fig: full-finetuning+tvhh} [Left | Middle], where the reward accuracy of NS-DPO matches that of IPO and DPO for early change points and outperforms the stationary baselines at later change points on \texttt{llama-2-7b-chat-hf} and \texttt{llama-3-1b-it} respectively. \looseness=-1

\begin{table}[t]
  \begin{center}
  \vspace{-5pt}
    
        \begin{tabular}{|c|c|c|c|c|}
        \hline
             & & \multicolumn{3}{c|}{LCWR} \\
        \hline
             $\rhodiff$ & $\tcp$ & NS-DPO & SW-DPO & DPO \\
        \hline
            0.7 & 21 & \textbf{8.93} & 6.09 & 7.29 \\
            0.7 & 51 & \textbf{8.38} & 4.93 & 7.85 \\
            0.7 & 81 & \textbf{7.85} & 4.63 & 7.17 \\
        \hline
            1.0 & 21 & \textbf{9.00} & 6.71 & 8.23  \\
            1.0 & 51 & \textbf{7.41} & 5.59 & 6.99 \\
            1.0 & 81 & \textbf{7.36} & 4.83 & 6.49 \\
        \hline
            0 & 0 & \textbf{9.12} & 8.81 & 8.81 \\
        \hline
        \end{tabular}
        \label{tab:win rates}
    
  \end{center}
  \vspace{-0.7em}
  \captionsetup{font=footnotesize}
  \caption{
      Length-Controlled Win Rates (LCWRs) of \texttt{Llama-3.2-1b-it} models, evaluated by AlpacaEval2. The models are trained with UltraFeedback-LM dataset (See \Cref{app: non-stationary preference dataset creation}). NS-DPO outperforms stationary DPO under various types of sudden preference drift, with higher preference by GPT-4 evaluator.
  }
  \label{fig: win rate}
  \vspace{-10pt}
\end{table}

Stationary algorithms treat the non-stationary preferences within the data as label noise. Thus, for early change points, the stationary algorithms see broadly correct preference labels as the majority of the dataset consists of data after the change point. As the change point occurs later, far more of the dataset is likely to have the incorrect preference, which is then learned by the stationary baseline, whilst NS-DPO focuses learning only on the later correctly labeled samples. It is also important to consider how many data points are affected by the preference change. We capture this in the $\rhodiff$ parameter. As more of the data experiences preference drift (higher values of $\rhodiff$) the more beneficial a non-stationary algorithm like NS-DPO is. We can see this in \Cref{fig: llama3_1b_chpt_nsdpo_vs_dpo} where NS-DPO achieves the strongest win rate at later change points when $\rhodiff=0.9$.\looseness=-1    

\textbf{How does NS-DPO perform under \emph{gradual} preference drifts?} Here we investigate how LLMs trained with NS-DPO perform when preference drift happens gradually over time. In \Cref{fig: full-finetuning+tvhh} [Right], we see that NS-DPO outperforms the DPO reward accuracy by over $ 10\%$ on the TV-HH dataset with gradual preference drift. We note that the performance of NS-DPO is dependent upon the value of $\gamma$ chosen, however both approaches outperform the stationary baseline. The experiment results on the 2C NSGO dataset, which also simulates a gradual drift of preferences, are given in \Cref{fig: experiment_nsgo2c}. NS-DPO shows significantly better performance compared to stationary DPO, showing a performance gap of nearly 10\% in reward accuracy. This difference is mainly caused by stationary methods failing to efficiently learn from datapoints at later time steps. tDPO, which trains the policy with time step information appended to the prompt, does not show a significant difference from stationary DPO. 

\begin{figure}[h!]
  \begin{center}
    \includegraphics[width=0.35\textwidth]{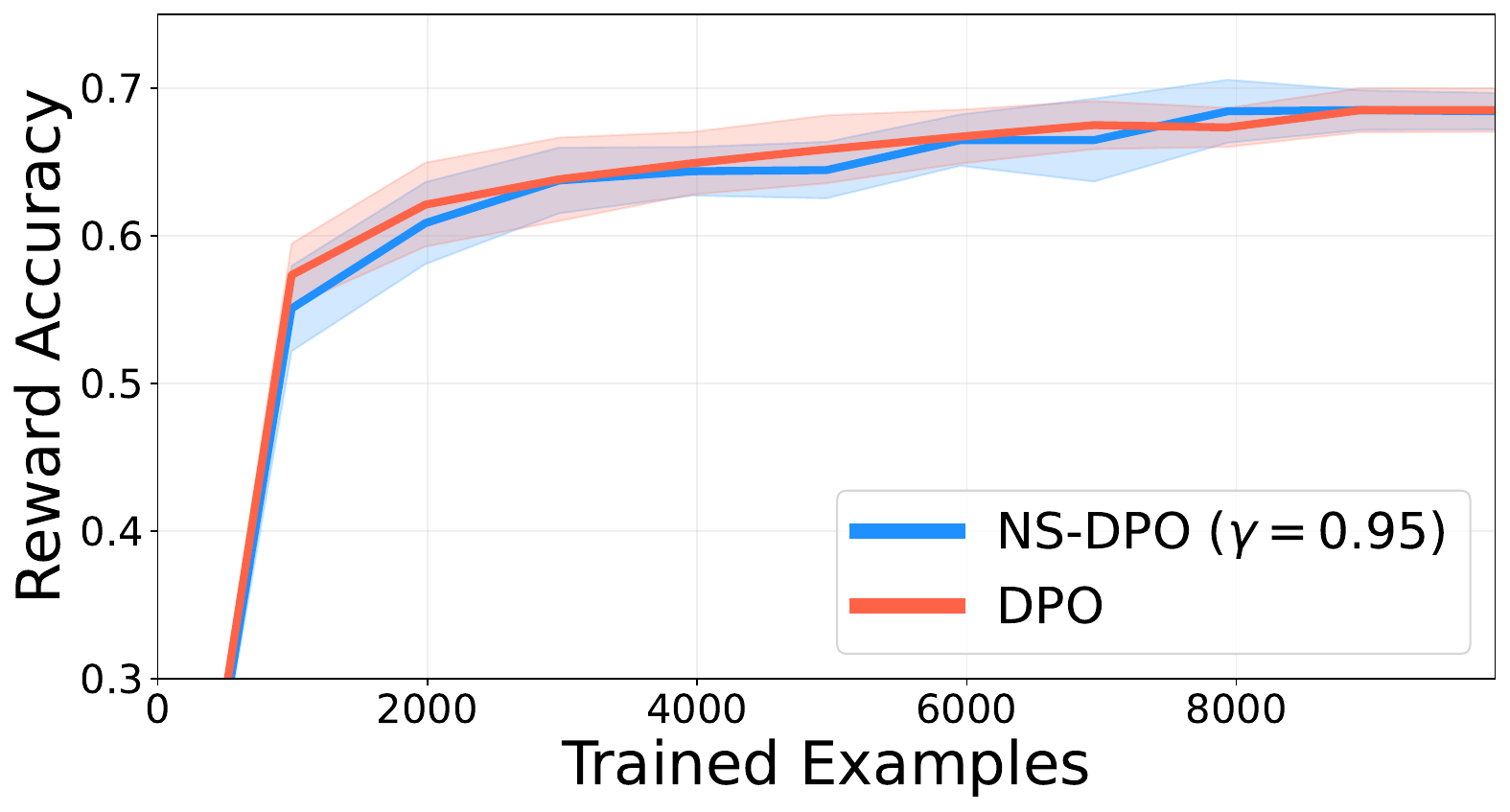}
  \end{center}
  \captionsetup{font=footnotesize}
  \caption{Training curves of NS-DPO and DPO trained with the UltraFeedback dataset without preference drift ($\tcp=0$).The trained model is initialised from \texttt{Llama-2-7b-chat-hf}. NS-DPO matches the performance of DPO even in stationary settings.} 
  \label{fig: ufb-stationary}
\end{figure}

\textbf{How does NS-DPO perform under \emph{no} preference drift?} While NS-DPO is proposed to address scenarios where preference drift happens, one natural question that follows is whether NS-DPO performs worse than stationary DPO when the dataset is stationary, containing no preference drift. In order to answer this question, we train NS-DPO and stationary DPO on the UltraFeedback-RM and UltraFeedback-RM datasets without introducing any preference drift. Note that based on the definition of $\tcp$, stationary datasets are equivalent to datasets with $\tcp=0$. As demonstrated in \Cref{fig: ufb-stationary}, the model trained with NS-DPO using the stationary UltraFeedback dataset shows almost identical reward accuracy as the model trained with stationary DPO. The AlpacaEval2 results in \Cref{tab:win rates} also support this, where NS-DPO actually shows higher LCWR than DPO even without any preference drift in the dataset.

\section{Conclusion}
In this work we propose NS-DPO, a practical and provably efficient approach for preference optimization on non-stationary offline datasets. With standard assumptions on the offline learning setting and a minimal assumption of having only a upper bound $B_T$ on the preference drift, we provide a theoretical analysis on the performance of NS-DPO in the case of log-linear policies. NS-DPO achieves a sample complexity of $O(n^{-1/4})$. We also show that as the preference drift in the environment diminishes, $B_T \rightarrow 0$, the complexity of the regret recovers $O(n^{-1/2})$, found in the stationary setting. We further support this result with a suit of empirical results on a synthetic setting. We also investigate the application of NS-DPO to LLMs, creating several non-stationary preference datasets with varied levels of preference drift, and show that NS-DPO shows superior performance to standard preference optimization algorithms and In Context Learning approaches on these datasets. Even in stationary settings, NS-DPO matches the performance of stationary algorithms. This motivates the usefulness of our approach when the existence of preference drift in a dataset is unknown, as applying NS-DPO will not hurt performance even if the preference drift is too small to matter. NS-DPO introduces only a single parameter $\gamma$ in the stationary DPO loss, which is notably easy to implement and test with. 
Our approach can be easily extended to the online setting where data is sequentially provided as time passes. NS-DPO can also be adapted to learn at a time step that is not the present by discounting both past and future preference as a function of their distance from the time step of interest. We leave these ideas for future work.\looseness=-1

\newpage

\section*{Acknowledgements}

IB was supported by the EPSRC New Investigator Award EP/X03917X/1; the Engineering and Physical Sciences Research Council EP/S021566/1; and Google Research Scholar award. WB was supported by the Engineering and Physical Sciences Research Council EP/S021566/1; and Fluidstack AI Cloud Platform Research Funding.

\section*{Impact Statement}
The development of NS-DPO provides a significant advancement in the field of preference optimization for non-stationary scenarios, addressing a critical challenge in adapting to environments with temporal preference drift. By introducing a practical and theoretically rigorous approach, NS-DPO enables robust optimization with minimal assumptions, achieving competitive sample complexity and empirical performance in both stationary and non-stationary settings.\looseness=-1

Whilst NS-DPO offers a clear approach to address temporal drift within offline dataset, it does not take the qualitative aspects of the inputs into consideration. Aligning LLMs can be a complex task and great care should be taken to avoid introducing sources of bias during training, which lead to misaligned harmful models. When using NS-DPO practitioners should be cautious of newer data that may contain biases as such biases may be amplified in training by the NS-DPO discount parameter. 

\bibliography{references}
\bibliographystyle{icml2025}

\newpage

\onecolumn
\appendix

\section*{Appendix Contents}
\label{appendix: appendix contents}

In \Cref{appendix: more related works}, we provide further related works on DPO algorithms, different alignment settings, and a discussion of works that consider time varying alignment problems. \Cref{appendix: gradient analysis} analyses the gradient of the NS-DPO objective. \Cref{appendix: further experiment details} explains the details of experiments conducted, including the creation of non-stationary datasets for LLM experiments and the behaviour of NS-DPO and SW-DPO in the synthetic setting. We provide proofs of our theoretical analysis in \Cref{appendix: offline learning analysis} step by step. In-depth derivations necessary for deriving the learning error are separately presented in \Cref{appendix: applying Bernstein's inequality}.

\section{Further Related Works}
\label{appendix: more related works}

Recent interest in the alignment of LLMs has lead to a wide variety of works. We briefly discuss further works that focus upon direct preference alignment algorithms.  

Several approaches examine preference optimisation from a game theory perspective, avoiding the implicit assumptions of the BT model. In these settings the current policy plays against previous versions to further improve performance \citep{swamy2024minimaximalist, rosset2024direct, wu2024self, yuan2024self, chen2024self, pang2024iterative, munos2023nash}. \cite{xu2023some} propose a cringe loss based objective whilst \cite{hong2024reference, pentyala2024paft, hua2024intuitive} try to combine the supervised fine-tuning and preference optimization steps. \cite{hong2024reference, hua2024intuitive} propose a single training objective to do this and \cite{pentyala2024paft} examine combining two different models trained on an SFT and direct preference objective respectively. Finally, \cite{lu2024discovering} propose a meta algorithm which uses an LLM to optimize the form of the direct preference learning objective itself.\looseness=-1

An orthogonal direction of work is the online setting \citep{qi2024online, zhang2024self, guo2024direct, xie2024exploratory}, where feedback is returned by a human labeler or superior model. \cite{khaki2024rs, liu2023statistical} adapt the offline settings using techniques such as rejection sampling to approximate an online setting. In this work we only consider the offline setting for simplicity, however the approach we propose can easily be adapted to the online setting. 
Other important directions of research include safety and robustness. \cite{dai2023safe, ramesh2024group, wu2024towards} consider robust settings where safety or group information is known at training time and \cite{dai2023safe} analyse a constrained optimization problem through the lens of safety in LLMs. Whilst these approaches look to address a wide range of settings, our work is the first to provide a solution to the case of non-stationary preferences.\looseness=-1 

\cite{carroll2024ai} consider how to correctly align LLMs under preference drift, showing several possible goals for alignment in an online setting. Whilst in the online non-stationary setting the LLM can adapt to the changing preferences of the user, our setting considers aligning the model on an offline dataset before deploying the static model to users at test time. As such our approach is most similar to the \textit{Privileged Reward} and \textit{Initial Reward} settings \cite{carroll2024ai} proposes, as we determine that the preferences exhibited in the present are the most important (\textit{Privileged Reward}) and future users will interact with a model aligned to preferences from their past (\textit{Initial Reward}).

\section{Analysis of NS-DPO Gradient}
\label{appendix: gradient analysis}

Here we analyse the gradient of the NS-DPO loss objective. The gradient of \Cref{eq: NS-DPO Loss} with respect to the model parameters $\theta$ is as follows:
\begin{equation} \label{eq: NS-DPO-grad}
    \nabla_{\theta}\Lns(\theta) = \sum_{(x_i, a_i, a'_i, t_i) \in \mathcal{D}} \underbrace{-\tau \gamma^{T-t_i-1} \sigma \left(-h_{\theta}(x_i, a_i, a'_i)\right)}_{\text{Gradient scaling}} \underbrace{\left(\nabla_{\theta} \log \pi_{\theta}(a_i | x_i) - \nabla_{\theta} \log \pi_{\theta}(a'_i | x_i)\right)}_{\text{Gradient Direction}}. 
\end{equation}
The gradient of the NS-DPO objective consists of two terms. The first term $\sigma \left(-h_{\theta}(x_i, a_i, a'_i)\right)$ scales the gradient update, which increases when the model incorrectly prefers response $a'_i$ to $a_i$ and decreases when the model correctly predicts the response preference. \textbf{NS-DPO only adjusts the scaling term} of the gradient by discounting the scaling term further when points are temporally far away from $T$. The second term, $\nabla_{\theta}\log \pi_{\theta}(a_i | x_i) - \nabla_{\theta} \log \pi_{\theta}(a'_i | x_i)$, controls the direction of the gradient update.

In the case of stationary preferences in the dataset (points whose preference does not change at any time $t_i$), the gradient of these points is still applied to the parameters $\theta$ by the NS-DPO Loss with scaling by the term $\gamma^{T-t_i-1}$. Whilst this downweights these gradients this is price of not knowing which points have changing preferences and which points have fixed preferences within our setting. When we know that there is no preference drift, we set the value of $\gamma$ to 1 to remove discounts (see \Cref{appendix: stationary case}).

\section{Further Experiment Details}
\label{appendix: further experiment details}

\subsection{Controlling the Strength of Preference Drift}
\label{appendix: controlling the strength of preference drift}
In this section, we give more details on how $\rhodiff$ is calculated, which is used to control the degree of preference drift as reward models are changed in the experiments. We first note that when $t < \tcp$, \textit{old} reward model is used to evaluate the preference of the given prompt-response pair, while we use \textit{new} reward model to evaluate datapoints with $t \geq \tcp$:
\begin{align*}
    r(x, a, t) = 
    \begin{cases}
        r^\mathrm{old}(x, a), & \text{if } t < \tcp\\
        r^\mathrm{new}(x, a), & \text{if } t \geq \tcp.
    \end{cases}
\end{align*}
We then use $o_i^\mathrm{old}$ and $o_i^\mathrm{new}$ to denote the preference given by old and new reward model respectively, on the response pairs $(a_i, a'_i)$ of prompt $x_i$:
\begin{align*}
    o_i^\mathrm{old} \sim \sigma(r^\mathrm{old}(x_i, a_i) - r^\mathrm{old}(x_i, a'_i)), \\
    o_i^\mathrm{new} \sim \sigma(r^\mathrm{new}(x_i, a_i) - r^\mathrm{new}(x_i, a'_i)).
\end{align*}
Using $o_i^\mathrm{old}$ and $o_i^\mathrm{new}$, we calculate the portion of datapoints whose preferences differ between the old and new reward models:
\begin{align}
    \rhodiff = \frac{1}{n}\sum_i^n \mathds{1}(o_i^\mathrm{old} \neq o_i^\mathrm{new}). \label{eq: def_rhodiff}
\end{align}
If the value of $\rhodiff$ is large, it means that the preference drift from the old reward model to the new reward model is happening stronger in the dataset. When $\tcp$ is fixed for the dataset, which means that the number of datapoints from each reward model is fixed, datasets with higher $\rhodiff$ will result in worse performance of the algorithms. This is because more datapoints evaluated with the old reward model will have conflicting preference with the new reward model, causing harm to learning the true preference.

\subsection{Non-Stationary Preference Dataset Creation}
\label{app: non-stationary preference dataset creation}

\begin{wrapfigure}{r}{0.30\textwidth}
  \begin{center}
  \vspace{-2.0em}
    \includegraphics[width=0.29\textwidth]{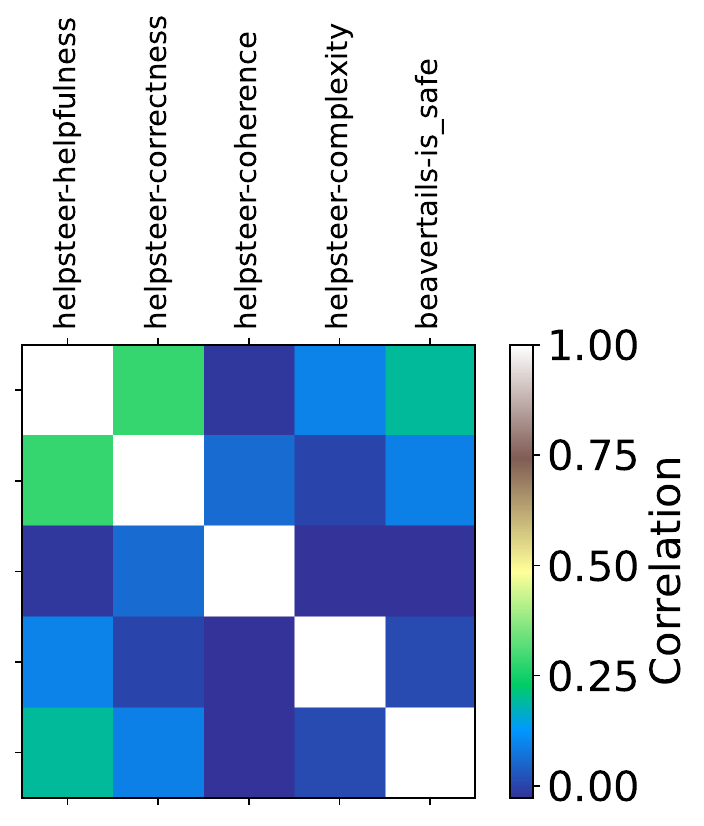}
  \end{center}
  \vspace{-1em}
  \captionsetup{font=footnotesize}
  \caption{The correlation of different preference labels generated by rewards from the \textsc{ArmoRM} reward model on the Helpful Harmless  \textit{harmless-base} dataset \citep{bai2022training}. We observed that concepts such as safety and helpfulness have more correlated preferences, whilst the \textit{helpsteer-coherence} reward model is un-correlated with the other models we analysed.\looseness=-1}
  \label{fig: tvhh2 armorm reward cov}
  \vspace{-2em}
\end{wrapfigure}

\textbf{1) NSGO Datasets.} We modify the GlobalOpinionQA dataset\footnote{\href{https://huggingface.co/datasets/Anthropic/llm_global_opinions}{https://huggingface.co/datasets/Anthropic/llm\_global\_opinions}} \citep{durmus2023towards} to create a time varying dataset. GlobalOpinionQA consists of questions regarding global issues, different responses, and preferences from several countries represented as a probability vector. We copy the questions and responses to create multiple time steps $t \in [100]$. We then vary the preferences with time by linearly interpolating between the preferences of two different countries. This simulates gradual preference drifts that can be caused by demographic shift or a series of external events. We generate preference drift using three pairs of countries. In each pair the starting country is the US, and the ending country is either Brazil, Japan or Germany. The preferences at the first and last time step correspond to either country in the pair. The last time step is held out as a test dataset and treated as the current time $T=101$. We divide the prompt-response pairs so that training and test data do not share any prompts.

\textbf{2) UltraFeedback-RM Datasets.} Using the prompts and response candidates of UltraFeedback\footnote{We modify the \href{https://huggingface.co/datasets/HuggingFaceH4/ultrafeedback_binarized}{binarized version of UltraFeedback}.} \citep{cui2023ultrafeedback}, we obtain preferences from two different reward models, \textsc{PairRM}\footnote{\href{https://huggingface.co/llm-blender/PairRM}{https://huggingface.co/llm-blender/PairRM}}\citep{jiang2023llm} and \textsc{ArmoRM}\footnote{\href{https://huggingface.co/RLHFlow/ArmoRM-Llama3-8B-v0.1}{https://huggingface.co/RLHFlow/ArmoRM-Llama3-8B-v0.1}} \citep{wang2024interpretable}. The datapoints in the training set are randomly assigned to one of $t \in [100]$ time steps, and assigned preferences of \textsc{PairRM} if the time step $t$ is earlier than the change point $\tcp \in \{51, 66, 81\}$. We assign the preferences of \textsc{ArmoRM} for the datapoints with time steps $t \geq \tcp$ and datapoints in the test set with $T = 101$. To test the effect of varied degrees of preference drift, we also vary the portion of datapoints whose preferences flip as reward model changes. We denote this portion as $\rhodiff$ and use $\rhodiff \in \{0.7, 0.9, 0.95, 1.0\}$ to create both training and test data. We use 10k datapoints for training and 500 datapoints for testing.\looseness=-1

\textbf{3) UltraFeedback-LM Datasets.} Using the same UltraFeedback dataset as above, we construct another dataset with the information of language models used for generations. The datapoints in the training set are randomly assigned to one of $t \in [100]$ time steps. Among the datapoints whose time step is earlier than the change point $\tcp \in \{21, 51, 81\}$, $\rhodiff \in \{0.7, 1.0\}$ of the datapoints have responses that are generated by \emph{smaller} language models as preferred responses. The other datapoints have responses generated by \texttt{gpt-4} as preferred. We use 23.3k datapoints for training. We use the generations of \texttt{starchat}, \texttt{llama-2-7b-chat}, \texttt{wizardlm-7b}, \texttt{pythia-12b}, \texttt{alpaca-7b}, \texttt{llama-2-13b-chat}, \texttt{wizardlm-13b}, \texttt{ultralm-13b} for \emph{smaller} language models in the dataset.
\looseness=-1

\textbf{4) Time Varying Helpful Harmless Datasets.} Using the \textit{harmless-base} subset of the Helpful Harmless dataset\footnote{\href{https://huggingface.co/datasets/Anthropic/hh-rlhf}{https://huggingface.co/datasets/Anthropic/hh-rlhf}}\citep{bai2022training}, we create a time varying preference dataset. To do so, we use two reward models, the \textit{helpsteer-helpfulness} and \textit{beavertails-is\_safe} outputs from the \textsc{ArmoRM} model \citep{wang2024interpretable}. \Cref{fig: tvhh2 armorm reward cov} shows that these rewards result in different preferences on the \textit{harmless-base} dataset. We then assign each datapoint in the dataset a random time value from $t \in [100]$. We construct two methods to assign preferences using the time step information: change point preference shift and gradual variation. Under the change point preference shift, datapoints are assigned preferences according to \textit{helpsteer-helpfulness} before the change point $t_{cp}$ and \textit{beavertails-is\_safe} after the change point. Under gradual variation, we use the following reward model
\begin{equation*}
    r(x,y,t) \!=\! \begin{cases} r_{0}(x,y) & t < 33 \\
    r_{0}(x,y)\frac{(t - 33)}{33} + r_1(x,y)\big(1 - \frac{t - 33}{33}\big) & 33 \leq t < 66 \\
    r_{1}(x,y) & t \geq 66,
    \end{cases}
\end{equation*}
where $r_0$ is the \textit{helpsteer-helpfulness} reward and $r_1$ is the \textit{beavertails-is\_safe} reward. We use this type of schedule for gradual change to simulate preference drifts that happens gradually over a finite time horizon. We use $15k$ points for training and $2k$ for testing. We use reward models for helpfulness and safety, as these are both desired properties of an LLM but often result in differing preferences; for example, rewarding helpfulness can often lead to unsafe outputs when an LLM is asked a dubious question, like how to best rob a store.\looseness=-1

\subsection{The Two Countries (2C) Non-Stationary Global Opinions Dataset}
\label{app: 2c nsgo dataset creation}

To test NS-DPO, we create a synthetic non-stationary dataset in which the temporal trends are known. To do this, we use the GlobalOpinionsQA dataset \citep{durmus2023towards}. We preprocess the dataset in three major ways. \looseness=-1

\textbf{Binary Preferences.} We convert the dataset to a dataset of binary preferences. For each set of prompt and responses, we create a row for each possible combination of prompt and binary response pairs. We calculate the preference probability for these response pairs as follows. Assuming the non-binary responses follow a Plackett-Luce preference framework, we can find the reward associated with responses (up to an additive constant) by taking the log of the preference probability. We can then take the sigmoid of these responses to find a normalised binary preference.

\textbf{Country Filter.} We filter the dataset down to the following countries: Nigeria, Egypt, India, China, Japan, Germany, France, Spain, United States, Canada, Brazil, Argentina, Australia and New Zealand.

\textbf{Country Level Prompts.} We filter the dataset such that each row of the dataset is the prompt, response, preference probability of a single country.

After the preprocessing, we copy the dataset and assign a different timestep to each unique instance of (prompt, response, preference). We simulate the drift in preferences by using preference probabilities of two countries, shifting from one to another over time. Out of $100$ time steps in the training dataset, the first 33 time steps consisted of preference probabilities from the US. Preference labels sampled from the last 33 time steps are from probabilities of the target country. We use Germany, Japan and Brazil as target countries, creating three different datasets. In the intermediate 33 time steps, preference labels are sampled from interpolated probabilities between these two countries. To introduce sufficient shift in preferences, we selected responses in which probabilities for the same response from two countries differed at least by $0.2$. We subsample prompt-response pairs down to 10,000 datapoints, allowing each time step to consist of different prompts and responses. For evaluation, we use prompts and response candidates that are not present in the training data.

\subsection{Compute Resources Uses}
\label{sec: compute}
To run the LLM experiments, we use A100 GPUs with 40GB VRAM. The synthetic experiments are run locally on a laptop without using GPUs.  

\subsection{Synthetic Experiments}
\label{appendix: synthetic experiments}

To analyse the performance of NS-DPO in the log-linear policy class, we construct a synthetic environment with a known feature space and preference drift. We use the feature space from \citep{Li2023PolicyOI}, where $x \in \X = [0, 1]^{d_x}$, $a \in \A = [n_a]$ and $\phi(x,a)$ is computed as
\begin{equation}\label{eq: feature space}
    \phi(x, a) = \Bigg[(a+1) \cdot \cos(x_0 \cdot \pi), \frac{1}{a + 1} \cdot \sin(x_0 \cdot \pi), \cdots, (a+1) \cdot \cos(x_{d_x-1} \cdot \pi), \frac{1}{a + 1} \cdot \sin(x_{d_x-1} \cdot \pi) \Bigg].
\end{equation}
The dimensions of the feature space and the policy parameter are both $2 \cdot d_x$. We use $d_x = 4, d_\theta = 8, |\A| = 16$ for all synthetic experiments.\looseness=-1

\textbf{Non-stationary Dataset.} To construct a dataset $\mathcal{D} = \{x, a, a', t\}_{i=1}^{n}$, we randomly sample $x \sim X$ and $a_1, a_2 \sim \A$. We assign 20 datapoints per time step $\forall t\in[100]$. We sample 100 datapoints for evaluation at $T=101$. To introduce preference drift, we follow an approach similar to \cite{faury2021regret}. We sample the preferences over $a_1$ and $a_2$ from the class of log-linear policies given in \Cref{eq: policy class Pi}, parameterised by $\theta^*_t$. We denote preferred response as $a$ and the rejected response as $a'$. When $t \leq 33$, we set the optimal parameter as $\theta^*_t = (1, 0, 1, 0, 1, 0, 1, 0)^\intercal$. Between $34 \leq t \leq 66$, the parameter $\theta^*_t$ varies as
\begin{align}
    \theta^*_t = \left[
        \cos(\tfrac{t - 33}{33}\cdot \tfrac{\pi}{2}), \sin(\tfrac{t - 33}{33}\cdot \tfrac{\pi}{2}), \ldots, \cos(\tfrac{t - 33}{33}\cdot \tfrac{\pi}{2}), \sin(\tfrac{t - 33}{33}\cdot \tfrac{\pi}{2})
    \right]^\intercal. \label{eq: synthetic_param_shift}
\end{align}
For the remaining time steps $67 \leq t \leq 100$, we use $\theta^*_t = (0, 1, 0, 1, 0, 1, 0, 1)^\intercal$.\looseness=-1

\textbf{Algorithms for Synthetic Experiments.} We compare NS-DPO with DPO and SW-DPO in synthetic experiments. SW-DPO uses a "sliding" window to only consider points close to the current timestep $T$, which is commonly used in the non-stationary bandit literature \citep{garivier2008upper}. We test the performance of NS-DPO and SW-DPO over several values of $\gamma\in \{0.7,0.9\}$ and window size $w \in \{33, 50\}$. The regularisation coefficient is $\tau = 1.0$ for all algorithms. We normalise the scale of the gradient for each method to address the differences caused by the application of exponential weighting and sliding window. For the reference policies, we use a uniform policy, whose parameter $\thetaref \in \Rb^d$ is a zero vector. \looseness=-1 

\textbf{Evaluation Metrics.} To analyse the performance of the algorithms, we use the reward accuracy of the trained policies. The reward accuracy is computed by the portion of test response pairs with correctly estimated preferences, using the implicit rewards defined in \Cref{eq: ns implicit reward}. For each tested algorithm, we report averaged results of the experiments across 10 different random seeds.\looseness=-1

\begin{figure*}[t]
    \centering
    \includegraphics[width=0.45\textwidth]{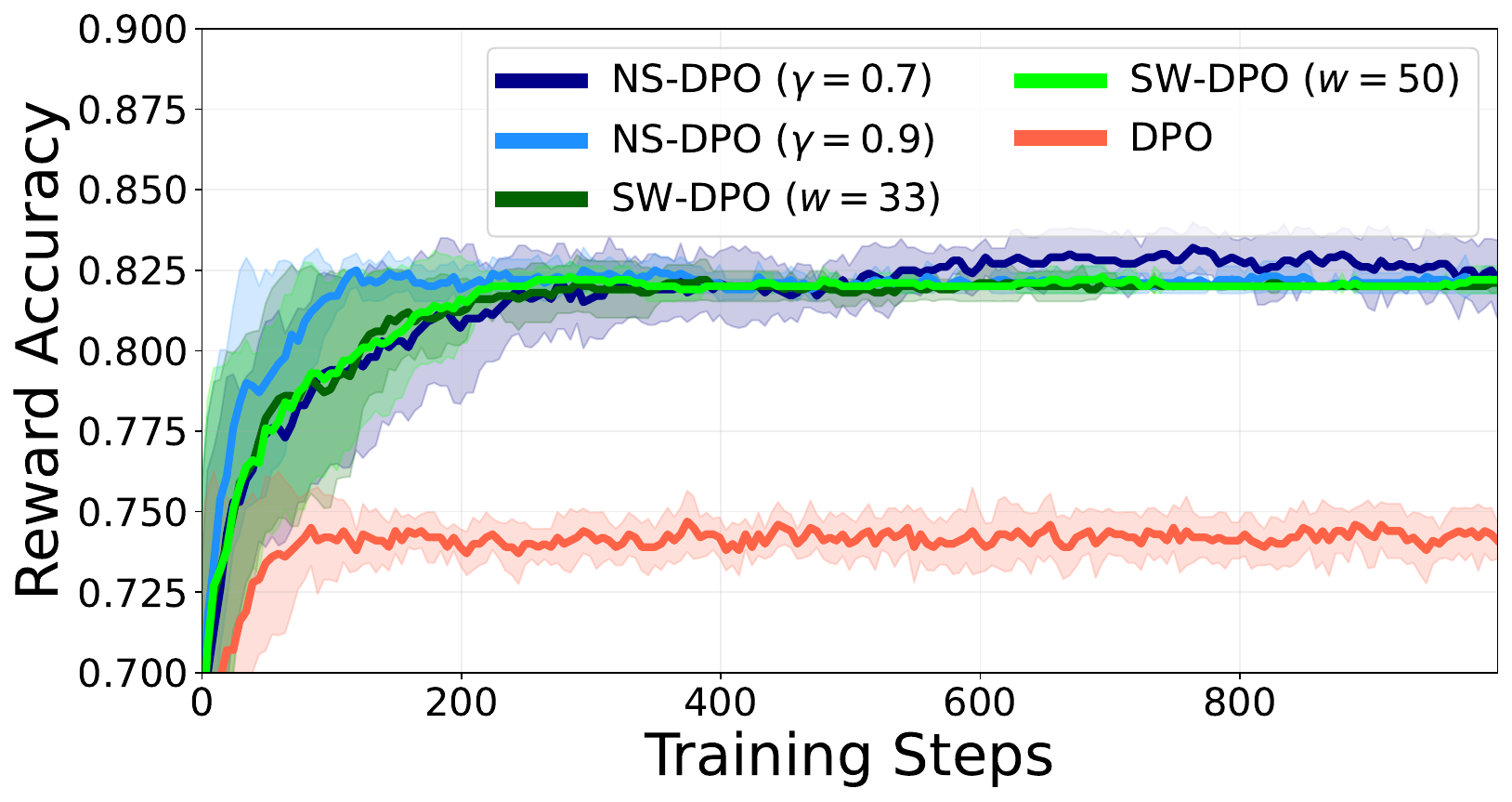}
    \includegraphics[width=0.45\textwidth]{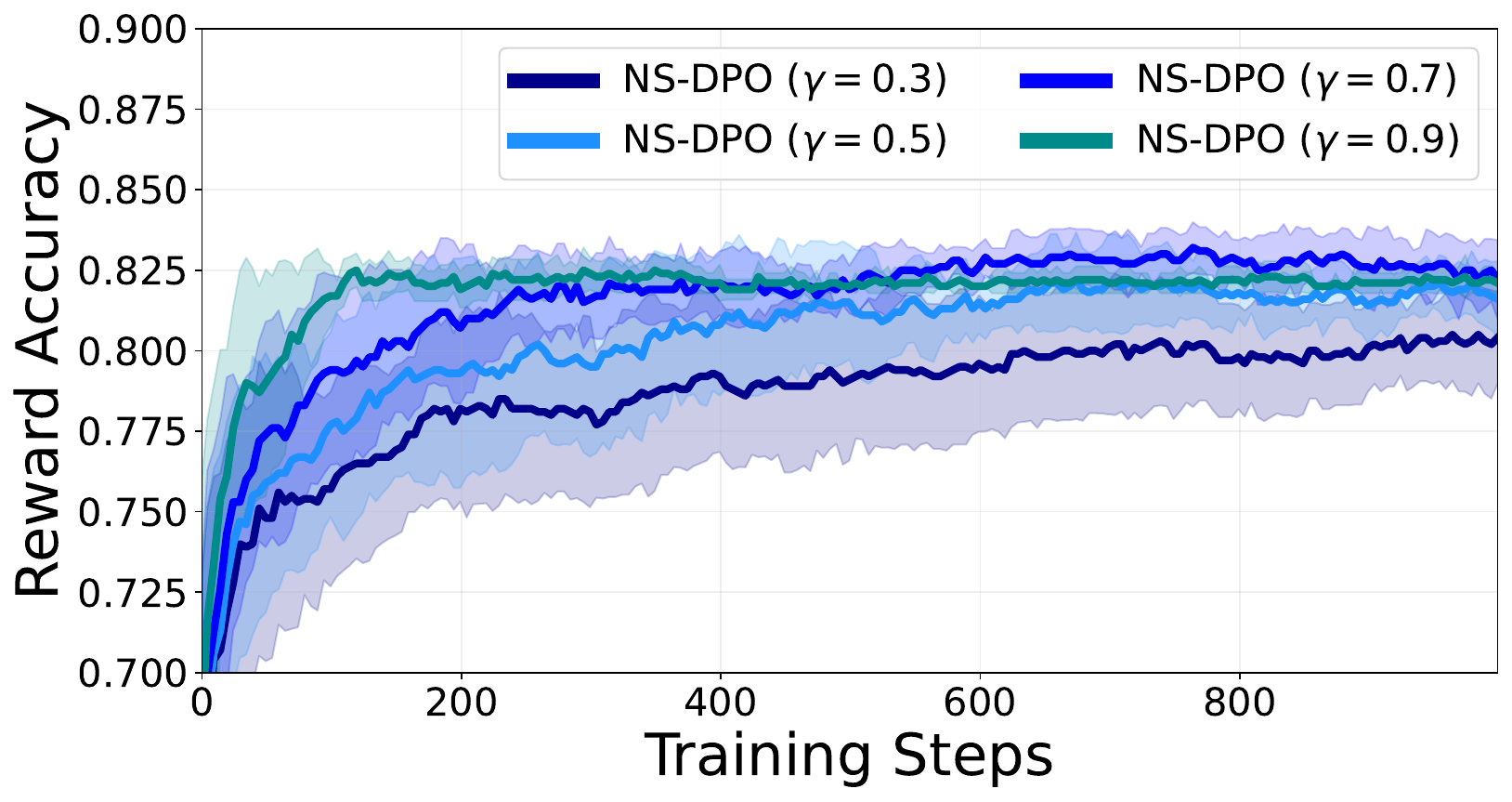}
    \captionsetup{font=footnotesize}
    \caption{Synthetic experiment results with $d_x = 4, |\A| = 16$. The shaded area represents the standard deviation of each algorithm. [Left] NS-DPO and SW-DPO successfully addresses the non-stationarity present in the dataset, while stationary DPO fails to do so. NS-DPO shows faster training than SW-DPO, even compared to the case where the value of the window parameter $w$ for SW-DPO is set to the optimal value of 33. [Right] An ablation study on how different values of the discount factor $\gamma$ affect the training of NS-DPO. As the value of $\gamma$ becomes larger, the final test accuracy of the policy is achieved in fewer training steps.\looseness=-1} 
    \label{fig: experiment_synthetic}
\end{figure*}

\vspace{-5pt}

\begin{figure}[t]
    \centering
    \includegraphics[width=0.45\textwidth]{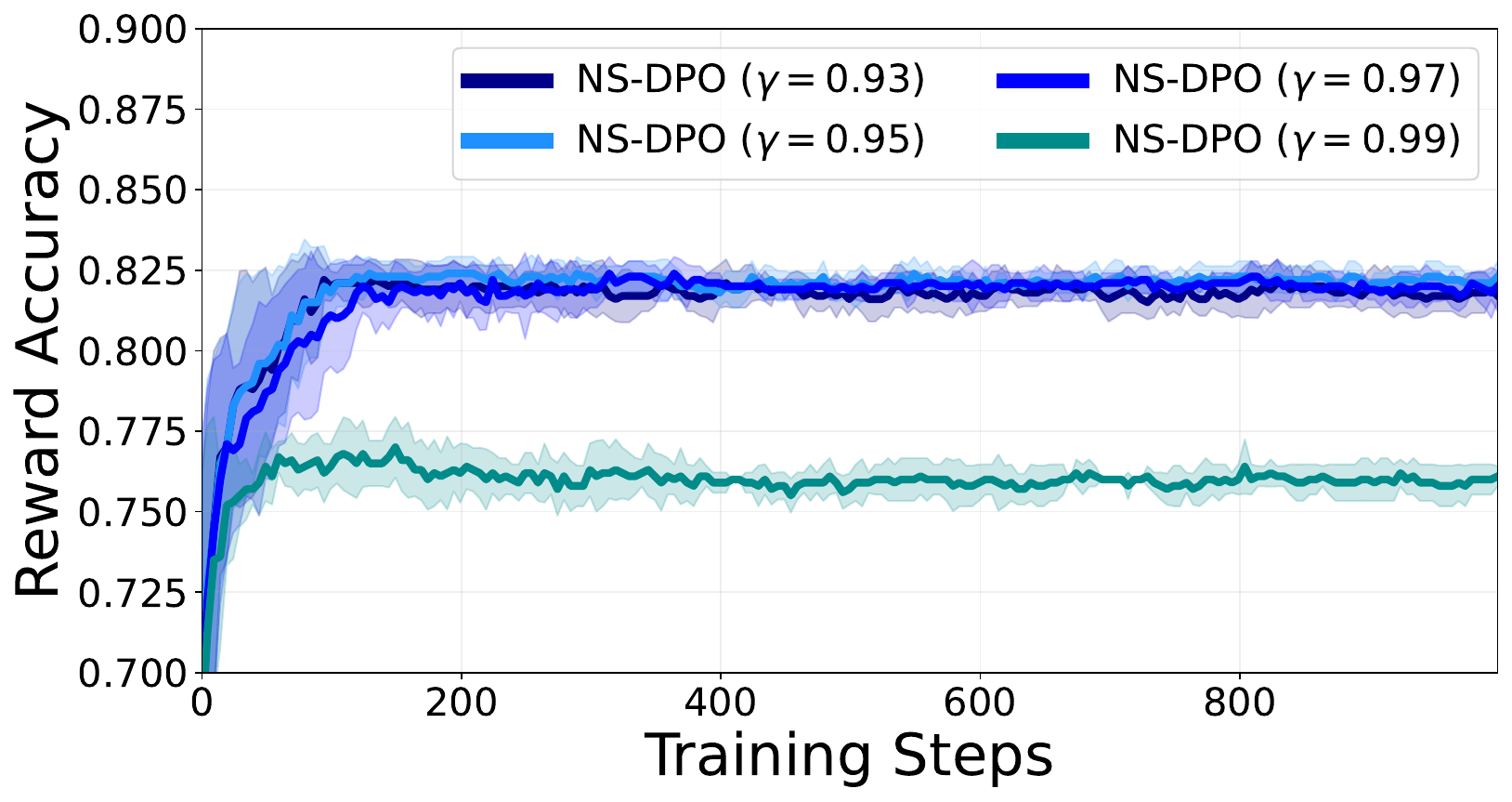}
    \includegraphics[width=0.45\textwidth]{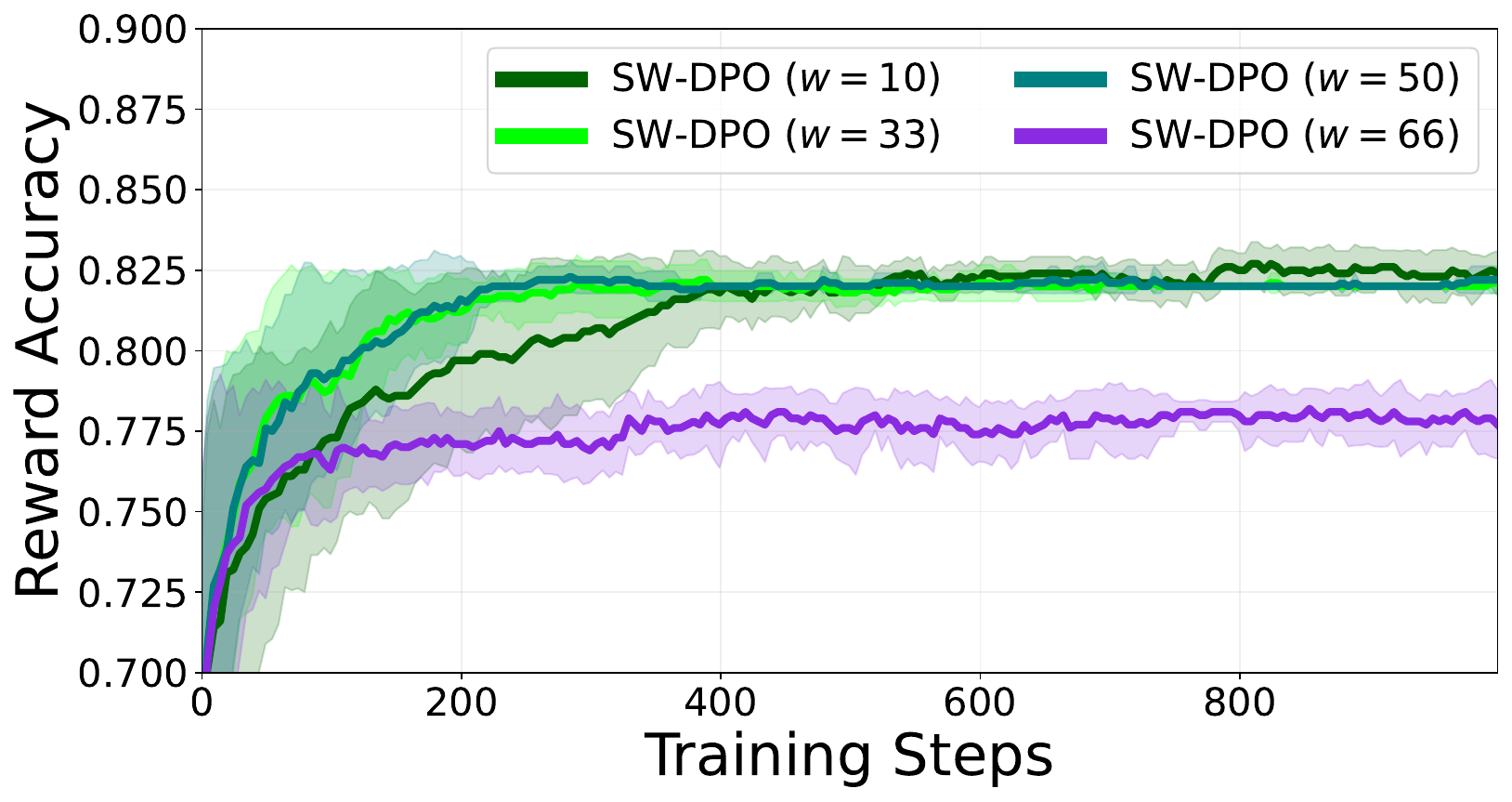}
    \captionsetup{font=footnotesize}
    \caption{[Left] Performance of NS-DPO with values of $\gamma > 0.9$. NS-DPO shows robust performance with respect to the value of $\gamma$, while it starts resembling the performance of stationary DPO as the value approaches very close to 1, $\gamma > 0.97$. [Right] Expected RLHF objective gap of SW-DPO in the same experiments. The performance of SW-DPO improves as the value of $w$ gets closer to 33, when the algorithm is only learning from datapoints where the preference distribution stays stationary in the given setting. The setting with $w=10$ also shows final performance similar to the case of $w=33$, but it shows slower training because of the reduced amount of data used for training.
    } 
    \label{fig: synthetic_further}
\end{figure}

\textbf{How does NS-DPO perform when specialized to log-linear policy classes?} As shown in the left image of \Cref{fig: experiment_synthetic}, when compared to NS-DPO and SW-DPO, DPO shows the worst performance with respect to the test data. Both NS-DPO and SW-DPO, which account for the preference drift present in the data, show significantly better performance. SW-DPO achieves similar performance to NS-DPO in the later stages of training, but NS-DPO achieves this performance in fewer training steps. As NS-DPO only varies the weights of datapoints, rather than removing them entirely, it can still leverage the information of datapoints in the earlier time steps. The right image of \Cref{fig: experiment_synthetic} shows a comparison of different values of $\gamma$, ranging from 0.3 to 0.9. The results show that the performance of NS-DPO is stable in terms of the final test accuracy across a wide range of values $\gamma \in [0.5, 0.9]$. As the value of $\gamma$ is reduced, only the points closest to the current time step contribute significantly to the gradient update of the model. Thus, as $\gamma$ decreases, NS-DPO requires more training steps to converge the reward accuracy on the test set. 

\textbf{Further Results of NS-DPO and SW-DPO.} We present the experiment results of NS-DPO and SW-DPO on the synthetic dataset with varied values of hyperparameters $\gamma$ and $w$. As shown in \Cref{fig: synthetic_further}, The performance of NS-DPO is robust across varied values of $\gamma$, maintaining its reward accuracy over 80\% when $0.5 \leq \gamma \leq 0.97$. In the case of SW-DPO, the performance is more sensitive to the change of the window size $w$. When $w=10$, it shows similar test performance in the later stage of the training, while the process is visibly slowed down due to the reduced amount of datapoints actually being used. On the other hand, as the window size gets bigger and starts including datapoints where parameter shift introduces conflicting preferences, SW-DPO also shows degrading performance. These results provide further support the advantages of using NS-DPO over SW-DPO, as it shows faster training and less sensitivity to the hyperparameter. 

\newpage
\section{Offline Learning Analysis}
\label{appendix: offline learning analysis}

In this section, we provide the remaining details of the analysis on the offline learning of non-stationary dataset. \looseness=-1

\textbf{Non-Linearity Coefficients.}\label{sec: non-linearity coefficients} Following the analysis from \cite{filippi2010parametric, faury2021regret}, we capture the non-linearity of the sigmoid function in the NS-DPO loss. We use the coefficients $\ks, \cs$, which are the supremum and infimum of $\dot{\sigma}(\tau \langle \phi(x, a) - \phi(x, a'), \theta \rangle)$ over $x\in\X, (a, a')\in\A^2, \theta\in\Theta$ respectively:
\begin{align}
    \ks = \sup_{x\in\X, (a, a')\in\A^2, \theta\in\Theta} \dot{\sigma}(\tau \langle \phi(x, a) - \phi(x, a'), \theta \rangle), \label{eq:def of ks} \\ 
    \cs = \inf_{x\in\X, (a, a')\in\A^2, \theta\in\Theta} \dot{\sigma}(\tau \langle \phi(x, a) - \phi(x, a'), \theta \rangle),\label{eq:def of cs}
\end{align}
while we use $\Rs = \ks / \cs$ to denote the ratio between $\ks$ and $\cs.$

\textbf{Loss and gradient.} We recap the loss of NS-DPO with $\ell_2$ regularisation term: \looseness=-1
\begin{align}
    \Lns_{\texttt{reg}}(\theta) &= -\frac{1}{n}\sum_{i = 1}^n\Big[\gamma^{T - t_i - 1} \big\{
        o_i\log \sigma(h_\theta(x_i, a_i, a'_i)) + (1 - o_i)\log \sigma(h_\theta(x_i, a'_i, a_i))\big\}
    \Big] + \frac{\lambda \cs \tau^2}{2} \|\theta\|^2. \label{eq: NS-DPO loss - offline analysis - recap}
\end{align}
We use \Cref{eq: NS-DPO loss - offline analysis - recap} to draw parallels between the NS-DPO objective in \Cref{eq: NS-DPO Loss} and the logistic regression objective used in the generalised linear bandit setting of \citep{faury2021regret}. We assume the preference label $o_i$ is sampled from a Dynamic Bradley-Terry model with the true unknown environment parameter $\theta^*_{t_i}$. Under this assumption, the mean of the preference label is $\mathbb{E}[o_i | \{x_i, a_i, a_i', t_i\}] = \sigma(h_{\theta_{t_i}^*}(x_i, a_i, a'_i))$. When there is only a unilateral preference sampled for a given prompt-response pairs, the sigmoid function forces the implicit rewards of DPO to have infinitely large scale, driving $p(a \succ a')$ to either 1 or 0 \citep{azar2024general}. The $\ell_2$ regularisation term in our analysis mitigates this problem, by controlling the parameter norm. 
Differentiating \Cref{eq: NS-DPO loss - offline analysis} with respect to the parameter $\theta$ results in\looseness=-1
\begin{align}
    \nabla_\theta\Lns_{\texttt{reg}}(\theta) &= -\frac{1}{n}\sum_{i = 1}^n \tau \gamma^{T - t_i - 1}o_i\phihat_i + \underbrace{\frac{1}{n}\sum_{i = 1}^n\left[
        \tau \gamma^{T - t_i - 1}\sigma(h_\theta(x_i, a_i, a'_i))\phihat_i
    \right] + \lambda \cs \tau^2 \theta}_{:=\gtau(\theta)},
    \label{eq: NS-DPO grad - offline analysis}
\end{align}
where $\phihat_i = \phi(x_i, a_i) - \phi(x_i, a'_i)$ is also introduced for brevity. We denote the parameter-dependent part of the gradient as $\gtau(\theta) = \frac{1}{n}\sum_{i = 1}^n\left[\tau \gamma^{T - t_i - 1}\sigma(h_\theta(x_i, a_i, a'_i))\phihat_i\right] + \lambda \cs\tau^2 \theta$ which we will use to analyse the parameter estimation error.

\textbf{Parameter Projection.} \label{paragraph: parameter projection} Let $\thetahat_T$ denote the parameter minimising the NS-DPO loss defined in \Cref{eq: NS-DPO loss - offline analysis}, $\thetahat_T = \argmin_{\theta \in \Rb^d} \Lns(\theta)$. Due to both learning and tracking aspects of the estimation error, we cannot guarantee that $\thetahat_T$ is within the boundary of the parameter presented in \Cref{assumption: boundedness}, $\thetahat_T \in \Theta$. This motivates a parameter projection method, which enables finding an admissible parameter $\thetatilde_T \in \Theta$ while minimising its deviation from $\thetahat_T$ \citep{faury2021regret, wang2023revisiting}. Using $\thetatilde_T$ in the performance analysis of NS-DPO allows preventing the potential violation of \Cref{assumption: boundedness} when $\thetahat_T$ is used. We perform parameter projection by calculating $\thetahat_T$ by
\begin{equation}\label{eq: parameter_projection_wang}
    \thetatilde_T = \argmin_{\theta \in \Theta}\|\gtau(\thetahat_T) - \gtau(\theta)\|_{(\sigmahat + \lambda I)^{-1}},
\end{equation}
using $\sigmahat$ defined in \Cref{eq: sigma_hat - weighted sample covariance matrix} and $\gtau(\theta)$ defined in \Cref{eq: NS-DPO grad - offline analysis}.

\textbf{Covariance matrices.} In addition to $\sigmahat$ defined in \Cref{eq: sigma_hat - weighted sample covariance matrix} we also define  $\sigmatilde$, to which squared discount weights are applied:
\begin{align}
    \sigmatilde = \frac{1}{n}\sum_{i=1}^n\gamma^{2T - 2t_i - 2}(\phi(x_i, a_i) - \phi(x_i, a'_i))(\phi(x_i, a_i) - \phi(x_i, a'_i))^\intercal. \label{eq: sigma_tilde - square-weighted sample covariance matrix}
\end{align}
Due to its squared application of the exponential weighting, $\sigmahat \succ \sigmatilde$.

\subsection{Estimation Error}
\label{appendix: confidence bounds - offline}

\begin{thmmod}{theorem: estimation error - offline - uniform}{}
    (Estimation error of $\thetatilde_T$.) Let $\delta \in (0, 1], \lambda > 0, \tau > 0$. Let $\thetahat_T$ denote the minimiser of the NS-DPO loss defined in \Cref{eq: NS-DPO loss - offline analysis} on an offline dataset. Let $\thetatilde_T$ denote the parameter obtained by performing the parameter projection procedure on $\thetahat_T$. Then with probability at least $1 - \delta$:
    \begin{align}
        \|\thetatilde_T - \theta^*_T\|_{\sigmahat + \lambda I} &\leq \underbrace{2\sqrt{\lambda}W + \frac{2C_1}{\tau \cs}\sqrt{\frac{d + \log(1 / \delta)}{n}}}_{\text{learning}} + \underbrace{\frac{16L \Rs \mbar}{T(1 - \gamma)^{\tfrac{3}{2}}} \sqrt{\frac{d\mbar}{n}} B_T}_{\text{tracking}}
        \label{eq: confidence bound - offline - combined}
    \end{align}
    where $C_1>0$ is a constant.
\end{thmmod}

Estimation errors in typical stationary settings can be considered as \emph{learning} errors, which are caused by having finite data sampled stochastically. In time-varying settings, the parameter estimation suffers from \emph{tracking} error as well, which is caused by the drift of the underlying true parameter along the time steps \citep{faury2021regret, wang2023revisiting}. In this section, we show how these errors can be disentangled and bounded separately. To do this, we apply the approach of \citep{wang2023revisiting} in contextual bandit setting to our setting of offline preference learning. 

\subsubsection{Bound Decomposition}
\label{appendix: estimation error - bound decomposition}

We begin with the deviation between the optimal parameter $\theta^*_T$ and $\thetatilde_T$, the projected parameter of the NS-DPO estimator $\thetahat_T$:
\begin{align}
    \gtau(\thetatilde_T) - \gtau(\theta^*_T) &= \frac{1}{n}\sum_{i=1}^n \tau \gamma^{T - 1 - t_i}\left[
        \sigma(h_{\thetatilde_T}(x_i, a_i, a'_i)) - \sigma(h_{\theta^*_T}(x_i, a_i, a'_i))
    \right]\phihat_i + \lambda \cs \tau^2(\thetatilde_T - \theta^*_T). \label{eq: bound decomposition - thetatilde and theta^*}
\end{align}

Applying the mean value theorem to the difference of sigmoid functions in \Cref{eq: bound decomposition - thetatilde and theta^*} we get 
\begin{align}
    \gtau(\thetatilde_T) - \gtau(\theta^*_T) &= \frac{1}{n}\sum_{i=1}^n \tau^2 \gamma^{T - 1 - t_i}\left[
        \int_{v=0}^1 \dot{\sigma}(\tau \langle \phihat_i, (1 - v)\theta^*_T + v \thetatilde_T \rangle)dv
    \right]\phihat_i\phihat_i^\intercal(\thetatilde_T - \theta^*_T) \nonumber \\
    &\quad + \lambda \cs \tau^2(\thetatilde_T - \theta^*_T). \nonumber
\end{align}
We can now define a matrix $\GT$ to define the relation between $\gtau(\thetatilde_T) - \gtau(\theta^*_T)$ and $\thetatilde_T - \theta^*_T$:
\begin{align}
    \GT &:= \frac{1}{n}\sum_{i=1}^n \gamma^{T - 1 - t_i} \underbrace{\left[
        \int_{v=0}^1 \dot{\sigma}(\tau \langle \phihat_i, (1 - v)\theta^*_T + v \thetatilde_T \rangle)dv
    \right]}_{\alpha(i, \theta^*_T, \thetatilde_T)} \phihat_i\phihat_i^\intercal + \lambda \cs I, 
    \label{eq: def_GT} \\
    &\gtau(\thetatilde_T) - \gtau(\theta^*_T) = \tau^2 \cdot \GT \cdot (\thetatilde_T - \theta^*_T). \label{eq: gtau-param.diff}
\end{align}
We make a brief aside to show $\GT \succeq \cs (\sigmahat + \lambda I) \succeq 0$ \citep{faury2020improved, filippi2010parametric}, as this is an important property of $\GT$ and one we will use later in the main proof. 
To prove this, we first show that $\alpha(i, \theta^*_T, \thetatilde_T) > \cs$. $\alpha(i, \theta_1, \theta_2)$ is the mean value of $\dot{\sigma}$ along the path between some points $\langle \phihat, \theta_1 \rangle$ and $\langle \phihat ,\theta_2 \rangle$. This is greater than the infimum of $\dot{\sigma}$ at a point along that path, which is in turn greater than the infimum of $\dot{\sigma}$ in the space of parameters $\theta \in \Theta$. The last infimum is the definition of  $\cs$ \Cref{eq:def of cs}. Then  
\begin{align}
    \alpha(i, \theta_1, \theta_2) = \int_{v=0}^{v=1} \dot{\sigma}(\tau ( v\phi_i^{\intercal}\theta_1 - (1- v)\phi_i^{\intercal}\theta_2))dv
    &\geq \mathrm{inf}_{c \in [\phi_i^{\intercal}\theta_1, \phi_i^{\intercal}\theta_2]} [\dot{\sigma}(c)] \nonumber \\
    &\geq \mathrm{inf}_{\phi \in \Phi, \theta \in \Theta} [\dot{\sigma}(\tau \phi^{\intercal}\theta)] = \cs > 0. \label{eq: def_alpha_i}
\end{align}
$\alpha(i, \theta_1, \theta_2) > 0$ comes from the fact that the logistic sigmoid function is strictly increasing and has a gradient greater than zero at every point. Because of this inequality, each element of $\GT$ denoted by $[\GT]_{lk} \forall l,k \in [d]$, is strictly larger than each element of $\cs[\sigmahat]_{lk}$. We use this to prove that $\GT \succeq \cs (\sigmahat + \lambda I)$ for any $v = \theta_1 - \theta_2$. We first remind the reader of the definition of $\sigmahat$: 
\begin{equation*}
\sigmahat = \frac{1}{n}\sum_{i=1}^n \gamma^{T-t_i-1}(\phi(x_i, a_i) - \phi(x_i, a'_i))(\phi(x_i, a_i) - \phi(x_i, a'_i))^\intercal.
% + \lambda I.
\end{equation*}
We then prove the inequality, using the fact that $\alpha$ and $\gamma$ do not depend upon the indices $l, k$ of the vector $v$ to move the sum across indices within the sum over the datapoints
\begin{align}
    v^{\intercal}\GT v &= \sum_{(l,k) \in [d]^2} \Big[ \frac{1}{n}\sum_{i=1}^n \gamma^{T - 1 - t_i} \alpha(i, \theta_1, \theta_2) \phihat_i\phihat_i^\intercal + \lambda \cs I\Big]_{lk} v_l v_k \nonumber \\
    &= \left(\frac{1}{n}\sum_{i=1}^n \gamma^{T - 1 - t_i} \alpha(i, \theta_1, \theta_2) \sum_{(l,k) \in [d]^2} \Big[\phihat_i\phihat_i^\intercal\Big]_{lk}v_l v_k \right) + \lambda \cs \sum_{l \in [d]} v_l^2 \nonumber \\
    &\geq \left(\frac{1}{n}\sum_{i=1}^n \gamma^{T - 1 - t_i} \cs \sum_{(l,k) \in [d]^2} \Big[\phihat_i\phihat_i^\intercal\Big]_{lk}v_l v_k \right) + \lambda \cs \sum_{l \in [d]} v_l^2 \\
    &= \cs \sum_{(l,k) \in [d]^2} \Big[ \underbrace{\frac{1}{n}\sum_{i=1}^n \gamma^{T - 1 - t_i} \phihat_i\phihat_i^\intercal + \lambda I}_{\sigmahat + \lambda I}\Big]_{lk} v_l v_k = \cs v^{\intercal} (\sigmahat + \lambda I) v. \label{eq: GT and sigmahat} 
\end{align}

We now continue applying \Cref{eq: gtau-param.diff} to bound the estimation error term:
\begin{align}
    \|\thetatilde_T - \theta^*_T\|_{\sigmahat + \lambda I} &= \frac{1}{\tau^2}\|\GTinv (\gtau(\thetatilde_T) - \gtau(\theta^*_T))\|_{\sigmahat + \lambda I}.
\end{align}

We use \Cref{eq: GT and sigmahat} to apply $\GT^{-1} \prec \frac{1}{\cs}(\sigmahat + \lambda I)^{-1}$:
\begin{align}
    \frac{1}{\tau^2}\|\GTinv (\gtau(\thetatilde_T) - \gtau(\theta^*_T))\|_{\sigmahat + \lambda I} &\prec \frac{1}{\tau^2 \cs}\| \gtau(\thetatilde_T) - \gtau(\theta^*_T)\|_{(\sigmahat + \lambda I)^{-1}}. \label{eq: before_introducing_thetahat}
\end{align}

We add and subtract $\gtau(\thetahat_T)$ inside \Cref{eq: before_introducing_thetahat}, and apply triangle inequality to derive
\begin{align}
    \frac{1}{\tau^2 \cs}\| \gtau(\thetatilde_T) &- \gtau(\theta^*_T)\|_{(\sigmahat + \lambda I)^{-1}} \nonumber \\
    &= \frac{1}{\tau^2 \cs}\|\gtau(\thetatilde_T) - \gtau(\thetahat_T) + \gtau(\thetahat_T) - \gtau(\theta^*_T)\|_{(\sigmahat + \lambda I)^{-1}} \nonumber \\
    &\leq \frac{1}{\tau^2 \cs}\left(
    \|\gtau(\thetatilde_T) - \gtau(\thetahat_T)\|_{(\sigmahat + \lambda I)^{-1}} + \|\gtau(\thetahat_T) - \gtau(\theta^*_T)\|_{(\sigmahat + \lambda I)^{-1}}
    \right).
\end{align}

We use the definition of $\thetatilde_T$ from \Cref{eq: parameter_projection_wang} to derive $\|\gtau(\thetatilde_T) - \gtau(\thetahat_T)\|_{(\sigmahat + \lambda I)^{-1}} \leq \|\gtau(\thetahat_T) - \gtau(\theta^*_T)\|_{(\sigmahat + \lambda I)^{-1}}$ and get
\begin{align}
    \frac{1}{\tau^2 \cs}&\left(
    \|\gtau(\thetatilde_T) - \gtau(\thetahat_T)\|_{(\sigmahat + \lambda I)^{-1}} + \|\gtau(\thetahat_T) - \gtau(\theta^*_T)\|_{(\sigmahat + \lambda I)^{-1}}
    \right) \nonumber \\
    &\leq \frac{2}{\tau^2 \cs} \|\gtau(\thetahat_T) - \gtau(\theta^*_T)\|_{(\sigmahat + \lambda I)^{-1}}. \label{eq: estimation_error_in_gt}
\end{align}

We remind the definition of $\thetahat_T$, which minimises the gradient of the loss defined in \Cref{eq: NS-DPO grad - offline analysis}, making $\nabla \Lns_{\texttt{reg}}(\theta) = 0$:
\begin{align}
    \nabla\Lns_{\texttt{reg}}(\theta) = \frac{1}{n}\sum_{i = 1}^n \tau \gamma^{T - 1 - t_i} \Big[ 
        \sigma(\tau \langle \phihat_i, \thetahat_T - \thetaref\rangle) - o_i
    \Big]\phihat_i + \lambda \cs \tau^2 \thetahat_T = 0. \label{eq: gradient of thetahat_T}
\end{align}
We rearrange the terms in \Cref{eq: gradient of thetahat_T} to derive $\gtau(\thetahat_T)$ on one side of the equation:
\begin{align}
    \underbrace{\frac{1}{n}\sum_{i=1}^n \tau \gamma^{T-1-t_i} \sigma(\tau \langle \phihat_i, \thetahat_T - \thetaref\rangle)\phihat_i + \lambda \cs \tau^2 \thetahat_T}_{=\gtau(\thetahat_T)} = \frac{1}{n}\sum_{i=1}^n \tau \gamma^{T-1-t_i} o_i \phihat_i. \label{eq: gtau_thetahat}
\end{align}

We apply the result of \Cref{eq: gtau_thetahat} to obtain
\begin{align}
    \gtau(\thetahat_T) - \gtau(\theta^*_T) &= \frac{1}{n}\sum_{i=1}^n \tau \gamma^{T-1-t_i} [o_i - \sigma(h_{\theta^*_T}(x_i, a_i, a'_i))]\phihat_i - \lambda \cs \tau^2\theta^*_T. 
\end{align}

Using the fact that the preference label $o_i$ is obtained from the optimal parameter at time step $t_i$, we define $\epsilon_i = o_i - \sigma(\tau\langle \phihat_i, \theta^*_{t_i} - \thetaref\rangle)$, and use $o_i = \epsilon_i + \sigma(\tau\langle \phihat_i, \theta^*_{t_i} - \thetaref\rangle)$ to get
\begin{align}
    \frac{1}{n}\sum_{i=1}^n &\tau \gamma^{T-1-t_i} [o_i - \sigma(h_{\theta^*_T}(x_i, a_i, a'_i))]\phihat_i - \lambda \cs \tau^2\theta^*_T \nonumber \\
    &= \frac{1}{n}\sum_{i=1}^n \tau \gamma^{T-1-t_i} [\epsilon_i + \sigma(\tau\langle \phihat_i, \theta^*_{t_i} - \thetaref\rangle) - \sigma(h_{\theta^*_T}(x_i, a_i, a'_i))]\phihat_i - \lambda \cs \tau^2\theta^*_T \nonumber \\
    &= \underbrace{\frac{1}{n}\sum_{i=1}^n \tau \gamma^{T-1-t_i} [\sigma(\tau\langle \phihat_i, \theta^*_{t_i} - \thetaref\rangle) - \sigma(h_{\theta^*_T}(x_i, a_i, a'_i))]\phihat_i}_{\text{tracking}} \nonumber \\
    &\quad + \underbrace{\frac{1}{n}\sum_{i=1}^n \tau \gamma^{T-1-t_i}\epsilon_i\phihat_i - \lambda \cs \tau^2\theta^*_T}_{\text{learning}}. \label{eq: result_decomposition}
\end{align}

We use terms in \Cref{eq: result_decomposition} with \Cref{eq: estimation_error_in_gt} to define learning error and tracking error:
\begin{align}
    \xi^\mathrm{learn} &= \frac{2}{\tau^2 \cs}\|\frac{1}{n}\sum_{i=1}^n \tau \gamma^{T-1-t_i}\epsilon_i\phihat_i - \lambda \cs \tau^2\theta^*_T\|_{(\sigmahat + \lambda I)^{-1}} \label{eq: xi_learn} \\
    \xi^\mathrm{track} &= \frac{2}{\tau^2 \cs}\|\frac{1}{n}\sum_{i=1}^n \tau \gamma^{T-1-t_i} [\sigma(\tau\langle \phihat_i, \theta^*_{t_i} - \thetaref\rangle) - \sigma(h_{\theta^*_T}(x_i, a_i, a'_i))]\phihat_i\|_{(\sigmahat + \lambda I)^{-1}}. \label{eq: xi_track}
\end{align}

Bounding each of \Cref{eq: xi_learn} and \Cref{eq: xi_track} results in \Cref{theorem: estimation error - offline - uniform}. The detailed bounds for the tracking and learning terms are provided in \Cref{appendix: confidence set - learning - offline} and \Cref{appendix: confidence set - tracking - offline} respectively.

\subsubsection{Confidence Sets: Learning}
\label{appendix: confidence set - learning - offline}

We begin with the definition of the learning error:
\begin{align}
    \xi^\mathrm{learn} &= \frac{2}{\tau^2 \cs}\|\frac{1}{n}\sum_{i=1}^n \tau \gamma^{T-1-t_i}\epsilon_i\phihat_i - \lambda \cs \tau^2\theta^*_T\|_{(\sigmahat + \lambda I)^{-1}}. \label{eq: beginning_learning error}
\end{align}

We bound the norm of \Cref{eq: beginning_learning error} with respect to $\sigmatilde + \lambda I$, using the fact that $\sigmahat \succ \sigmatilde$ and $\sigmatilde + \lambda I\succeq \lambda I$:
\begin{align}
   \Big\|\frac{1}{n}\sum_{i=1}^n \tau \gamma^{T - 1 - t_i}\epsilon_i \phihat_i &-\lambda \cs \tau^2 \theta^*_T\Big\|_{(\sigmahat + \lambda I)^{-1}} \nonumber \\
   &\leq \Big\|\frac{1}{n}\sum_{i=1}^n \tau \gamma^{T - 1 - t_i}\epsilon_i \phihat_i -\lambda \cs \tau^2 \theta^*_T\Big\|_{(\sigmatilde + \lambda I)^{-1}} \nonumber \\
    &\leq \|\lambda \cs \tau^2 \theta^*_T\|_{(\lambda I)^{-1}} + \Big\|\frac{1}{n}\sum_{i=1}^n \tau \gamma^{T - 1 - t_i}\epsilon_i \phihat_i\Big\|_{(\sigmatilde + \lambda I)^{-1}} \nonumber \\
    &\leq \tau^2\sqrt{\lambda} \cs  W + \Big\|\frac{1}{n}\sum_{i=1}^n \tau \gamma^{T - 1 - t_i}\epsilon_i \phihat_i\Big\|_{(\sigmatilde + \lambda I)^{-1}}. \label{eq: learning - theta^*_T}
\end{align}
We can use the $\epsilon_i$'s property of being a sub-Gaussian random variable, sampled i.i.d. during the creation of the dataset. We apply Theorem 2.1 of \citep{hsu2012tail} to \Cref{eq: learning - theta^*_T}, resulting in a bound holding with probability at least $1 - \delta$:
\begin{align}
    \Big\|\frac{1}{n}\sum_{i=1}^n \tau \gamma^{T - 1 - t_i}\epsilon_i \phihat_i\Big\|_{(\sigmatilde + \lambda I)^{-1}} 
    &\leq \tau C_1 \sqrt{\frac{d + \log(1/\delta)}{n}} \label{eq: learning-gtau-bound} = \betaTd,
\end{align}
where $C_1$ denotes a constant introduced for bounding purpose. We provide the details of applying \citep{hsu2012tail}'s theorem in \Cref{appendix: applying Bernstein's inequality}.

We now go back to the original definition of learning error term $\xi^\mathrm{learn}$ and bound it. We use the result in \Cref{eq: learning - theta^*_T} and \Cref{eq: learning-gtau-bound} to derive
\begin{align}
    \xi^\mathrm{learn} &= \frac{2}{\tau^2 \cs}\|\frac{1}{n}\sum_{i=1}^n \tau \gamma^{T-1-t_i}\epsilon_i\phihat_i - \lambda \cs \tau^2\theta^*_T\|_{(\sigmahat + \lambda I)^{-1}} \nonumber \\
    &= \frac{2}{\tau^2 \cs} \left(
        \tau^2\sqrt{\lambda} \cs  W + \tau C_1 \sqrt{\frac{d + \log(1/\delta)}{n}}
    \right) \nonumber \\
    &= 2\sqrt{\lambda} W + \frac{2C_1}{\tau \cs}\sqrt{\frac{d + \log(1/\delta)}{n}} ,\label{eq: xi_learn-bounding-0}
\end{align}
which finishes the bounding of the learning error.

\subsubsection{Estimation Error: Tracking}
\label{appendix: confidence set - tracking - offline}

We begin with the definition of the tracking error:
\begin{align}
    \xi^\mathrm{track} &= \frac{2}{\tau^2 \cs}\Big\|\frac{1}{n}\sum_{i=1}^n \tau \gamma^{T-1-t_i} [\sigma(\tau\langle \phihat_i, \theta^*_{t_i} - \thetaref\rangle) - \sigma(h_{\theta^*_T}(x_i, a_i, a'_i))]\phihat_i\Big\|_{(\sigmahat + \lambda I)^{-1}} \nonumber \\
    &= \frac{2}{\tau^2 \cs}\Big\|\frac{1}{n}\sum_{i=1}^n \tau \gamma^{T-1-t_i} [\sigma(\tau\langle \phihat_i, \theta^*_{t_i} - \thetaref\rangle) - \sigma(\tau \langle \phihat_i, \theta^*_T - \thetaref \rangle)]\phihat_i\Big\|_{(\sigmahat + \lambda I)^{-1}}. \label{eq: tracking_offline_0}
\end{align}

We remind that using \Cref{eq: def_alpha_i}, $\alpha(i, \theta^*_{t_i}, \theta^*_T)$ is
\begin{align}
    \alpha(i, \theta^*_{t_i}, \theta^*_T) &:= \int_{v=0}^1 \dot{\sigma}(\tau \langle \phihat_i, (1 - v)\theta^*_T + v \theta^*_{t_i} \rangle)dv. \label{eq: alpha_tracking}
\end{align}

Applying the man value theorem to \Cref{eq: tracking_offline_0}, we obtain
\begin{align}
    \frac{2}{\tau^2 \cs}\Big\|\frac{1}{n}\sum_{i=1}^n &\tau \gamma^{T-1-t_i} [\sigma(\tau\langle \phihat_i, \theta^*_{t_i} - \thetaref\rangle) - \sigma(\tau \langle \phihat_i, \theta^*_T - \thetaref \rangle)]\phihat_i\Big\|_{(\sigmahat + \lambda I)^{-1}} \nonumber \\
    &= \frac{2}{\tau^2 \cs}\Big\|\frac{1}{n}\sum_{i=1}^n \tau^2 \gamma^{T-1-t_i}\alpha(i, \theta^*_{t_i}, \theta^*_T) \phihat_i\phihat_i^\intercal (\theta^*_{t_i} - \theta^*_T) \Big\|_{(\sigmahat + \lambda I)^{-1}}. \label{eq: tracking - applying alpha}
\end{align}

We apply telescopic sum, which separates $\theta^*_{t_i} - \theta^*_T$ into differences of the optimal parameters between each datapoint:
\begin{align}
     \Big\|\frac{1}{n}\sum_{i=1}^n \tau^2 \gamma^{T-1-t_i}\alpha(i, \theta^*_{t_i}, \theta^*_T)& \phihat_i\phihat_i^\intercal (\theta^*_{t_i} - \theta^*_T) \Big\|_{(\sigmahat + \lambda I)^{-1}} \nonumber \\
     &= \Big\|\frac{1}{n}\sum_{i=1}^{n}\tau^2\gamma^{T-1-t_i} \alpha(i, \theta^*_{t_i}, \theta^*_T) \phihat_i \phihat_i^\intercal \Big(
        \sum_{p=i}^{n} (\theta^*_{t_p} - \theta^*_{t_{p+1}})
    \Big)\Big\|_{(\sigmahat + \lambda I)^{-1}},
\end{align}
where we use $t_{n+1}$ to denote $T$.

Then we use $\sum_{i=k}^n \sum_{j=i}^n a_{i, j} = \sum_{j=k}^n \sum_{i=k}^j a_{i, j}$ to rearrange the terms inside the summation:
\begin{align}
    \Big\|\frac{1}{n}\sum_{i=1}^{n}&\tau^2\gamma^{T-1-t_i} \alpha(i, \theta^*_{t_i}, \theta^*_T) \phihat_i \phihat_i^\intercal \Big(
        \sum_{p=i}^{n} (\theta^*_{t_p} - \theta^*_{t_{p+1}})
    \Big)\Big\|_{(\sigmahat + \lambda I)^{-1}} \nonumber \\
    &= \Big\|\sum_{p=1}^{n}\frac{1}{n}\sum_{i=1}^{p}\tau^2\gamma^{T-1-t_i} \alpha(i, \theta^*_{t_i}, \theta^*_T) \phihat_i \phihat_i^\intercal 
         (\theta^*_{t_p} - \theta^*_{t_{p+1}})
    \Big\|_{(\sigmahat + \lambda I)^{-1}}.
\end{align}

We use $\alpha(i, \theta^*_{t_i}, \theta^*_T) \leq \ks$ using the definition of $\alpha_i$ in \Cref{eq: def_alpha_i} to get
\begin{align}
    \Big\|\sum_{p=1}^{n}&\frac{1}{n}\sum_{i=1}^{p}\tau^2\gamma^{T-1-t_i} \alpha(i, \theta^*_{t_i}, \theta^*_T) \phihat_i \phihat_i^\intercal 
         (\theta^*_{t_p} - \theta^*_{t_{p+1}})
    \Big\|_{(\sigmahat + \lambda I)^{-1}} \nonumber \\
    &\leq \tau^2 \ks \Big\|\sum_{p=1}^{n}\frac{1}{n}\sum_{i=1}^{p}\gamma^{T-1-t_i} \phihat_i \phihat_i^\intercal 
         (\theta^*_{t_p} - \theta^*_{t_{p+1}})
    \Big\|_{(\sigmahat + \lambda I)^{-1}}.
\end{align}

We then apply triangle inequality and Cauchy-Schwarz inequality to get 
\begin{align}
    \tau^2 \ks &\Big\|\sum_{p=1}^{n}\frac{1}{n}\sum_{i=1}^{p}\gamma^{T-1-t_i} \phihat_i \phihat_i^\intercal 
         (\theta^*_{t_p} - \theta^*_{t_{p+1}})
    \Big\|_{(\sigmahat + \lambda I)^{-1}} \nonumber \\
    &\leq \tau^2 \ks \sum_{p=1}^{n}\Big\|\frac{1}{n}\sum_{i=1}^{p}\gamma^{T-1-t_i} \phihat_i \|\phihat_i^\intercal\|_2
         \|\theta^*_{t_p} - \theta^*_{t_{p+1}}\|_2
    \Big\|_{(\sigmahat + \lambda I)^{-1}}.
\end{align}

We use $\|\phihat\| \leq 2L$ and arrange terms to obtain
\begin{align}
    \tau^2 \ks &\sum_{p=1}^{n}\Big\|\frac{1}{n}\sum_{i=1}^{p}\gamma^{T-1-t_i} \phihat_i \|\phihat_i^\intercal\|_2
         \|\theta^*_{t_p} - \theta^*_{t_{p+1}}\|_2
    \Big\|_{(\sigmahat + \lambda I)^{-1}} \nonumber \\
    &\leq 2L \tau^2 \ks \sum_{p=1}^{n}\underbrace{\frac{1}{n}\sum_{i=1}^{p}\gamma^{T-1-t_i} \|\phihat_i\|_{(\sigmahat + \lambda I)^{-1}}}_{=v_1}\|\theta^*_{t_p} - \theta^*_{t_{p+1}}\|_2. \label{eq: tracking_before_v1}
\end{align}

Here we bound the term $v_1$. We first apply Jensen's inequality to derive
\begin{align}
    v_1 &\leq \sqrt{\frac{1}{n}\sum_{i=1}^{p}\gamma^{T-1-t_i} }\sqrt{\frac{1}{n}\sum_{i=1}^{p}\gamma^{T-1-t_i}\|\phihat_i\|^2_{(\sigmahat + \lambda I)^{-1}}} 
    \nonumber \\
    &= \gamma^{\frac{T - 1}{2}}\sqrt{\frac{1}{n}\sum_{i=1}^{p}\gamma^{-t_i} }\sqrt{\frac{1}{n}\sum_{i=1}^{p}\gamma^{T-1-t_i}\|\phihat_i\|^2_{(\sigmahat + \lambda I)^{-1}}}. \label{eq: tracking_before_trace}
\end{align}

We then use the property of trace operation and $\sigmahat \succ \sum_{i=1}^{p}\gamma^{T-1-t_i}\phihat_i\phihat_i^\intercal$ from \Cref{eq: sigma_hat - weighted sample covariance matrix} to get
\begin{align}
    \frac{1}{n}\sum_{i=1}^{p}\gamma^{T-1-t_i}\|\phihat_i\|^2_{(\sigmahat + \lambda I)^{-1}} &= \frac{1}{n}\sum_{i=1}^{p}\gamma^{T-1-t_i}\mathrm{tr}\left(\phihat_i^\intercal (\sigmahat + \lambda I)^{-1} \phihat_i \right) \nonumber \\
    &= \mathrm{tr}\left((\sigmahat + \lambda I)^{-1}\frac{1}{n}\sum_{i=1}^{p}\gamma^{T-1-t_i}\phihat_i\phihat_i^\intercal  \right) \nonumber \\
    &\leq \mathrm{tr}\left(I_d\right) = d. \label{eq: tracking_after_trace}
\end{align}
We apply \Cref{assumption: temporal_distribution_new} here. Because each time step can have at maximum $\mbar$ datapoints, we can upper bound $\frac{1}{n}\sum_{i=1}^{p}\gamma^{-t_i}$ with
\begin{align}
    \frac{1}{n}\sum_{i=1}^{p}\gamma^{-t_i} &\leq \frac{\mbar}{n}\sum_{k=1}^{t}\gamma^{-k} = \frac{\mbar\gamma (\gamma^{-(t+1)} - 1)}{n(1 - \gamma)}, \label{eq: tracking_mbar0}
\end{align}
where $t = \left\lceil\frac{|[p]|}{\mbar} \right\rceil$. 
We combine \Cref{eq: tracking_after_trace} and \Cref{eq: tracking_mbar0} to obtain
\begin{align}
    2L \tau^2 \ks &\sum_{p=1}^{n}\frac{1}{n}\sum_{i=1}^{p}\gamma^{T-1-t_i} \|\phihat_i\|_{(\sigmahat + \lambda I)^{-1}} \|\theta^*_{t_p} - \theta^*_{t_{p+1}}\|_2 \nonumber \\
    &\leq 2L \tau^2 \ks \sum_{p=1}^{n}\gamma^{\frac{T-1}{2}} \sqrt{\frac{d \mbar\gamma (\gamma^{-(t+1)} - 1)}{n(1 - \gamma)}} \|\theta^*_{t_p} - \theta^*_{t_{p+1}}\|_2.
\end{align}
We apply \Cref{assumption: temporal_distribution_new} again to upper bound the summation as $\sum_{p=1}^n v_p \leq \mbar \sum_{t=1}^{T-1} v_t$, getting
\begin{align}
    2L \tau^2 \ks &\sum_{p=1}^{n}\gamma^{\frac{T-1}{2}} \sqrt{\frac{d \mbar\gamma (\gamma^{-(t+1)} - 1)}{n(1 - \gamma)}} \|\theta^*_{t_p} - \theta^*_{t_{p+1}}\|_2 \nonumber \\
    &\leq 2L \tau^2 \ks \mbar \sum_{t=1}^{T-1}\gamma^{\frac{T-1}{2}} \sqrt{\frac{d \mbar\gamma (\gamma^{-(t+1)} - 1)}{n(1 - \gamma)}} \|\theta^*_{t} - \theta^*_{t+1}\|_2. \label{eq: tracking_uniform_0}
\end{align}
We apply $v = \frac{1}{T}\sum_{k=1}^{T} v$ to introduce another summation:
\begin{align}
    2L \tau^2 \ks \mbar &\sum_{t=1}^{T-1}\gamma^{\frac{T-1}{2}} \sqrt{\frac{d \mbar \gamma (\gamma^{-(t+1)} - 1)}{n(1 - \gamma)}} \|\theta^*_{t} - \theta^*_{t+1}\|_2 \nonumber \\
    &= \frac{2L \tau^2 \ks \mbar}{T}\sum_{k=1}^T\sum_{t=1}^{T-1}\gamma^{\frac{T-1}{2}} \sqrt{\frac{d \mbar\gamma (\gamma^{-(t+1)} - 1)}{n(1 - \gamma)}} \|\theta^*_{t} - \theta^*_{t+1}\|_2. \label{eq: tracking_before_gamma_bounding}
\end{align}

Because $\gamma < 1$, we can bound 
\begin{equation}
    \sum_{k=1}^T \sum_{t=1}^{T-1}\gamma^{\frac{T-1}{2}}\sqrt{\frac{d \mbar\gamma (\gamma^{-(t+1)} - 1)}{n(1 - \gamma)}} \leq 2 \sum_{t=1}^{T-1} \sum_{k=t+1}^T\gamma^{\frac{k-1}{2}} \sqrt{\frac{d\mbar\gamma (\gamma^{-(t+1)} - 1)}{n(1 - \gamma)}}   
\end{equation}
and apply geometric sum to obtain
\begin{equation}
    2 \sum_{t=1}^{T-1} \sum_{k=t+1}^T\gamma^{\frac{k-1}{2}} \sqrt{\frac{d\mbar\gamma (\gamma^{-(t+1)} - 1)}{n(1 - \gamma)}} = 2 \sum_{t=1}^{T-1} \frac{\gamma^{\tfrac{t}{2}} - \gamma^{\tfrac{T}{2}}}{1 - \gamma^{\tfrac{1}{2}}} \sqrt{\frac{d \mbar \gamma (\gamma^{-(t+1)} - 1)}{n(1 - \gamma)}}.
\end{equation}

We use $\gamma < 1$ again to derive $\frac{1 + \gamma^{\tfrac{1}{2}}}{2} < 1$, and get
\begin{align}
    2 \sum_{t=1}^{T-1} \frac{\gamma^{\tfrac{t}{2}} - \gamma^{\tfrac{T}{2}}}{1 - \gamma^{\tfrac{1}{2}}} \sqrt{\frac{d \mbar\gamma (\gamma^{-(t+1)} - 1)}{n(1 - \gamma)}} &\leq 2 \sum_{t=1}^{T-1} \frac{\gamma^{\tfrac{t}{2}} - \gamma^{\tfrac{T}{2}}}{1 - \gamma^{\tfrac{1}{2}} \frac{1 + \gamma^{\tfrac{1}{2}}}{2}} \sqrt{\frac{d \mbar\gamma (\gamma^{-(t+1)} - 1)}{n(1 - \gamma)}} \nonumber \\
    &= 4 \sum_{t=1}^{T-1} \frac{\gamma^{\tfrac{t}{2}} - \gamma^{\tfrac{T}{2}}}{1 - \gamma} \sqrt{\frac{d \mbar \gamma (\gamma^{-(t+1)} - 1)}{n(1 - \gamma)}}.
\end{align}

We then use $\left(\gamma^{\tfrac{t}{2}} - \gamma^{\tfrac{T}{2}}\right) \sqrt{\gamma (\gamma^{-(t+1)} - 1)}\leq \gamma^{\tfrac{t}{2}} \gamma^{-\tfrac{t}{2}} = 1$ to derive
\begin{equation}
     4 \sum_{t=1}^{T-1} \frac{\gamma^{\tfrac{t}{2}} - \gamma^{\tfrac{T}{2}}}{1 - \gamma} \sqrt{\frac{d \mbar \gamma (\gamma^{-(t+1)} - 1)}{n(1 - \gamma)}} \leq  4 \sqrt{\frac{d\mbar}{n}}\sum_{t=1}^{T-1} \frac{1}{(1 - \gamma)^{\tfrac{3}{2}}}. \label{eq: tracking_wang_cleaned0}
\end{equation}

We use the result from \Cref{eq: tracking_wang_cleaned0} to \Cref{eq: tracking_before_gamma_bounding}, and use the definition of variation budget $B_T$ from \Cref{assumption: variation budget} to get
\begin{align}
    \frac{2L \tau^2 \ks \mbar}{T}&\sum_{k=1}^T\sum_{t=1}^{T-1}\gamma^{\frac{T-1}{2}} \sqrt{\frac{d \mbar \gamma (\gamma^{-(t+1)} - 1)}{n(1 - \gamma)}} \|\theta^*_{t} - \theta^*_{t+1}\|_2 \nonumber \\
    &\leq \frac{8L \tau^2 \ks \mbar}{T} \sqrt{\frac{d\mbar}{n}}\sum_{t=1}^{T-1} \frac{1}{(1 - \gamma)^{\tfrac{3}{2}}} \|\theta^*_{t} - \theta^*_{t+1}\|_2 \nonumber \\
    &\leq \frac{8L \tau^2 \ks \mbar}{T(1 - \gamma)^{\tfrac{3}{2}}} \sqrt{\frac{d\mbar}{n}} B_T. \label{eq: tracking_wang_result_uniform}
\end{align}
We now combine \Cref{eq: tracking_wang_result_uniform} with \Cref{eq: xi_track} to derive the full bound of the tracking error:
\begin{align}
    \xi^\mathrm{track} = \frac{16L \Rs \mbar}{T(1 - \gamma)^{\tfrac{3}{2}}} \sqrt{\frac{d\mbar}{n}} B_T.
    \label{eq: xi_track-bounding-0}
\end{align}

We now use \Cref{eq: xi_track-bounding-0} with \Cref{eq: xi_learn-bounding-0} to obtain the full estimation error:
\begin{align}
    \|\thetahat_T - \thetahat^*_T\|_{\sigmahat + \lambda I} &\leq \xi^\mathrm{learn} + \xi^\mathrm{track} \nonumber \\
    &\leq 2\sqrt{\lambda}W + \frac{2C_1}{\tau \cs}\sqrt{\frac{d + \log(1 / \delta)}{n}} + \frac{16L \Rs \mbar}{T(1 - \gamma)^{\tfrac{3}{2}}} \sqrt{\frac{d\mbar}{n}} B_T, \label{eq: estimation_error_uniform_0}
\end{align}
which concludes the analysis for \Cref{theorem: estimation error - offline - uniform}.

\subsection{Regret Bound}
\label{appendix: regret bound}

\begin{thmmod}{theorem: regret bound - offline - uniform}{}
    (Regret bound of $\thetatilde_T$) Let $\delta \in (0, \tfrac{1}{2}], \tau > 0$. 
    Let $\thetatilde_T$ denote the parameter in $\Theta$ which minimises the NS-DPO loss (\Cref{eq: NS-DPO loss - offline analysis}) on an offline dataset. The following bound holds with probability at least $1 - 2\delta$ and when $\lambda \geq \conccoef$:\looseness=-1
    \begin{align}
        \Roff \leq \frac{\tau \kappa \mbar T (1 - \gamma)}{2 \mubar (1 - \gamma^{T-1})}& \Bigg(
        2\sqrt{\lambda}W + \frac{2C_1}{\tau \cs}\sqrt{\frac{d + \log(1 / \delta)}{n}} + \frac{16L \Rs \mbar}{T(1 - \gamma)^{\tfrac{3}{2}}} \sqrt{\frac{d\mbar}{n}} B_T \Bigg)^2, \nonumber
    \end{align}
    where $C_1 > 0$ denotes a constant. When $\gamma = 1- \left(\tfrac{B_T}{T}\right)^{3/4}$, $\Roff$ satisfies:\looseness=-1
    \begin{align}
        \Roff = \tilde{O}\left(d\ B_T^{3/4}\ n^{-1/4}\right). \nonumber
    \end{align}
\end{thmmod}

\subsubsection{Population Covariance of Feature Differences} 
\label{appendix: population covariance of feature differences}

Let $\sigmadiff$ define the population covariance matrix of the feature differences:
\begin{equation}\label{eq: def_sigmadiff}
    \sigmadiff = \Eb[\phihat \phihat^\intercal],
\end{equation}
where $\phihat = \phi(x, a) - \phi(x, a')$ denotes the feature difference vector, and  the expectation is computed with respect to ${x \sim \X, t \sim \Tcal, a, a'\sim \piref(\cdot | x)}$. We also define the discounted population covariance matrix $\sigmagdiff$:
\begin{equation}\label{eq: def_sigmagdiff}
    \sigmagdiff = \Eb[\gamma^{T-1-t}\phihat \phihat^\intercal],
\end{equation}
where the expectation is computed with respect to the same distributions as $\sigmadiff$.

We then define $\omegaupp(T, \gamma)$:
\begin{align}
    \omegaupp(T, \gamma) = \sup_{v \in \Rb^d} \frac{v^\intercal \sigmadiff v}{v^\intercal \sigmagdiff v}, \label{eq: omegaupp_gamma_piref}
\end{align}

Without any assumptions on the time distribution, $\omegaupp(T, \gamma) \leq \gamma^{-(T-1)}$, which happens when all the datapoints come from the oldest time step.
We use \Cref{assumption: temporal_distribution_new} to obtain a tighter upper bound of $\omegaupp$. Using $\mubar (T-1) \leq n \leq \mbar (T-1)$, we can get
\begin{align}
    \frac{1}{n}\sum_{i=1}^{n} \gamma^{T-1-t_i} \geq \frac{\mubar}{n} \cdot \sum_{t=1}^{T-1} \gamma^{T-1-t} \geq \frac{\mubar}{\mbar (T - 1)} \cdot \sum_{t=1}^{T-1} \gamma^{T-1-t}. \label{eq: lower_bound_gammasum}
\end{align}
We note that the prompt distribution $\X$ and the reference policy $\piref$ are independent from the time step distribution $\mathcal{T}$. Using \Cref{eq: lower_bound_gammasum}, we obtain
\begin{align}
    v^\intercal\sigmagdiff v &\geq \left(\frac{\mubar}{\mbar (T-1)}\sum_{i=0}^{T-2}\gamma^{i}\right) \cdot (v^\intercal\sigmadiff v) =\frac{\mubar(1 - \gamma^{T-1})}{\mbar (T-1)(1 - \gamma)} \cdot (v^\intercal\sigmadiff v), \label{eq: ratio - sigmagdiff and sigmadiff}
\end{align}
which implies $\omegaupp(T, \gamma) \leq \frac{\mbar (T-1)(1 - \gamma)}{\mubar(1 - \gamma^{T-1})}$.

\subsubsection{Decomposing Regret Bound}
\label{appendix: decomposing regret bound}

In order to decompose and bound the detailed elements of the regret bound, we first show the relation between the regret and the estimation error of the model parameters. 

\begin{theorem} \label{theorem: offline regret bound + estimation error}
Let $\delta \in [0, 1]$. Let $\thetatilde_T$ denote the parameter obtained by performing the parameter projection in \Cref{paragraph: parameter projection}, after training with the NS-DPO loss defined in \Cref{eq: NS-DPO loss - offline analysis} on an offline dataset. When $\lambda \geq \conccoef$, with probability at least $1 - \delta$:
    \begin{equation}
        \Roff \leq \frac{\tau \kappa \mbar T (1 - \gamma)}{2 \mubar (1 - \gamma^{T-1})} \|\theta^*_T - \thetatilde_T\|^2_{\sigmahat + \lambda I}.
    \end{equation}
\end{theorem} 

Let $\pi_{\thetatilde_T}$ denote the policy we obtained by training with NS-DPO and performing parameter projection. We use $\Sigma_{\pi_{\thetatilde_T}}$ to denote the population covariance matrix, whose expectation taken with respect to $\pi_{\thetatilde_T}$. We assess the performance of $\pi_\thetatilde$ using the difference in expected non-stationary RLHF objective $\J_T(\pi)$ defined in \Cref{eq: ns rlhf objective}, which is 
\begin{align}
    \J_T(\pi) &= \Eb_{x \sim \X, a \sim \pi}\Big[r(x, a, T) - \tau \DKL[\pi(\cdot | x)\| \piref(\cdot | x)]\Big], \nonumber \\
    \Roff &= \J_T(\pi^*_T) - \J_T(\pi_{\thetatilde_T}) \nonumber \\
    &= \Eb_{x \sim \X}\Big[
        \Eb_{a \sim \pi^*_T(\cdot | x)}[r(x, a, T)] - \tau \DKL[\pi^*_T(\cdot | x)\| \piref(\cdot | x)] \nonumber \\
    &\quad \quad \quad - \Eb_{a' \sim \pi_{\thetatilde_T}(\cdot | x)}[r(x, a', T)] + \tau \DKL[\pi_{\thetatilde_T}(\cdot | x)\| \piref(\cdot | x)]
    \Big]. \label{eq: expected regret_new}
\end{align}

We plug \Cref{eq: ns implicit reward} in \Cref{eq: expected regret_new} to obtain
\begin{align}
    \Roff &= \Eb_{x \sim \X}\Big[
        \Eb_{a \sim \pi^*_T(\cdot | x)}[\tau \log \frac{\pi^*_T(a | x)}{\piref(a | x)}] - \tau \DKL[\pi^*_T(\cdot | x)\| \piref(\cdot | x)] \nonumber \\
    &\quad \quad \quad - \Eb_{a' \sim \pi_{\thetatilde_T}(\cdot | x)}[\tau \log \frac{\pi^*_T(a | x)}{\piref(a | x)}] + \tau \DKL[\pi_{\thetatilde_T}(\cdot | x)\| \piref(\cdot | x)]
    \Big], \label{eq: expected regret_new-2}
\end{align}
where terms with normalisation constant $Z^*_T(x)$ are cancelled out. By using the definition of KL divergence in \Cref{eq: expected regret_new-2} again, we obtain
\begin{align}
    \Roff &= \Eb_{x \sim \X}\Bigg[
        - \Eb_{a' \sim \pi_{\thetatilde_T}(\cdot | x)}\bigg[\tau \log \frac{\pi^*_T(a | x)}{\piref(a | x)}\bigg] + \Eb_{a' \sim \pi_{\thetatilde_T}(\cdot | x)}\bigg[\tau \log \frac{\pi_{\thetatilde_T}(a | x)}{\piref(a | x)}\bigg]
    \Bigg] \nonumber \\ 
    &= \Eb_{x \sim \X}\Bigg[
        \tau \Eb_{a' \sim \pi_{\thetatilde_T}(\cdot | x)}\bigg[\log \frac{\pi_{\thetatilde_T}(a | x)}{\pi^*_T(a | x)}\bigg] 
    \Bigg] \nonumber \\
    &= \Eb_{x \sim \X}\Bigg[
        \tau \DKL[\pi_{\thetatilde_T}(\cdot | x)\| \pi^*_T(\cdot | x)] 
    \Bigg]. \label{eq: expected regret_new-3}
\end{align}

Here, we borrow the analysis in Appendix A.5. of \cite{chowdhury2024provably}. We use the property of the Bergman divergence $\Bb_\Lx$ with its potential function $\Lx(\theta) = \log \sum_{a' \in \A}\langle \theta, \phi(x, a') \rangle$:
\begin{align}
    \DKL[\pi_{\thetatilde_T}(\cdot | x)\| \pi^*_T(\cdot | x)] &= \frac{1}{2}(\theta^*_T - \thetatilde_T)^\intercal \nabla^2 \Lx(\theta)(\theta^*_T - \thetatilde_T)
\end{align}
for a parameter $\theta \in \{t\thetatilde + (1 - t)\theta^*\ :\ t \in [0, 1]\}$ using Taylor's approximation. With log-linear policies, $\Eb_{x \sim \X}[\nabla^2 \Lx(\theta)] = \Sigma_{\pi_\theta}$. We use this to derive the upper bound of \Cref{eq: expected regret_new-3}:
\begin{align}
    \Roff &= \Eb_{x \sim \X}\big[
        \tau \DKL[\pi_{\thetatilde_T}(\cdot | x)\| \pi^*_T(\cdot | x)] 
    \big] \nonumber \\
    &\leq \tau \|\theta^*_T - \thetatilde_T\|^2_{\Sigma_{\pi_\theta}} \nonumber \\
    &= \tau \|\theta^*_T - \thetatilde_T\|^2_{\sigmahat + \lambda I} \frac{
        (\theta^*_T - \thetatilde_T)^\intercal \Sigma_{\pi_\theta}(\theta^*_T - \thetatilde_T)
    }{
        (\theta^*_T - \thetatilde_T)^\intercal (\sigmahat + \lambda I)(\theta^*_T - \thetatilde_T)
    } \label{eq: regret_new_squared}
\end{align}

We now use the following lemma from \citep{chowdhury2024provably}, which relies on the matrix concentration inequality to explain the difference between $\sigmahat$ and $\sigmagdiff$.
\begin{lemma}\label{lemma: chowdhury A.1. - matrix concentration inequality}
    (Lemma A.1. of \citep{chowdhury2024provably}) With probability at least $1 - \delta$, for some universal constant C, we have
    \begin{equation}
        \|\sigmahat - \sigmagdiff\|_2 \leq C \sqrt{d \log(4d/\delta) / n}.   
    \end{equation}
\end{lemma}

\Cref{lemma: chowdhury A.1. - matrix concentration inequality} implies that with probability at least $1-\delta$ and $\lambda \geq \conccoef$:
\begin{align} 
    \sigmahat + \lambda I &\succeq \sigmagdiff + \lambda I - \conccoef \nonumber \\
    &\succeq \sigmagdiff. \label{eq: implication_lemmaa1_chowdhury}
\end{align}

We use \Cref{eq: implication_lemmaa1_chowdhury} to derive
\begin{align}
    \tau \|\theta^*_T - \thetatilde_T\|^2_{\sigmahat + \lambda I} &\frac{
        (\theta^*_T - \thetatilde_T)^\intercal \Sigma_{\pi_\theta}(\theta^*_T - \thetatilde_T)
    }{
        (\theta^*_T - \thetatilde_T)^\intercal (\sigmahat + \lambda I)(\theta^*_T - \thetatilde_T)
    } \nonumber \\
    &\leq \tau \|\theta^*_T - \thetatilde_T\|^2_{\sigmahat + \lambda I} \frac{
        (\theta^*_T - \thetatilde_T)^\intercal \Sigma_{\pi_\theta}(\theta^*_T - \thetatilde_T)
    }{
        (\theta^*_T - \thetatilde_T)^\intercal \sigmagdiff(\theta^*_T - \thetatilde_T)
    }.\label{eq: sigmahat to sigmagdiff}
\end{align}

We then apply the result from \Cref{eq: omegaupp_gamma_piref} which implies $(\|v\|_{\sigmagdiff})^{-1} \leq \sqrt{\omegaupp(T, \gamma)}(\|v\|_{\sigmadiff})^{-1}$:
\begin{align}
    \tau \|\theta^*_T - \thetatilde_T\|^2_{\sigmahat + \lambda I} &\frac{
        (\theta^*_T - \thetatilde_T)^\intercal \Sigma_{\pi_\theta}(\theta^*_T - \thetatilde_T)
    }{
        (\theta^*_T - \thetatilde_T)^\intercal \sigmagdiff(\theta^*_T - \thetatilde_T)
    } \nonumber \\
    &\leq \tau \omegaupp(T, \gamma) \|\theta^*_T - \thetatilde_T\|^2_{\sigmahat + \lambda I} \frac{
        (\theta^*_T - \thetatilde_T)^\intercal \Sigma_{\pi_\theta}(\theta^*_T - \thetatilde_T)
    }{
        (\theta^*_T - \thetatilde_T)^\intercal \sigmadiff(\theta^*_T - \thetatilde_T)
    }. \label{eq: sigmagdiff to sigmadiff}
\end{align}
From the definition of $\sigmadiff$ in \Cref{eq: def_sigmadiff}, $a, a'$ are independently sampled. We combine this fact with the population covariance matrix $\Sigma_{\piref}$, deriving $\sigmadiff = 2\Sigma_{\piref}$. We use this to get
\begin{align}
    \tau \omegaupp(T, \gamma) \|\theta^*_T - \thetatilde_T\|^2_{\sigmahat + \lambda I} &\frac{
        (\theta^*_T - \thetatilde_T)^\intercal \Sigma_{\pi_\theta}(\theta^*_T - \thetatilde_T)
    }{
        (\theta^*_T - \thetatilde_T)^\intercal \sigmadiff(\theta^*_T - \thetatilde_T)
    } \nonumber \\
    &= \frac{\tau \omegaupp(T, \gamma)}{2} \|\theta^*_T - \thetatilde_T\|^2_{\sigmahat + \lambda I} \frac{
        (\theta^*_T - \thetatilde_T)^\intercal \Sigma_{\pi_\theta}(\theta^*_T - \thetatilde_T)
    }{
        (\theta^*_T - \thetatilde_T)^\intercal \Sigma_{\piref}(\theta^*_T - \thetatilde_T)
    }.\label{eq: sigmadiff to sigmapiref}
\end{align}

We use $\kappa = \max_{\pi \in \Pi} \kappa_\pi$ with the definition of $\kappa_\pi$ in \Cref{eq: relative condition number}, along with the result obtained in \Cref{eq: ratio - sigmagdiff and sigmadiff} to use $\omegaupp(T, \gamma) = \frac{(T-1)(1 - \gamma)}{1 - \gamma^{T-1}} \leq \frac{T(1 - \gamma)}{1 - \gamma^{T-1}}$:
\begin{align}
    \frac{\tau \omegaupp(T, \gamma)}{2} \|\theta^*_T - \thetatilde_T\|^2_{\sigmahat + \lambda I} &\frac{
        (\theta^*_T - \thetatilde_T)^\intercal \Sigma_{\pi_\theta}(\theta^*_T - \thetatilde_T)
    }{
        (\theta^*_T - \thetatilde_T)^\intercal \Sigma_{\piref}(\theta^*_T - \thetatilde_T)
    } \nonumber \\
    &\leq \frac{\tau \kappa \omegaupp(T, \gamma)}{2} \|\theta^*_T - \thetatilde_T\|^2_{\sigmahat + \lambda I} \nonumber \\
    &\leq \frac{\tau \kappa \mbar T (1 - \gamma)}{2 \mubar (1 - \gamma^{T-1})} \|\theta^*_T - \thetatilde_T\|^2_{\sigmahat + \lambda I}. \label{eq: regret_analysis_finish}
\end{align}

\subsubsection{Complexity Analysis}
\label{appendix: complexity analysis - regret - uniform}

In order to investigate the complexity of the regret bound, we set the value of $\gamma$ using $T, B_T$. We first set $\gamma$ as 
\begin{align} 
    \gamma &= 1 - \left(\frac{B_T}{T}\right)^{3/4}. \label{eq: gamma_form}
\end{align}

We apply \Cref{eq: gamma_form} in the estimation error $\|\theta^*_T - \thetatilde_T \|_{\sigmahat + \lambda I}$, with assumption of $\lambda \geq \conccoef$ from \Cref{lemma: chowdhury A.1. - matrix concentration inequality}, while ignoring the logarithmic factor:
\begin{align}
    2\sqrt{\lambda}W& &&&(=d^{\tfrac{1}{4}}\ n^{-\tfrac{1}{4}}) \nonumber \\
    \frac{2C_1}{\tau \cs}\sqrt{\frac{d + \log(1 / \delta)}{n}}& &&&(=d^{\tfrac{1}{2}} \ n^{-\tfrac{1}{2}}) \nonumber \\
    \frac{16L \Rs \mbar}{T(1 - \gamma)^{\tfrac{3}{2}}} \sqrt{\frac{d\mbar}{n}} B_T & &&&(=d^{\tfrac{1}{2}}\ B_T^{-\tfrac{1}{8}} \ T^{-\tfrac{3}{8}}) \label{eq: comparison_complexity_uniform}
\end{align}

Here, we note that from \Cref{assumption: temporal_distribution_new}, $n = \Theta(T)$. This allows us to consider the complexity with respect to the dataset size $n$ and $T$ together. We can conclude from \Cref{eq: comparison_complexity_uniform} that the complexity bound of the entire estimation error is $O(d^{\tfrac{1}{2}} \ T^{-\tfrac{1}{4}})$. 
By setting the value of $T$ to a sufficiently large one, making $1 - \gamma^{T-1} \geq \tfrac{1}{2}$, then the complexity of $\omegaupp(T, \gamma)$ is
\begin{align}
    T(1 - \gamma) &&&(=B_T^{\tfrac{3}{4}}\ T^{\tfrac{1}{4}}). \label{eq: complexity_omega_gamma_uniform}
\end{align}

Finally we present the total complexity bound of the algorithm, by applying the complexity of $\omegaupp(T, \gamma)$ in \Cref{eq: complexity_omega_gamma_uniform} to the squared estimation error $\|\theta^*_T - \thetatilde_T\|^2_{\sigmahat + \lambda I}$:
\begin{align}
    \Roff = O(d\ B_T^{\tfrac{3}{4}}\ T^{-\tfrac{1}{4}}) \nonumber \\
    = O(d\ B_T^{\tfrac{3}{4}}\ n^{-\tfrac{1}{4}}).
\end{align}

\newpage

\subsection{Theoretical Analysis of NS-DPO under stationary preferences}
\label{appendix: stationary case}

\begin{cormod}{corollary: regret bound - offline - stationary}{}
    (Regret bound under stationary preferences) Let   
    $B_T \rightarrow 0$, $\delta \in (0, \tfrac{1}{2}], \tau > 0$. Let $\thetatilde_T \in \Theta$ denote the minimiser of the NS-DPO loss (\Cref{eq: NS-DPO loss - offline analysis}). Then, for $\lambda \geq \conccoef$, some constant $C_1 > 0$, $\gamma = 1 - \left(\tfrac{B_T}{T}\right)^{\alpha}$ and $\alpha \in (0, \frac{2}{3})$, we have with probability at least $1 - 2\delta$:\looseness=-1
    \begin{align}
        \lim_{B_T \rightarrow 0}\Roff < \underbrace{\frac{4\tau \kappa \mbar}{\mubar}}_{\text{Pre-factor}} \Bigg(
        \sqrt{\lambda}W + \frac{C_1}{\tau \cs}\sqrt{\frac{d + \log(1 / \delta)}{n}} \Bigg)^2~, \nonumber
    \end{align}
and recover the complexity of $\Roff = O(n^{-\tfrac{1}{2}})$ under stationary preferences.
\end{cormod}

We show that under certain conditions, NS-DPO's regret bound recovers $O(n^{-\tfrac{1}{2}})$. We first analyse the estimation error in the limit $B_T \rightarrow 0$. Consider the estimation error bound in \Cref{theorem: estimation error - offline - uniform}:
\begin{align}
        \|\thetatilde_T - \theta^*_T\|_{\sigmahat + \lambda I} &\leq \underbrace{2\sqrt{\lambda}W + \frac{2C_1}{\tau \cs}\sqrt{\frac{d + \log(1 / \delta)}{n}}}_{\text{learning}} + \underbrace{\frac{16L \Rs \mbar}{T(1 - \gamma)^{\tfrac{3}{2}}} \sqrt{\frac{d\mbar}{n}} B_T}_{\text{tracking}},
    \end{align}
in which the \textit{tracking} term depends upon $\gamma$ and $B_T$. In the regret bound, we write $\gamma$ in terms of $B_T$ the form of
\begin{equation}
    \gamma = 1 - \left(\frac{B_T}{T}\right)^\alpha,
\end{equation}
where $\alpha \in \mathbb{R}$. We obtain $1 - \gamma = \left(\frac{B_T}{T}\right)^\alpha$ by rearranging terms. Substituting $B_T$ back into the estimation error bound, we find that the tracking term reduces to $16LR_{\sigma, \tau}\mbar T^{\tfrac{3}{2}\alpha-1}B_T^{1-\tfrac{3}{2}\alpha}\sqrt{\frac{d\mbar}{n}}$. By inspection, for $0 < \alpha < \frac{2}{3}$ the tracking term tends to $0$ as $B_T \rightarrow 0$. Thus we conclude that
\begin{equation} \label{eq: tracking-to-zero}
     \lim_{B_T \rightarrow 0} \Bigg(2\sqrt{\lambda}W + \frac{2C_1}{\tau \cs}\sqrt{\frac{d + \log(1 / \delta)}{n}} + \frac{16L \Rs \mbar}{T(1 - \gamma)^{\tfrac{3}{2}}} \sqrt{\frac{d\mbar}{n}} B_T\Bigg) = 2\sqrt{\lambda}W + \frac{2C_1}{\tau \cs}\sqrt{\frac{d + \log(1 / \delta)}{n}}.
\end{equation}

We now consider the regret bound in \Cref{theorem: regret bound - offline - uniform}:
\begin{equation}\label{eq: regret-restate}
    \Roff \leq \underbrace{\frac{\tau \kappa \mbar 
 T (1 - \gamma)}{2 \mubar (1 - \gamma^{T-1})}}_{\text{Pre-factor}} \Bigg(
        2\sqrt{\lambda}W + \frac{2C_1}{\tau \cs}\sqrt{\frac{d + \log(1 / \delta)}{n}} + \underbrace{\frac{16L \Rs \mbar}{T(1 - \gamma)^{\tfrac{3}{2}}} \sqrt{\frac{d\mbar}{n}} B_T}_{\text{Tracking}} \Bigg)^2.
\end{equation}
Here we note that the tracking term and the pre-factor term are dependent upon $\gamma$. Using the product rule of limits, we analyse the limit of the pre-factor and tracking terms independently and then multiply them together. Using L'Hopital's rule, the pre-factor term in \Cref{eq: regret-restate} in the limit $B_T \rightarrow 0$ becomes
\begin{align}
   \lim_{B_T \rightarrow 0}\frac{\tau \kappa \mbar 
 T (1 - \gamma(B_T))}{2 \mubar (1 - \gamma(B_T)^{T-1})} &= \lim_{B_T \rightarrow 0}\frac{\tau \kappa \mbar 
 T (\frac{B_T}{T})^{\alpha}}{2 \mubar (1 - (1 - (\frac{B_T}{T})^{\alpha})^{T-1}} \nonumber \\ 
\end{align}
We remove terms that do not depend upon $B_T$ for simplicity and then apply L'Hopital's rule:
\begin{align}
\lim_{B_T \rightarrow 0}\frac{(\frac{B_T}{T})^{\alpha}}{(1 - (1 - (\frac{B_T}{T})^{\alpha})^{T-1}} &= \lim_{B_T \rightarrow 0}\frac{1}{(T-1) (1- (\frac{B_T}{T})^\alpha)^{T-2}} \\
& = \frac{1}{T-1}
\end{align}

thus finding the limit of the pre-factor term. As $T > 1$, $\frac{\tau \kappa \mbar T}{2 \mubar (T-1)} < \frac{\tau \kappa \mbar}{\mubar}$, we use our analysis from the estimation bound and set $0 < \alpha < \frac{2}{3}$, such that the limit of the tracking term is $0$ as expected in stationary scenarios. We can now write the regret bound as
\begin{equation}\label{eq: regret-restate stationary}
    \lim_{B_T \rightarrow 0}\Roff < \underbrace{\frac{4\tau \kappa \mbar}{\mubar}}_{\text{Pre-factor}} \Bigg(
        \sqrt{\lambda}W + \frac{C_1}{\tau \cs}\sqrt{\frac{d + \log(1 / \delta)}{n}} \Bigg)^2.
\end{equation}
and recover the result of $\mathcal{O}(n^{-1/2})$ in \Cref{corollary: regret bound - offline - stationary}.

\subsection{Details of applying Bernstein's inequality}
\label{appendix: applying Bernstein's inequality}

We restate the norm to investigate:
\begin{equation}
    \|\frac{1}{n} \sum_{i=1}^n \tau \gamma^{T-1-t_i} \epsilon_i \phihat_i\|_{(\sigmatilde + \lambda I)^{-1}}. \label{eq: hsu_target}
\end{equation}

We then define two vectors $V$ and $Z$, followed by a matrix $M$:
\begin{align}
    V &= [\epsilon_1, \ldots, \epsilon_n], \label{eq: hsu_def_V}\\
    Z &= [\gamma^{T-1-t_1}\phihat_1, \ldots, \gamma^{T-1-t_n}\phihat_n], \label{eq: hsu_def_Z} \\
    M &= \frac{1}{n^2}Z (\sigmatilde + \lambda I)^{-1} Z^\intercal. \label{eq: hsu_def_M}
\end{align}

We then express \Cref{eq: hsu_target} using $V, Z, M$:
\begin{align}
    \|\frac{1}{n} \sum_{i=1}^n \tau \gamma^{T-1-t_i} \epsilon_i \phihat_i\|_{(\sigmatilde + \lambda I)^{-1}} &= \sqrt{\tau^2 V^\intercal M V}. \label{eq: hsu_target_to_vzm}
\end{align}

We here recall the definition of $\epsilon_i$, which is a 1-sub-Gaussian random variable:
\begin{align}
    \epsilon_i &= o_i - \sigma(\tau\langle \phihat_i, \theta^*_{t_i} - \thetaref\rangle), \nonumber \\
    \Eb_{o_i \sim p_{t_i}(a_i \succ a'_i | x_i)}[\epsilon_i] &= 0, \label{eq: hsu_epsilon_i} \\
    \mathrm{Var}_{o_i \sim p_{t_i}(a_i \succ a'_i | x_i)}[\epsilon_i] &= \Eb_{o_i \sim p_{t_i}(a_i \succ a'_i | x_i)}[\epsilon_i^2] - (\Eb_{o_i \sim p_{t_i}(a_i \succ a'_i | x_i)}[\epsilon_i])^2 \leq 1.
\end{align}

As stated in \citep{hsu2012tail}, the Bernstein’s inequality for sub-Gaussian random variables in quadratic form implies
\begin{align}
    \tau^2 V^\intercal M V &\leq \tau^2 \left(
        \text{tr}(M) + 2\sqrt{\text{tr}(M^\intercal M) \log (1/\delta)} + 2\|M\| \log (1/\delta)
    \right) \nonumber \\
    &\leq \tau^2 \cdot C_1 \cdot \frac{d + \log (1/\delta)}{n}, \label{eq: hsu_applying_theorem}
\end{align}
for some $C_1 > 0$, while $\|M\| = \lmax(M)$. Here we used the definition of $\sigmatilde$ in \Cref{eq: sigma_hat - weighted sample covariance matrix} to show $\sigmatilde = \frac{1}{n}Z^\intercal Z$, and derive for $\lambda > 0$
\begin{align}
    M &\prec \frac{1}{n^2}Z (\sigmatilde)^{-1} Z^\intercal = \frac{1}{n} I, \\
    \text{tr}(M) &\leq d/n, \\   
    \text{tr}(M^\intercal M) &\leq d/n^2, \\
    \|M\| &\leq 1/n. \label{eq: hsu_details}
\end{align}

\end{document}